\documentclass[sn-mathphys-num]{sn-jnl}

\usepackage{graphicx}%
\usepackage{multirow}%
\usepackage{amsmath,amssymb,amsfonts}%
\usepackage{amsthm}%
\usepackage{mathrsfs}%
\usepackage[title]{appendix}%
\usepackage{xcolor}%
\usepackage{textcomp}%
\usepackage{manyfoot}%
\usepackage{booktabs}%
\usepackage{algorithm}%
\usepackage{algorithmicx}%
\usepackage{algpseudocode}%
\usepackage{listings}%
\usepackage{subcaption}
\usepackage{cleveref}
\usepackage{rotating}  
\usepackage{booktabs}  
\usepackage{longtable}
\usepackage{adjustbox}
\usepackage{xltabular}
\usepackage{shellesc}

\theoremstyle{thmstyleone}%
\theoremstyle{thmstyletwo}%

\theoremstyle{thmstylethree}%

\begin{document}

\title[Article Title]{DRIFT open dataset: A drone-derived intelligence for traffic analysis in urban environment}
 
\author[1]{\fnm{Hyejin} \sur{Lee}}
\equalcont{These authors contributed equally to this work.}

\author[2]{\fnm{Seokjun} \sur{Hong}}
\equalcont{These authors contributed equally to this work.}

\author[2]{\fnm{Jeonghoon} \sur{Song}}

\author[1]{\fnm{Haechan} \sur{Cho}}

\author[3, 4]{\fnm{Zhixiong} \sur{Jin}}

\author[2]{\fnm{Byeonghun} \sur{Kim}}

\author[2]{\fnm{Joobin} \sur{Jin}}

\author[5]{\fnm{Jaegyun} \sur{Im}}

\author*[5]{\fnm{Byeongjoon} \sur{Noh}}\email{powernoh@sch.ac.kr (B. Noh)}

\author*[1]{\fnm{Hwasoo} \sur{Yeo}}\email{hwasoo@kaist.ac.kr (H. Yeo)}

\affil[1]{\orgdiv{Department of Civil and Environmental Engineering}, \orgname{Korea Advanced Institute of Science and Technology (KAIST)}, \orgaddress{\street{291 Daehak-ro}, \city{Daejeon}, \postcode{34141}, \country{South Korea}}}

\affil[2]{\orgdiv{Department of Future Convergence Technology}, \orgname{Soonchunhyang University}, \orgaddress{\street{22 Soonchunhyang-ro}, \city{Asan}, \postcode{31538}, \country{South Korea}}}

\affil[3]{\orgdiv{LICIT-ECO7}, \orgname{Univ Gustave Eiffel}, \city{Lyon}, \postcode{69675}, \country{France}}

\affil[4]{\orgdiv{Urban Transport Systems Laboratory (LUTS)}, \orgname{École Polytechnique Fédérale de Lausanne (EPFL)}, \city{Lausanne}, \postcode{1015}, \country{Switzerland}}

\affil[5]{\orgdiv{Department of AI and Big Data}, \orgname{Soonchunhyang University}, \orgaddress{\street{22 Soonchunhyang-ro}, \city{Asan}, \postcode{31538}, \country{South Korea}}}


\abstract{Reliable traffic data are essential for understanding urban mobility and developing effective traffic management strategies. This study introduces the \textbf{DR}one-derived \textbf{I}ntelligence \textbf{F}or \textbf{T}raffic analysis (DRIFT) dataset, a large-scale urban traffic dataset collected systematically from synchronized drone videos at approximately 250 meters altitude, covering nine interconnected intersections in Daejeon, South Korea. DRIFT provides high-resolution vehicle trajectories that include directional information, processed through video synchronization and orthomap alignment, resulting in a comprehensive dataset of 81,699 vehicle trajectories. Through our DRIFT dataset, researchers can simultaneously analyze traffic at multiple scales - from individual vehicle maneuvers like lane-changes and safety metrics such as time-to-collision to aggregate network flow dynamics across interconnected urban intersections. The DRIFT dataset is structured to enable immediate use without additional preprocessing, complemented by open-source models for object detection and trajectory extraction, as well as associated analytical tools. DRIFT is expected to significantly contribute to academic research and practical applications, such as traffic flow analysis and simulation studies. The dataset and related resources are publicly accessible at \textcolor{blue}{https://github.com/AIxMobility/The-DRIFT}.}







\maketitle

\section{Introduction}\label{chap:introduction}

Urban traffic research critically depends on reliable traffic data, including comprehensive traffic flow characteristics and detailed vehicle trajectories \citep{berghaus2024vehicle}. Traffic data form the foundation for predictive modeling, policy formulation, and advanced simulations, playing an indispensable role in understanding urban mobility and traffic flow dynamics \citep{antoniou2011synthesis, lemonde2021integrative, ouallane2022fusion}. However, conventional traffic data collection methods, such as static sensors (e.g., cameras, loop detectors, etc.) and mobile sensors (e.g., GPS, smartphones, etc.), frequently encounter significant limitations such as high installation costs, restricted spatial coverage, low penetration rates, and restricted accuracy \citep{tu2021estimating, tasgaonkar2020vehicle, chriki2021deep, park2017vision}. 

To overcome these limitations, previous studies have employed simulation-based datasets, such as AIMSUN \citep{casas2010traffic} and SUMO \citep{SUMO2018}, either exclusively or in combination with real-world data to support traffic modeling and analytical research. Nevertheless, simulation-based data inherently struggle to fully reflect the complexities and multimodal characteristics of real-world traffic interactions, owing to simplified assumptions and limited representation of actual human behaviors \citep{chen2024roadside}. As an alternative, drone-based traffic monitoring combined with advanced computer vision methods has emerged as a promising strategy. Drone imagery enables extensive coverage of road networks, capturing detailed spatial and temporal interactions among diverse traffic agents with minimal reliance on traditional ground-based infrastructure \citep{bakirci2025vehicular, gupta2022monitoring}. Existing drone-based datasets, such as HighD \citep{krajewski2018highd}, CitySim \citep{zheng2024citysim}, pNEUMA \citep{barmpounakis2020new}, Songdo \citep{fonod_2025_13828408}, underscore the significant potential of drone technology for traffic data collection and analysis. However, these datasets often present limitations in terms of spatial continuity necessary for analyzing complex network-level interactions in urban settings, lack systematic validation of trajectory accuracy, and may require additional preprocessing to ensure practical applicability.

Building upon this emerging approach, this study introduces a ``\textbf{DR}one-derived \textbf{I}ntelligent \textbf{F}or \textbf{T}raffic analysis (DRIFT)'' open dataset, collected via synchronized multi-drone operations conducted at nine interconnected roads including intersections in Daejeon, South Korea. The datasets were acquired during urban peak hours, capturing dynamic traffic states and real-world commuting patterns with continuous spatial coverage. This collaborative effort between the Korea Advanced Institute of Science and Technology (KAIST) and Soonchunhyang University aims to offer comprehensive, reliable, and systematically structured urban traffic data. 

This dataset provides detailed vehicle trajectories and a range of critical traffic parameters extracted using object detection model, the YOLOv11 model \citep{yolo11_ultralytics} trained on a large number of manually-annotated dataset with the oriented bounding box (OBB) method \citep{yao2022improving} and object tracking algorithm, Bytracker method \citep{zhang2022bytetrack}. The OBB-based method achieves higher precision in identifying vehicle orientations compared to the traditional axis-aligned bounding box (AABB) approaches, thus enabling more accurate extraction of vehicle states as microscopic indicators, such as vehicle speeds, positions, and lane occupancy. The resulting dataset demonstrates strong technical rigor and performance reliability, achieving high detection accuracy (99.2\%) and a trajectory continuity rate of 96.98\%, verified through systematic validation procedures. Its strengths lie not only in data precision but also in spatial continuity, extensive coverage of complex intersection networks, and full automation of the data processing pipeline. These attributes position DRIFT as a research-ready dataset, minimizing the preprocessing burden and directly supporting advanced traffic modeling, safety analysis, and simulation applications.

Unlike previous datasets that primarily focus on highway segments or short, isolated road sections, the DRIFT dataset simultaneously covers multiple contiguous intersections, facilitating integrated analysis of traffic flow from microscopic (individual vehicle behaviors) to mesoscopic (network-level interactions) scales. For example, time-space diagrams derived from DRIFT data effectively illustrate individual vehicle trajectories in spatiotemporal coordinates, supporting quantitative analyses of speed–density relationships, congestion propagation, and upstream-downstream interactions.

Ultimately, the DRIFT dataset serves as a comprehensive resource for investigating complex interactions within urban traffic networks. Its extensive coverage and methodological rigor provide valuable insights applicable to both scholarly research and practical applications such as traffic management initiatives, advancing data-driven urban mobility planning, and informed policymaking.

\section{Literature review}\label{chap:literature}
This section reviews existing studies on drone-based traffic data collection, compares the features and applicability of representative datasets, and discusses the distinct advantages of the proposed DRIFT dataset. Recent advancements in drone technology have facilitated the large-scale acquisition of high-resolution traffic trajectory data. Drone-derived datasets utilize an aerial perspective, which allows unobstructed observation of road users, thus holding significant potential for applications such as traffic safety evaluation, digital twin development, and traffic flow modeling. However, the usability of existing datasets varies depending on spatial coverage (e.g., freeways vs. urban networks, single intersections vs. multiple interconnected intersections), dataset scale, and data availability formats. Table~\ref{tbl:datasets} summarizes the specifications of currently available drone-based open traffic datasets.

\begin{sidewaystable}
\caption{Comparative table of drone-based traffic datasets} \label{tbl:datasets} 
\begin{tabular*}{\textheight}{@{}llllllll@{}}
\toprule%
\textbf{Dataset} 
& \textbf{\# of traj.} 
& \textbf{Road type} 
& \textbf{Coverage} 
& \textbf{OBB} 
& \textbf{Detection model}  \\ \midrule

highD \citep{krajewski2018highd}
& \textasciitilde110\,k 
& Highway
& 6 segments (\textasciitilde420\,m each) 
& \(\times\)
& - \\

INTERACTION  \citep{zhan2019interaction}
& -
& Urban/Highway
& 11 distinct sites 
& \(\times\)
& -\\

inD  \citep{bock2020ind}
& \textasciitilde13.5\,k 
& Urban
& Small-scale intersection areas 
& \(\times\)
& -\\

pNEUMA  \citep{barmpounakis2020new}
& \textasciitilde500\,k 
& Urban
& \textasciitilde1.3\,km\textsuperscript{2} 
& \(\times\)
& -\\

rounD  \citep{krajewski2020round}
& \textasciitilde13.7\,k 
& Urban
& Each roundabout area \textasciitilde140\(\times\)70\,m 
& \(\times\)
& -\\

exiD  \citep{moers2022exid}
& \textasciitilde69\,k 
& Highway
& 7 locations (each \textasciitilde420\,m) 
& \(\times\)
& -\\

SIND  \citep{xu2022drone}
& \textasciitilde13\,k 
& Urban
& -\
& \checkmark 
& YOLOv5 \citep{yolov5}\\

CitySim  \citep{zheng2024citysim}
& -
& Urban/Highway 
& - 
& \checkmark 
& Mask R-CNN \citep{he2017mask} + CSRT tracking \citep{lukezic2017discriminative}\\

Songdo \citep{fonod_2025_13828408}
& \textasciitilde1{,}000\,k 
& Urban
& 20 junctions in city center
& \(\times\)
& YOLOv8 \citep{yolov8_ultralytics}\\

\textbf{DRIFT (ours)}
& 81,699
& Urban
& 2.3 \,km
& \checkmark 
& YOLOv11 + ByteTracker\\

\bottomrule
\end{tabular*}
\footnotetext{-: not publicly disclosed or not applicable}
\footnotetext{\checkmark: yes, \(\times\): no}
\end{sidewaystable}

Among the earliest examples is the highD dataset \citep{krajewski2018highd}, which collected extensive naturalistic vehicle trajectories on German freeways near Cologne. Employing a DJI Phantom 4 Pro Plus to capture approximately 16.5 hours of 4K footage over a 420-meter segment, this dataset comprises more than 110,000 vehicle trajectories. The consistent capture altitude and detailed annotations—including lane changes and vehicle types—positioned highD as a benchmark for automated driving validation and driver behavior analysis. Subsequently, the exiD dataset \citep{moers2022exid} highlighted complex merging and diverging maneuvers at Autobahn on-ramps and off-ramps through approximately 69,000 vehicle trajectories, thereby further supporting automated driving system evaluations.

Efforts to apply these methods to urban settings include the rounD dataset \citep{krajewski2020round}, which examined three roundabouts in Germany and documented the interactions of 13,746 road users. The rounD dataset captured not only entering, circulating, and exiting traffic but also intricate negotiations among pedestrians, motorcycles, and bicycles. The inD dataset \citep{bock2020ind} built upon this foundation by gathering more than 13,500 trajectories at four urban intersections in Aachen, Germany, including detailed data on pedestrians and cyclists to promote research on intersection safety. INTERACTION \citep{zhan2019interaction} expanded coverage to a variety of roadway structures, such as roundabouts, ramps, and intersections, focusing on cooperative and critical driving scenarios. Nonetheless, it has been criticized for concentrating on particular road user types, thereby limiting multi-modal analyses.

Among large-scale urban data collection initiatives, the pNEUMA dataset \citep{barmpounakis2020new} is notable. Ten drones captured high-resolution footage spanning an area of 1.3 km\(^2\) in Athens, Greece, covering over 100 intersections and 30 bus stops across more than 10 km of roadway. The dataset features approximately 500,000 vehicle trajectories, facilitating systematic investigations of urban congestion propagation, lane changes, and multi-modal interactions. CitySim \citep{zheng2024citysim} similarly broadens coverage, encompassing freeways, intersections, and merging/diverging segments, as well as rotated bounding box annotations, 3D mapping, and signal timing data for advanced traffic safety research and modeling.

Another large-scale urban project is the Songdo dataset \citep{fonod_2025_13828408}, which gathered about one million vehicle trajectories from 20 major intersections in the Songdo International Business District in South Korea. By employing deep-learning-based object detection and trajectory stabilization techniques, this dataset achieved high accuracy, offering fresh opportunities for urban network analysis and intelligent transportation systems development. Nonetheless, even comprehensive datasets such as pNEUMA and Songdo have been found to require additional data cleaning in certain usage scenarios.

To address these issues, the DRIFT dataset proposed in this study is released in a preprocessed format following extensive object detection validation, stabilization, and error checking. Accordingly, users can directly proceed with traffic analysis or modeling without further cleansing and can more easily handle the spatial and temporal alignment challenges associated with multiple drone feeds. Furthermore, whereas previous datasets tended to focus on isolated intersections or specific highway stretches, DRIFT dataset observes a network of interconnected intersections, enabling integrated micro-, meso-, and macro-scale analyses—such as capturing congestion propagation between upstream and downstream intersections, lane changes, or fluctuations in traffic volume by time of day. In addition, DRIFT provides analytical and visualization tools, along with a pretrained model for detecting vehicles in drone imagery, allowing researchers to perform fine-tuning for customized applications. These features are especially advantageous in studies or policy-making processes that require a comprehensive view of metropolitan traffic flow.

\section{Experimental setup: Site and flight details}\label{chap:flight}

This section provides a concise overview of the experimental conditions and video recording outcomes from collecting traffic data via drones from an aerial perspective. Specifically, it defines the spatial boundaries of the study area, details the flight protocols—including flight altitude and recording duration—used for data acquisition and explains the video alignment and synchronization procedures necessitated by multi-drone operations.

\subsection{Site description}
\begin{figure}[h]
    \centering
    \includegraphics[width=\textwidth]{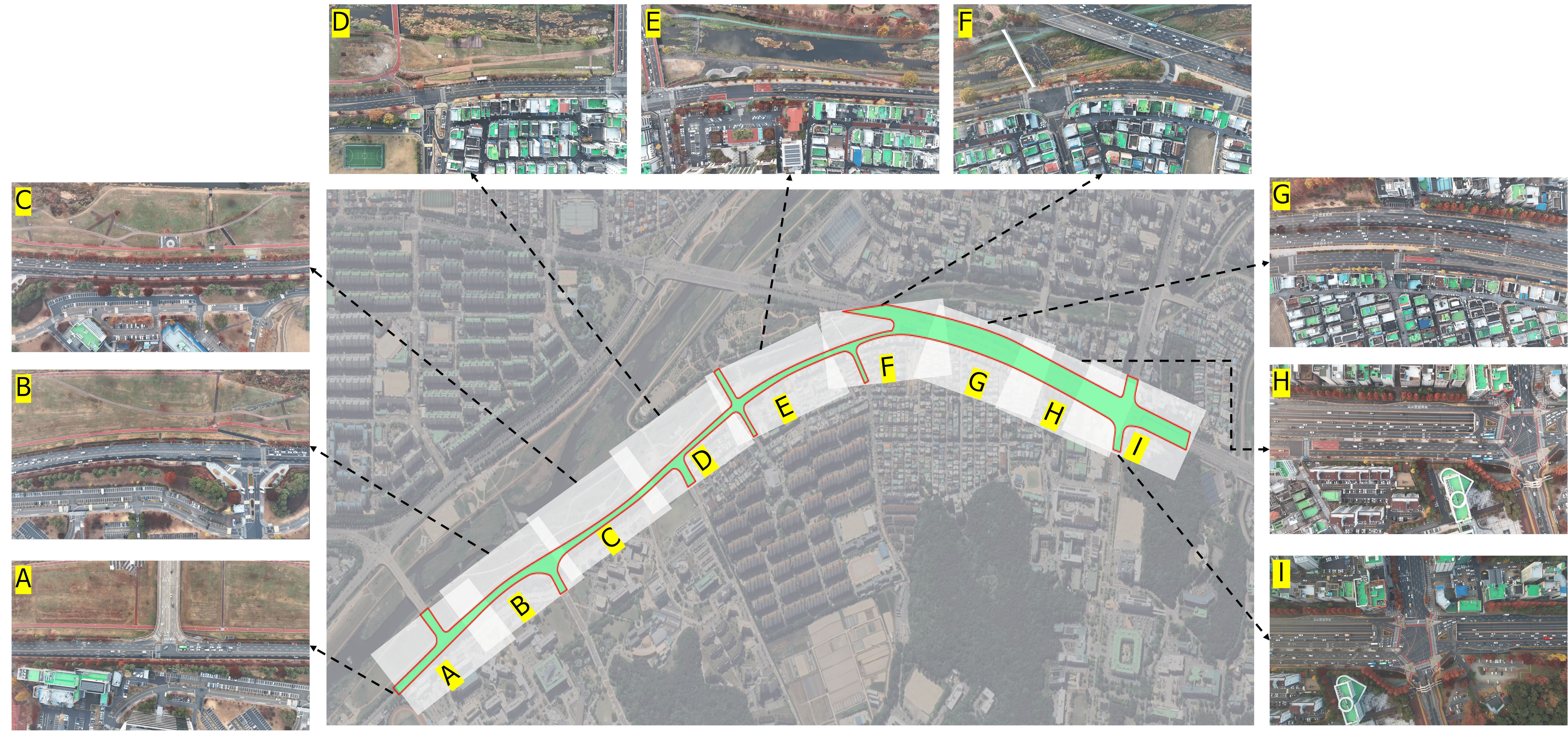}
    \caption{Drone-derived aerial views in nine sites in our experiment (Sties A to I). The green masks indicate the designated RoI for each site.}
    \label{fig:target-site}
\end{figure}

The experiment was conducted in a major traffic hub in Daejeon, South Korea, covering the section from 99 Daehak-ro (Chungnam National University Five-Way Intersection) to 291 Daehak-ro (KAIST). This section comprises approximately 2.6 km of continuous roadway. In our experiment, nine observation sites (Sites A to I) were strategically selected with spatially seamless monitoring. The designated study area and the highlighted lanes marking the Region of Interests (RoIs) are presented in Figure~\ref{fig:target-site}.

Each site consists of an average of 6 lanes, and is notably prone to severe traffic congestion during peak commuting hours. To capture representative traffic conditions, data collection was conducted between 7:30 AM and 9:20 AM.

\subsection{Flight protocols}
\begin{figure}[!ht]
    \centering
    \includegraphics[width=0.95\textwidth]{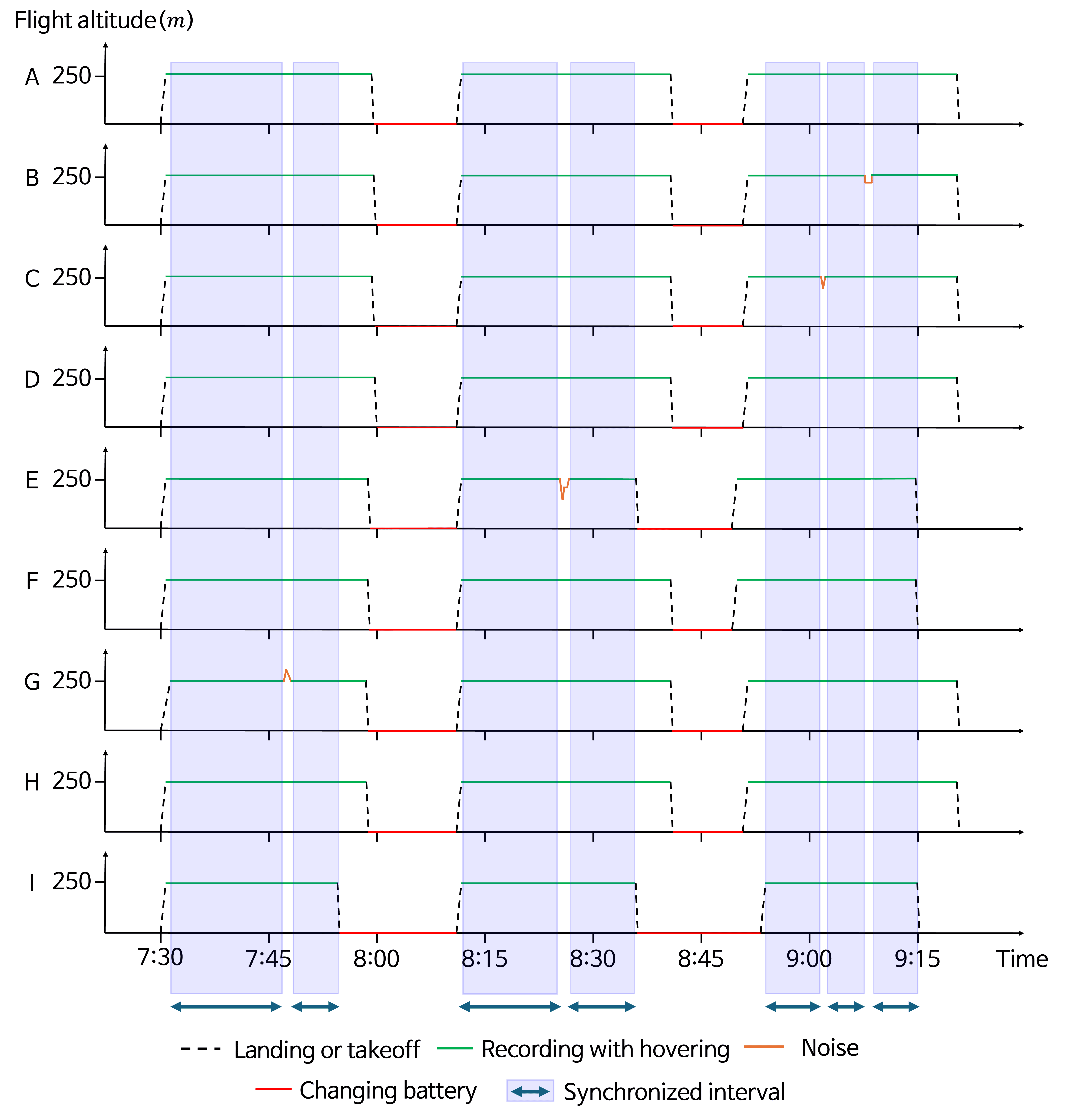}
    \caption{Synchronized drone flight periods across nine sites (A–I). Blue shading marks temporally synchronized intervals in our data collections, with a total synchronized duration of approximately 51 minutes and 43 seconds.}
    \label{fig:flight-result}
    \vspace{-10pt}
\end{figure}

Each drone operated at an altitude of approximately 250 meters, performing three synchronized flights, each lasting approximately 25 minutes. After each flight session, a 10-minute break was allocated for battery replacement. Figure~\ref{fig:flight-result} presents the flight schedules for each observation site, including intervals identified as noise due to battery replacement or altitude fluctuations exceeding the acceptable range (\(\pm\) 1 meter). Through this experiment, approximately 0.6 TB of raw video data was collected. Utilizing metadata extracted from the drone recordings, precise temporal synchronization across the interconnected intersection network was achieved.

All drone operations strictly complied with airspace regulations and safety guidelines, and prior flight approval was obtained through cooperation with Daejeon authorities.

\subsubsection{\textbf{Drone specification for recording}}
The drones used in our experiment was the commercially available Mavic 3 Pro model by DJI \citep{DJI_Mavic3Pro}. This drone offers a maximum flight time of approximately 43 minutes and hovering time of 37 minutes, with ascent and descent speeds of 8 m/s and 6 m/s, respectively. It has a maximum takeoff altitude of 6,000 meters and can withstand wind speeds of up to 12 m/s, ensuring stable operation under various weather conditions. In addition, when utilizing the vision positioning system, it provides high hover accuracy, with vertical and horizontal precision of 0.1 meters and 0.3 meters, respectively. The recording settings were configured to 4K resolution (3,840$\times$2,160 pixels) at approximately 30 fps to ensure efficient data processing while maintaining smooth and stable video quality. 

\subsection{Video alignment and synchronization}
The main objective of this process is to extract video frames captured simultaneously at all observation sites. 
Since our experiments are simultaneously conducted with multiple drones and multiple sites, it requires precise temporal alignment of the recorded footage. Multiple operators were involved, and variations in their experience led to differences in the timing of operations such as battery replacement, takeoff, and landing. As a result, the moments when drones reached their designated altitude and began recording were not always synchronized.

To address these inconsistencies, we meticulously performed temporal synchronization using drone metadata, including GPS timestamps, altitude logs, and takeoff/landing times. If the drone’s altitude change remained within 1 m over a 10-second interval during hovering, the recording was designated as stable. if it exceeded this range, it was treated as noise and removed. Guided by this criterion, we reviewed the recording status and timestamps of nine drones to ascertain whether the recorded data had been synchronized in time. As graphically depicted in Figure~\ref{fig:flight-result}, after aligning and synchronizing the videos, the total recording time is 51 minutes and 43 seconds. The first, second and third flights last 14:02, 20:29 and 17:12, respectively.

\section{DRIFT data curation}\label{chap:method}
This section outlines a systematic pipeline for extracting and structuring traffic data from drone-derived video footage. The proposed pipeline consists of three primary steps: preprocessing, object detection and tracking, and data validation and curation. In the preprocessing step, we convert drone video frames into real-world coordinates using homography-based orthophoto matching, apply frame stabilization to correct distortions from drone movements, and define lane RoIs for each experimental site. Next, vehicle detection is performed using the YOLOv11 model \citep{yolo11_ultralytics}, which uses polygon-based OBBs to enhance detection accuracy and effectively capture directional attributes inherent to aerial perspectives. The vehicle trajectories derived from detection results are then organized using the ByteTracker object-tracking algorithm \citep{zhang2022bytetrack}. Finally, the extracted trajectory data undergo quality validation to ensure reliability, after which they are systematically curated and integrated into the DRIFT open dataset.

The dataset constructed through the process supports comprehensive traffic-related studies at microscopic, mesoscopic, and macroscopic levels. In particular, the computer-vision automation in this pipeline enables the computation of vehicular metrics—such as speed, acceleration, and lane occupancy—as well as complex interaction-based measures, including Time-to-Collision (TTC) and lane-change behaviors. Consequently, the proposed approach is expected to enhance the scholarly utility of the dataset, supporting a wide spectrum of research from individual vehicle behavior analysis to large-scale traffic-flow assessments.

\subsection{Preprocessing}
\subsubsection{Orthophoto matching}
As the first step of preprocessing, we convert the video frame data into real-world coordinates with an orthophoto-based approach, providing the spatial foundation for accurate georeferencing. The orthomap corrects geometric distortions and enables precise spatial measurements.

\begin{figure}[!ht]
    \centering
    \includegraphics[width={0.75\textwidth}]{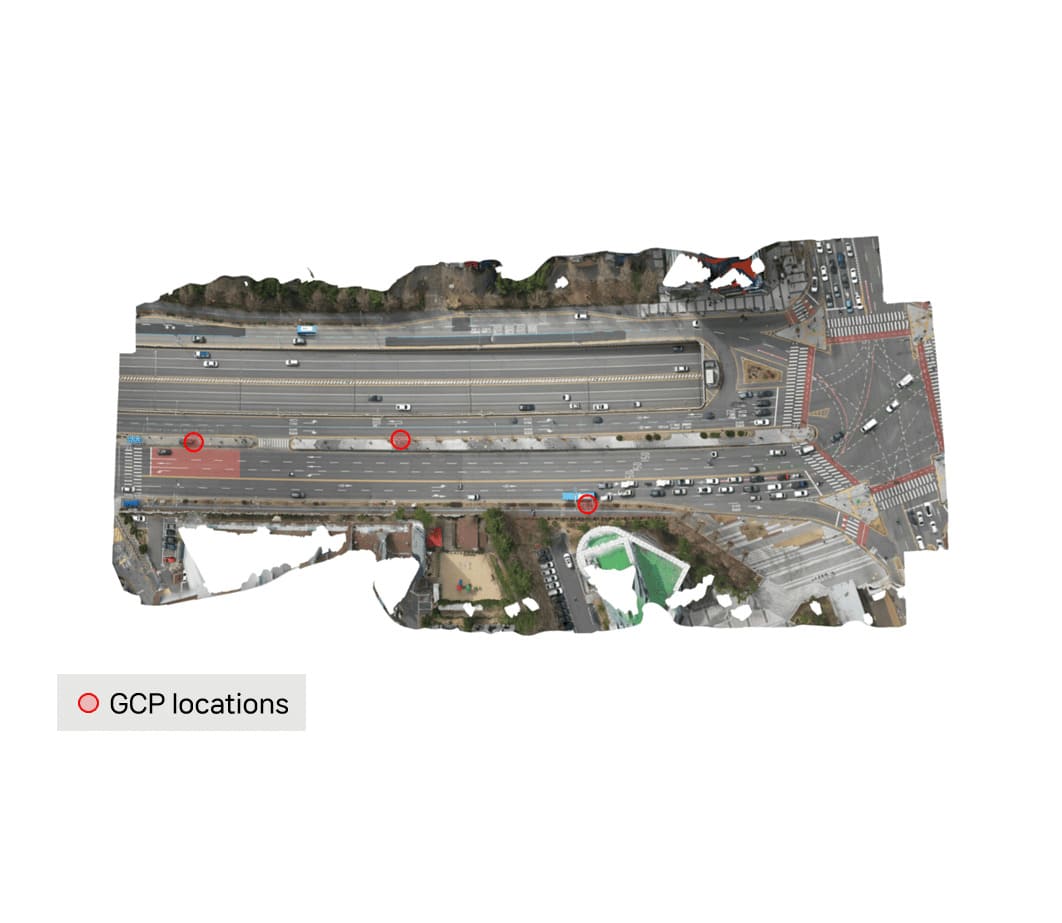}
    \caption{Example of orthophoto in Site H}
        \label{fig:orthomap}
\end{figure}

To construct the orthophoto, high-precision Ground Control Points (GCPs) were measured using the EMLID RS2+ GNSS receiver, as illustrated in Figure~\ref{fig:orthomap}. These GCPs were used in Agisoft Metashape \citep{AgisoftProfessionalEdition} to align a set of drone-captured images and generate a high-resolution orthophoto, which served as a reference for vehicle trajectory mapping. Feature points were then extracted from both the drone imagery and the orthophoto and matched to establish point correspondences. These correspondences enabled the estimation of a homography matrix, which was applied to the pixel-based vehicle trajectories in the drone footage. Because each pixel in the orthophoto corresponds to a real-world coordinate, the transformed output accurately reflects the spatial positions of vehicles within the observed environment.

\subsubsection{Frame stabilization in aerial drone videos}
The primary goal of frame stabilization is to correct fluctuations in drone videos caused by wind, drone vibration, and other environmental disturbances. To achieve precise object localization, this study performed frame stabilization to minimize spatial distortions and maintain frame-to-frame consistency. The stabilization process aligns all video frames to a single reference frame (\(I_r\)) selected from a segment where the drone maintains stable altitude and orientation.

\begin{figure}[h]
    \centering
    \includegraphics[width=\textwidth]{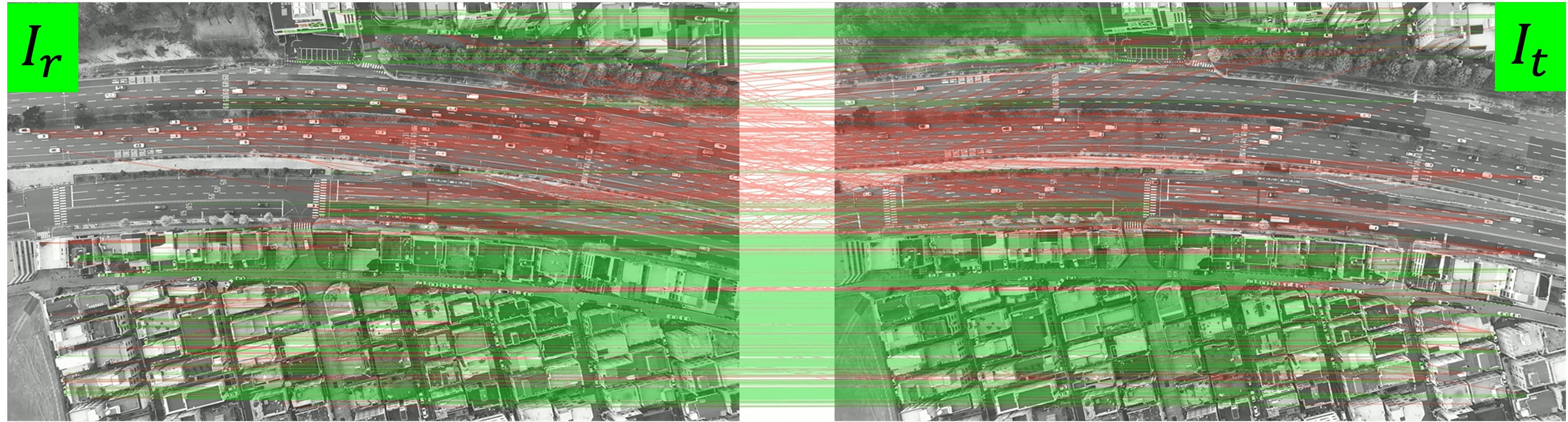}
    \caption{Feature detection and matching results between \( I_r \) and \( I_t \). Feature points were detected using ORB, and corresponding features were matched using the Brute-Force Matcher. Green lines: matched feature points, red lines: failed points.}
    \label{fig:stabilization1}
\end{figure}

Specifically, the Oriented FAST and Rotated BRIEF (ORB) algorithm \citep{rublee2011orb} is first employed to extract feature points \(\mathbf{k}_t=\{k_t^1,k_t^2,\dots,k_t^N\}\) and their corresponding descriptors \(\mathbf{d}_t=\{d_t^1,d_t^2,\dots,d_t^N\}\) from each input frame \(I_t\) in time \(t\). The Brute-Force Matcher is then applied to compute the Hamming distances between descriptors to establish correspondences between each input frame and the reference frame (see Figure~\ref{fig:stabilization1}).

From the set of matched feature points, a transformation matrix \(H_{\text{MAGSAC++}}\) is estimated using the MAGSAC++ algorithm \citep{barath2020magsac++}, known for its robustness against outliers. The resulting transformation matrix is then applied to perform a perspective transformation, generating a stabilized frame \(I^{\prime}_t\) that compensates for drone-induced movements and rotations.

However, if the number of matched feature points between frames is insufficient to reliably estimate the transformation matrix through standard techniques, a heuristic approach, denoted as the \textit{GeoAlign} transformation, is applied. This heuristic method comprises three sequential operations: scaling, rotation, and translation, ensuring effective alignment even in the absence of adequate feature matching. The \textit{GeoAlign} transformation is formulated as follows:

\begin{equation}
    H_{\text{GeoAlign}} = T_{\text{scale}} \cdot R_{\text{rotation}} \cdot T_{\text{translation}}
\end{equation}

\begin{figure}[h]
    \centering
    \includegraphics[width=\textwidth]{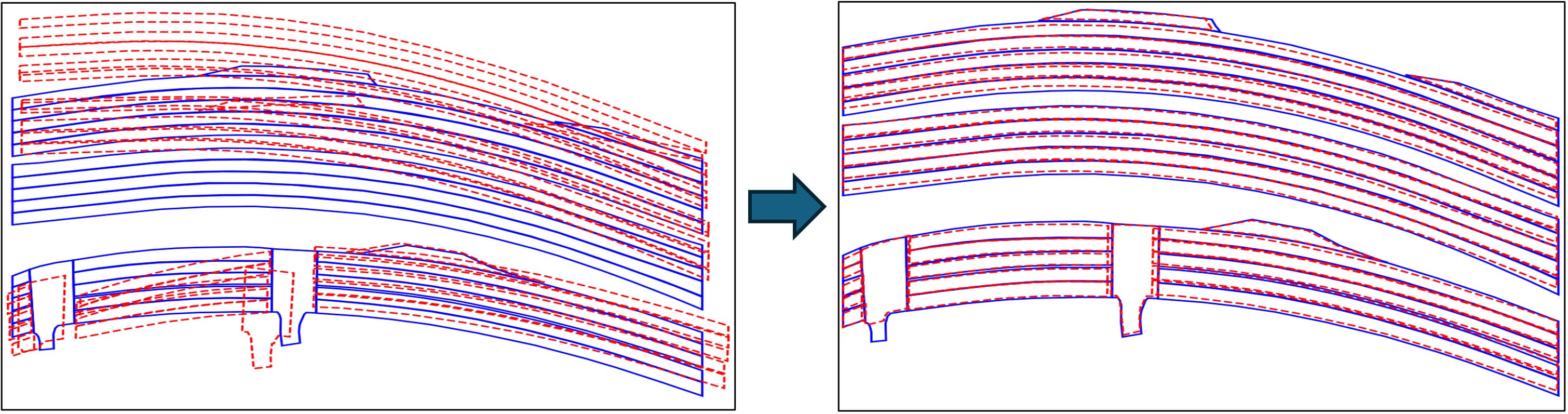}
    \caption{Transformation process using \( H_{\text{GeoAlign}} \) when feature matching fails. Blue regions: RoIs of \( I_r \), red dashed regions: RoI of \( I_t \).}
    \label{fig:stabilization2}
\end{figure}

Figure~\ref{fig:stabilization2} graphically depicts \textit{GeoAlign} transformation.

Finally, the stabilized frame \(I^{\prime}_t\) is generated by applying the transformation matrix to each input frame \(I_t\):

\begin{equation}
    I^{\prime}_t = \text{transform}(I_t, H),\quad H = \begin{cases}
    H_{\text{MAGSAC++}}, & \text{if feature matching succeeds}\\[6pt]
    H_{\text{GeoAlign}}, & \text{if feature matching fails}
    \end{cases}
\end{equation}

This streamlined stabilization enables to improve video consistency, providing reliable data for subsequent object detection, tracking, and comprehensive traffic analysis.

\subsubsection{Lane RoI settings}
\begin{figure}[!ht]
    \centering
    \begin{subfigure}[b]{0.32\textwidth}
        \includegraphics[width=\textwidth]{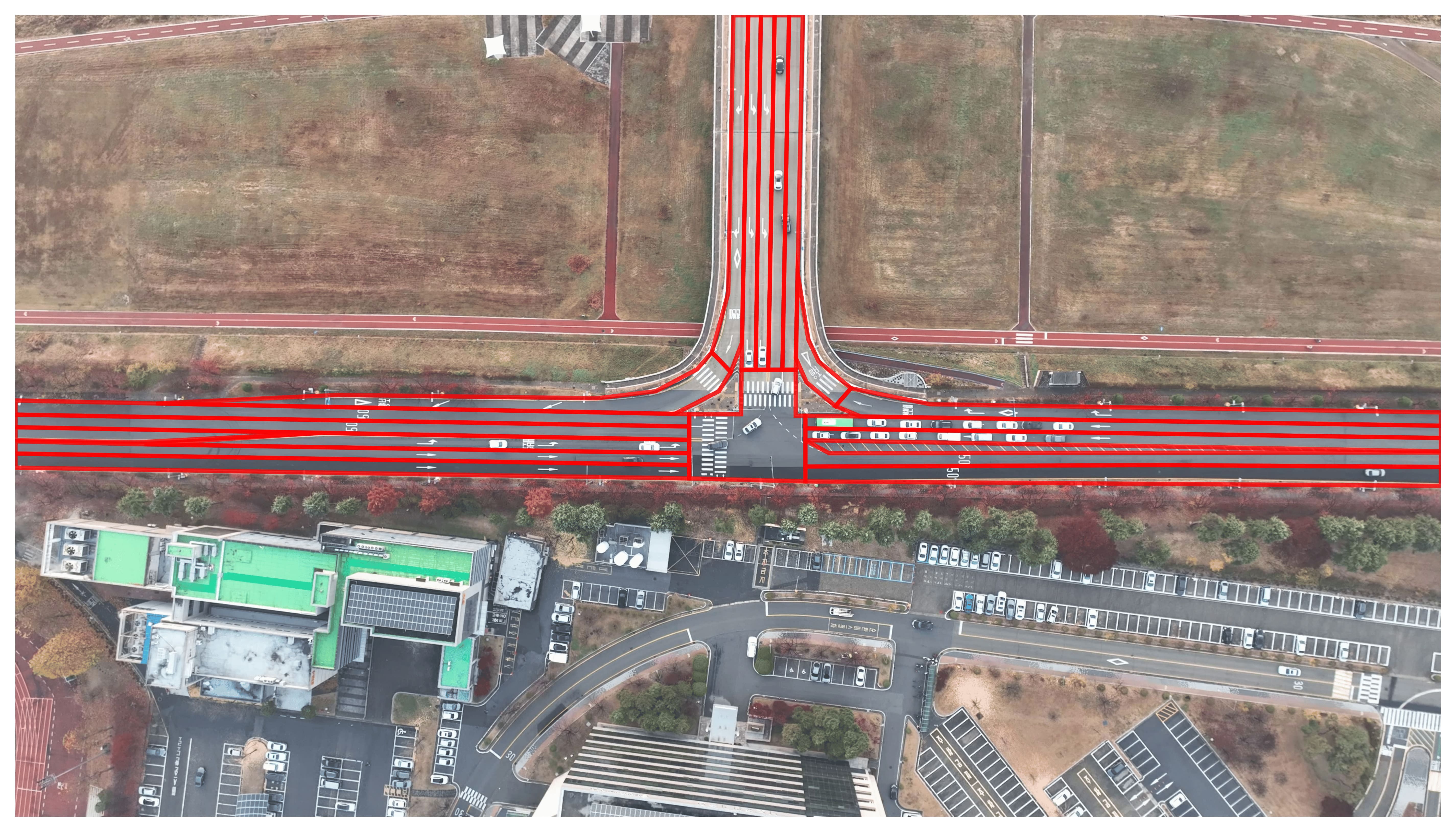}
        \caption{Site A}
        \label{subfig:site-A}
    \end{subfigure}
    \hfill
    \begin{subfigure}[b]{0.32\textwidth}
        \includegraphics[width=\textwidth]{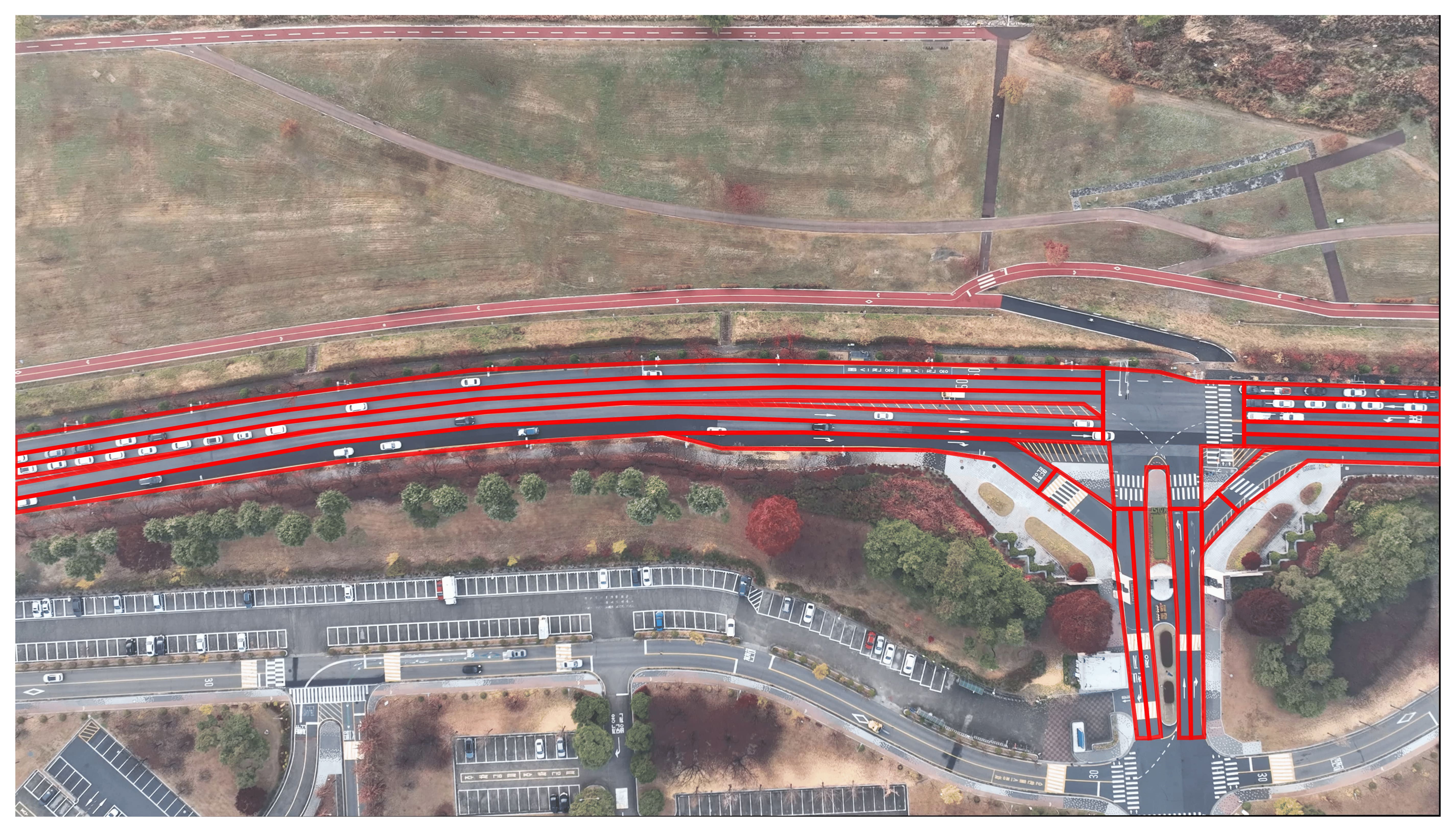}
        \caption{Site B}
        \label{subfig:site-B}
    \end{subfigure}
    \hfill
    \begin{subfigure}[b]{0.32\textwidth}
        \includegraphics[width=\textwidth]{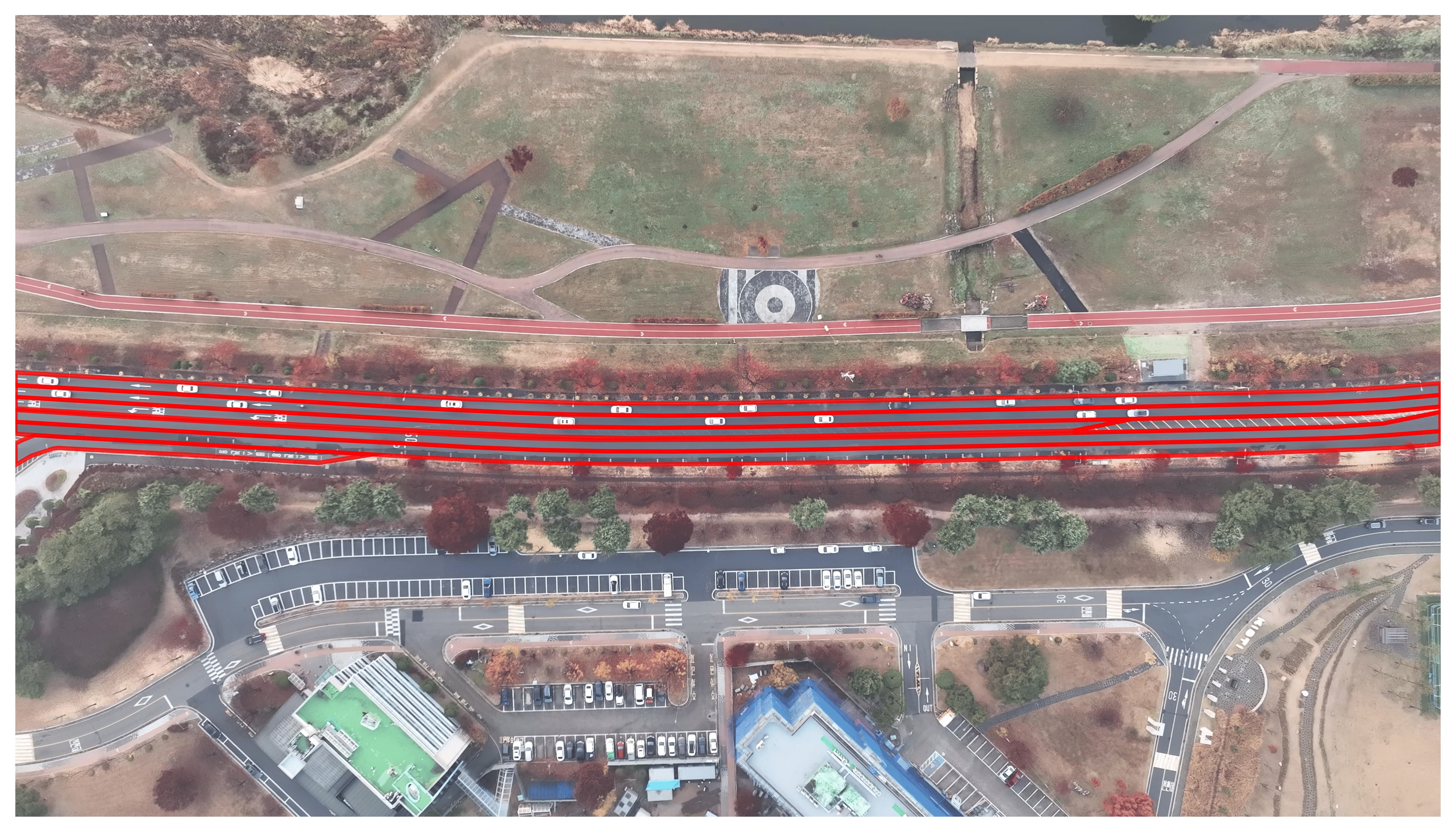}
        \caption{Site C}
        \label{subfig:site-C}
    \end{subfigure} 
    \hfill
    \begin{subfigure}[b]{0.32\textwidth}
        \includegraphics[width=\textwidth]{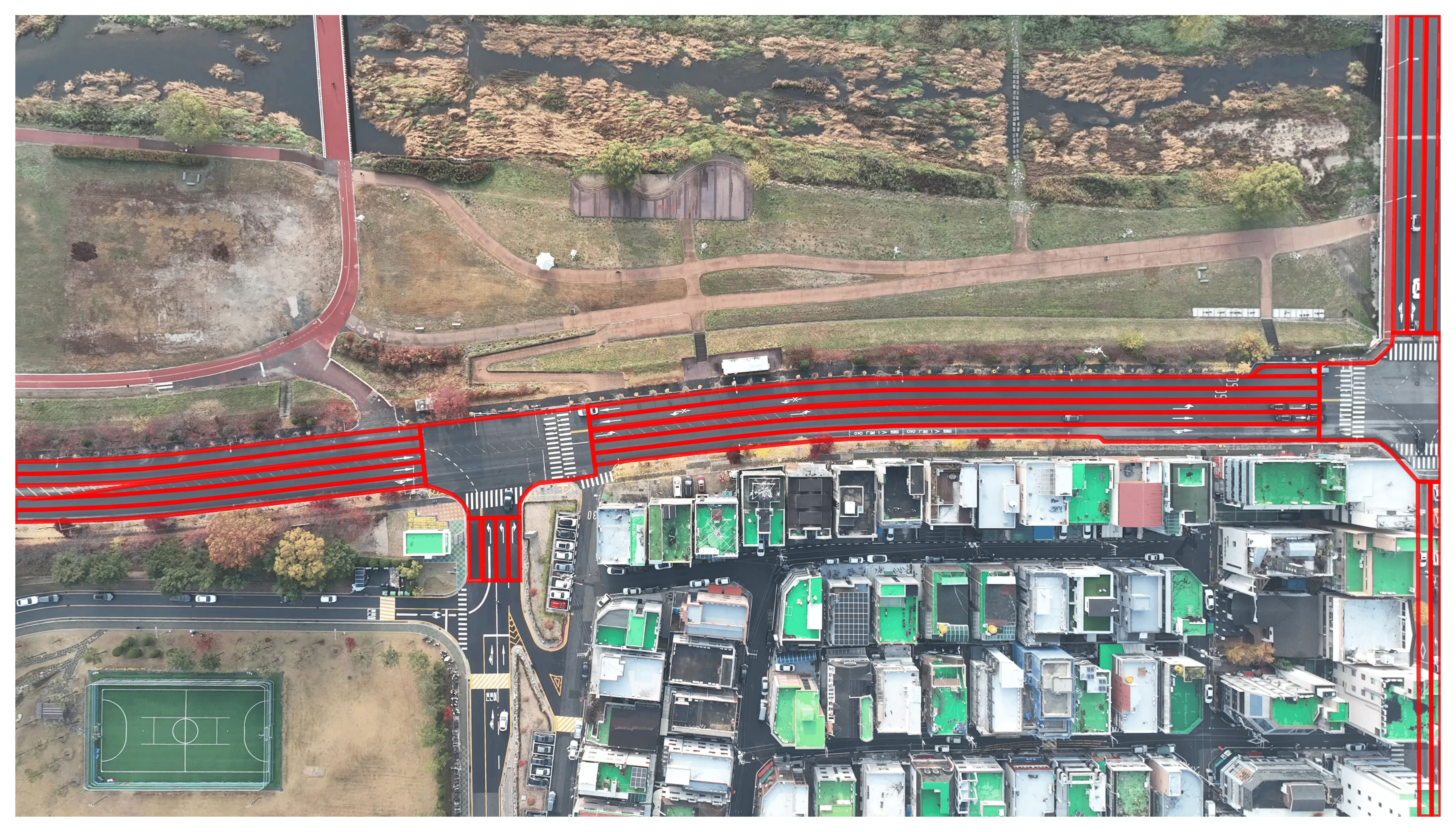}
        \caption{Site D}
        \label{subfig:site-D}
    \end{subfigure}    
    \hfill
    \begin{subfigure}[b]{0.32\textwidth}
        \includegraphics[width=\textwidth]{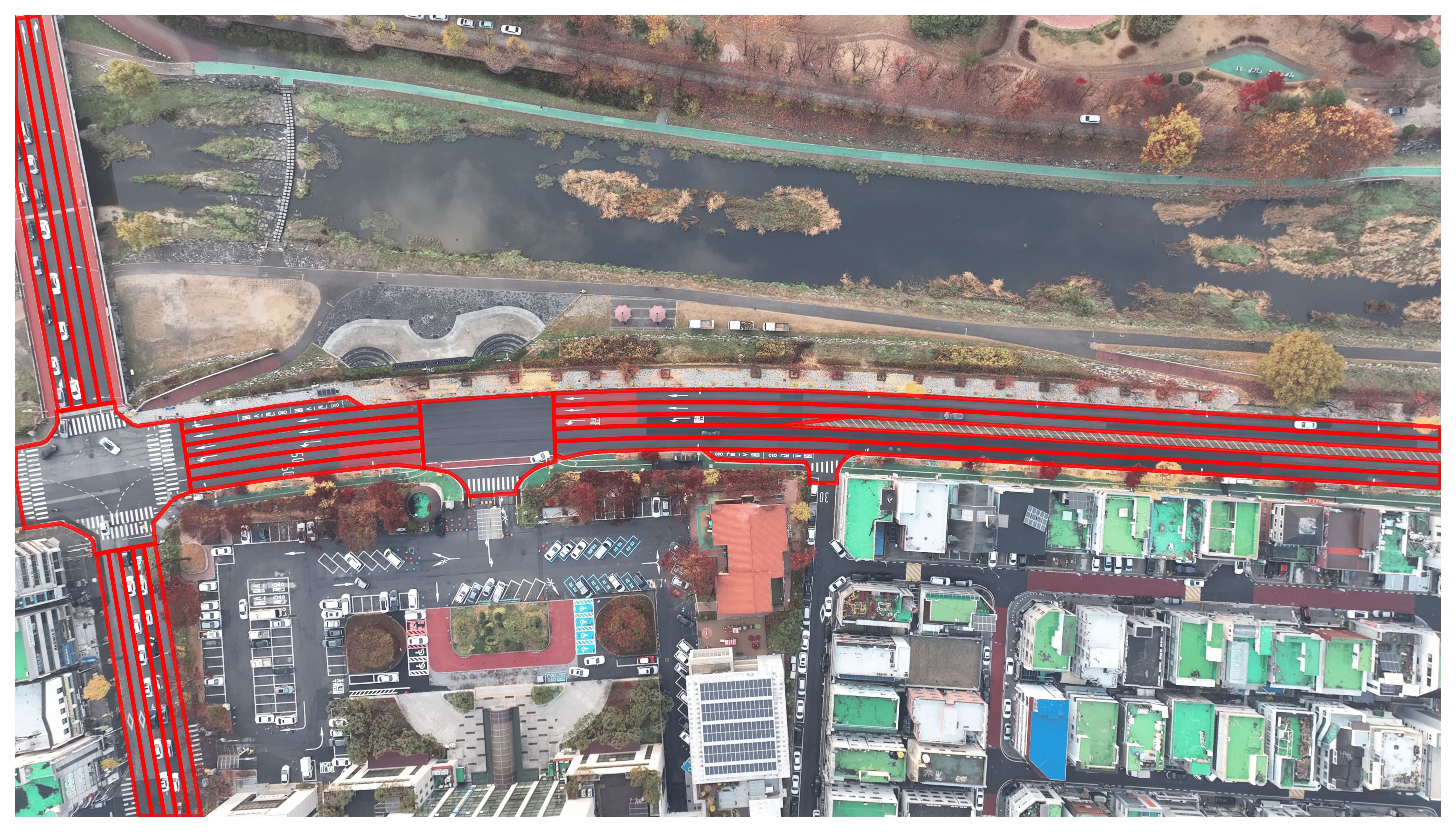}
        \caption{Site E}
        \label{subfig:site-E}
    \end{subfigure}
    \hfill
    \begin{subfigure}[b]{0.32\textwidth}
        \includegraphics[width=\textwidth]{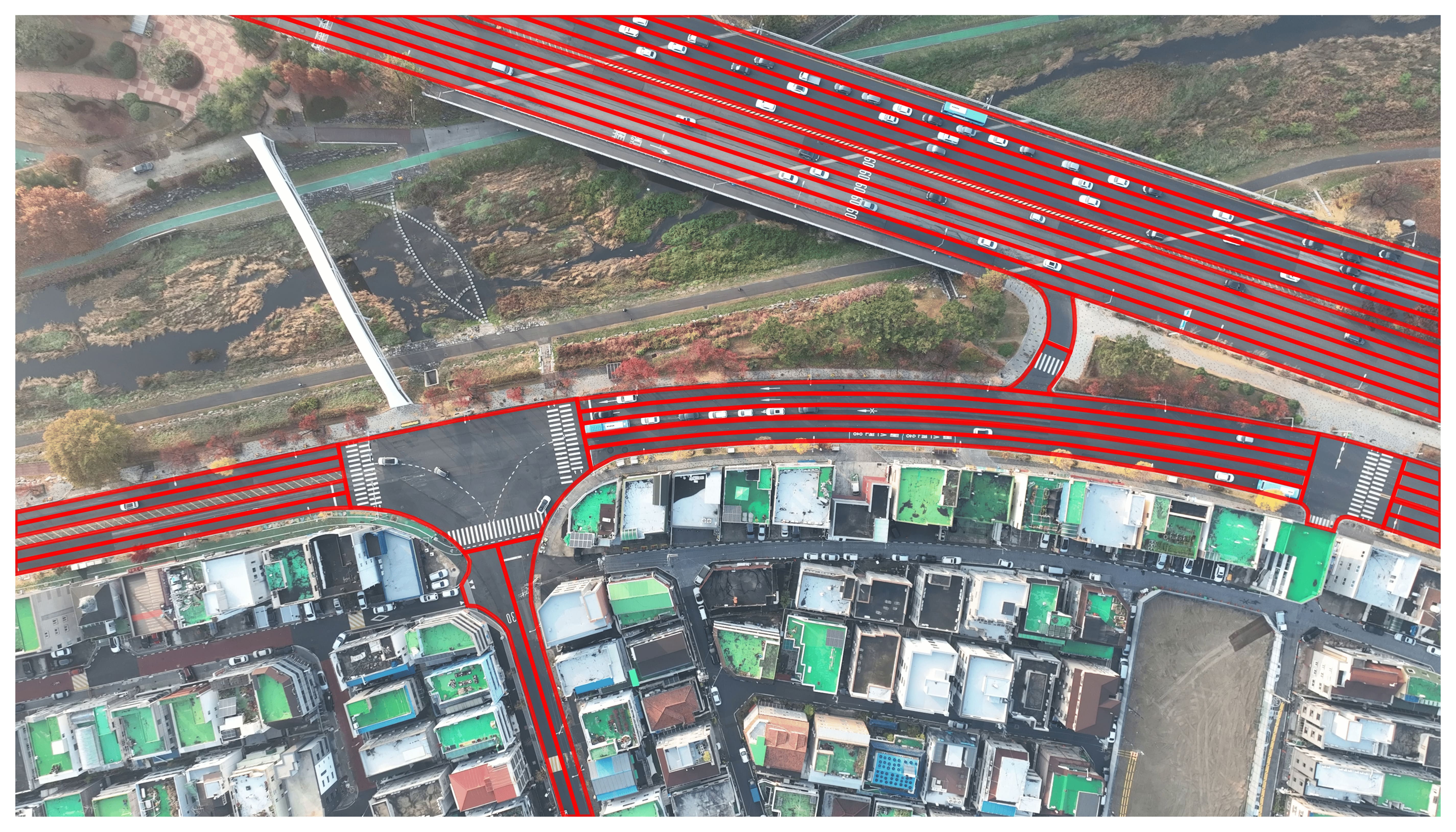}
        \caption{Site F}
        \label{subfig:site-F}
    \end{subfigure} 
    \hfill
    \begin{subfigure}[b]{0.32\textwidth}
        \includegraphics[width=\textwidth]{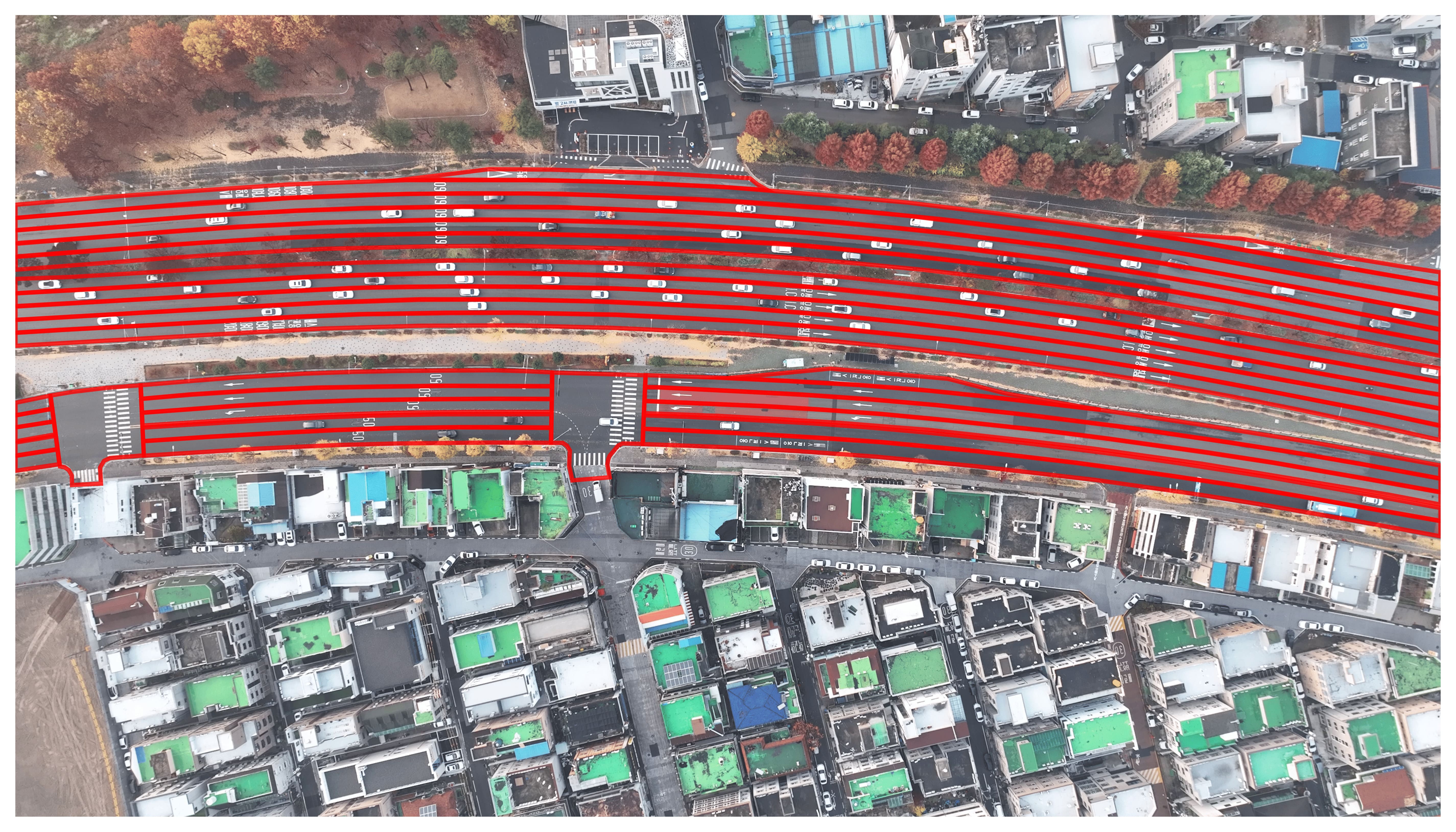}
        \caption{Site G}
        \label{subfig:site-G}
    \end{subfigure} 
    \hfill
    \begin{subfigure}[b]{0.32\textwidth}
        \includegraphics[width=\textwidth]{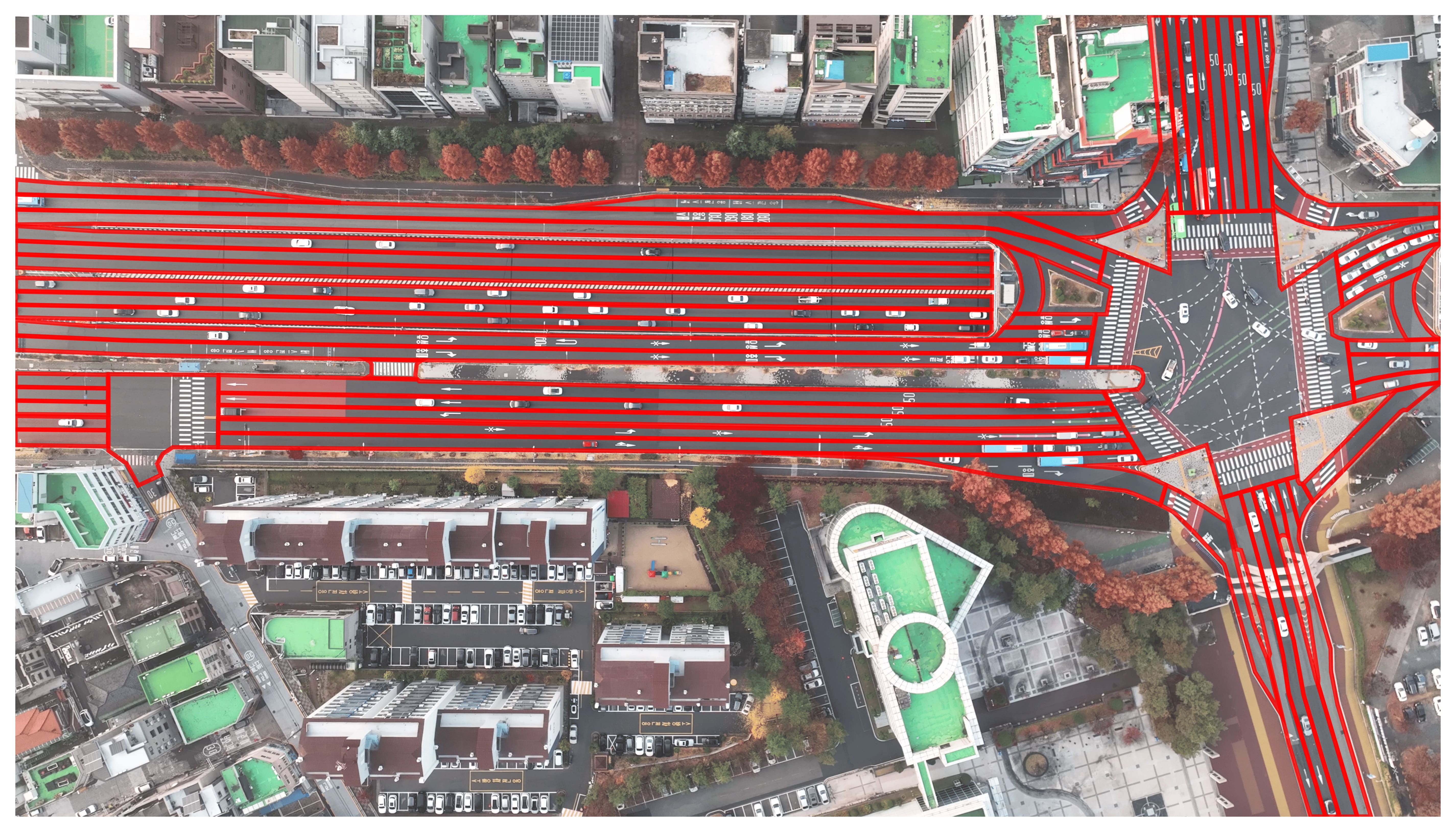}
        \caption{Site H}
        \label{subfig:site-H}
    \end{subfigure}
    \hfill
    \begin{subfigure}[b]{0.32\textwidth}
        \includegraphics[width=\textwidth]{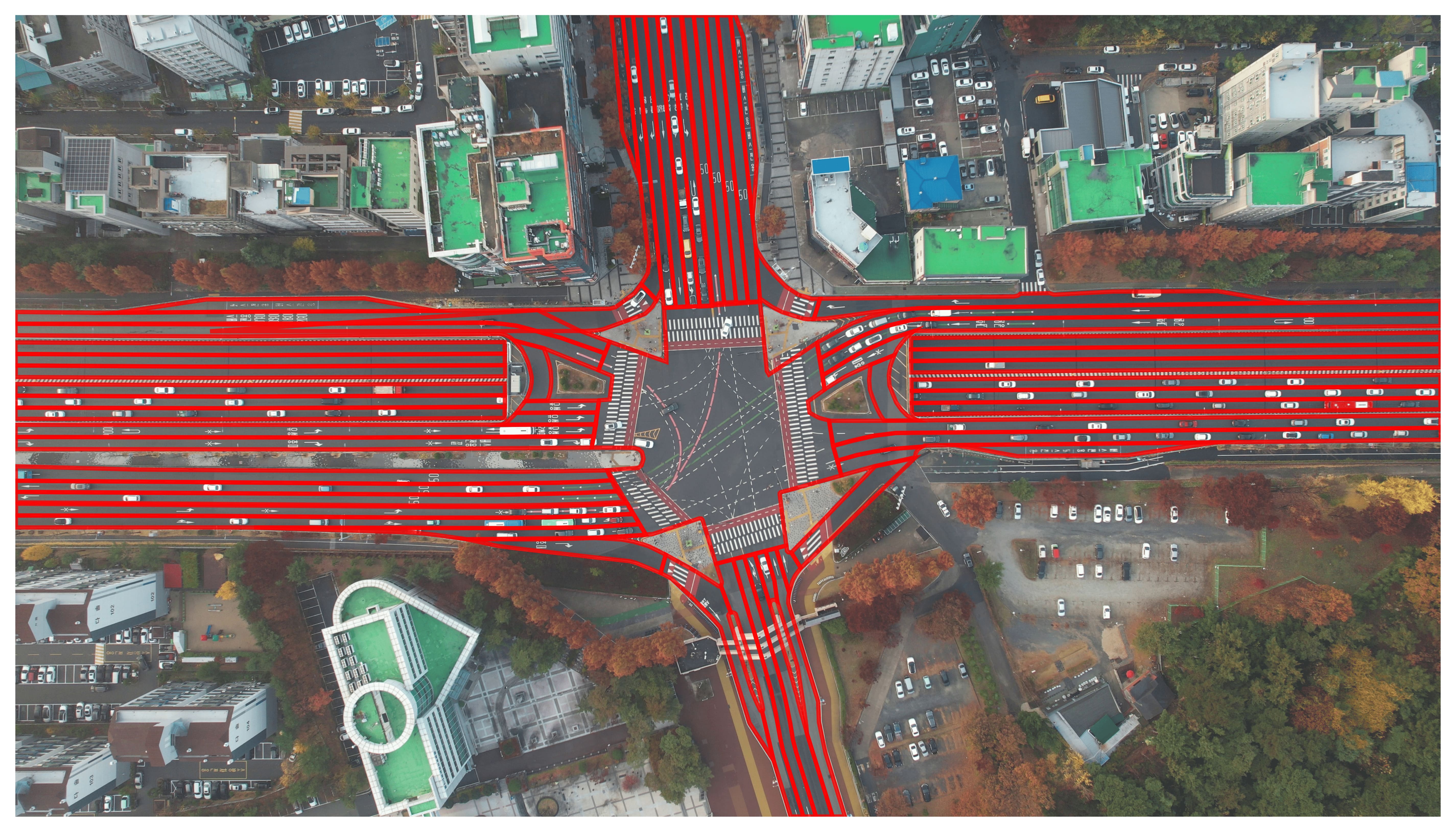}
        \caption{Site I}
        \label{subfig:site-I}
    \end{subfigure}
    \caption{Road RoIs for each observation site: (a) Site A, (b) Site B, (c) Site C, (d) Site D, (e) Site E, (f) Site F, (g) Site G, (h) Site H, and (i) Site I.}
    \label{fig:roi}
\end{figure}

As the drone footage includes areas irrelevant to this study’s objectives—such as alleyways and parking lots—in addition to major roads, RoIs are designated for each experimental site after frame stabilization. Moreover, lane-specific RoIs are defined by assigning unique codes to individual lanes, enabling the identification of lane information for each vehicle. The finalized RoIs and lane-specific RoIs are illustrated in Figure~\ref{fig:roi} and are provided along with the dataset in polygon format, accompanied by the respective lane RoI codes.

\subsection{Traffic object detection and tracking}
In this section, we describe the strategies for training the detection and tracking models along with their evaluation outcomes. For object detection and tracking tasks, YOLOv11 \citep{yolo11_ultralytics} and ByteTracker \citep{zhang2022bytetrack} are employed, respectively. The models identify vehicles from drone-derived videos and track their trajectories.

\begin{figure}[!ht]
    \centering    
    \begin{subfigure}[b]{0.44\textwidth}
        \includegraphics[width=\textwidth]{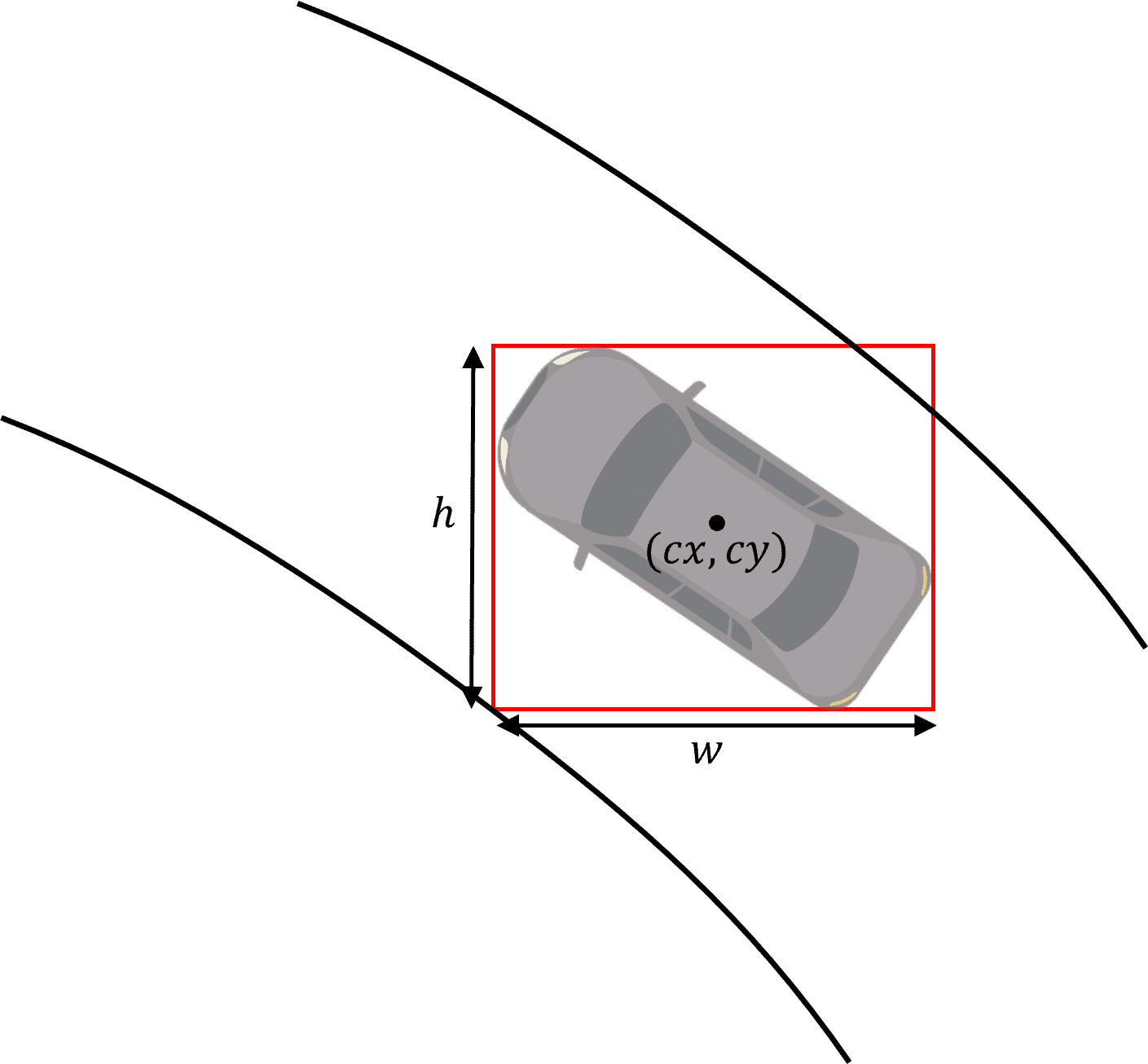}
        \caption{Axis-Aligned Bounding Box}
        \label{subfig:AABB}
    \end{subfigure}
    \begin{subfigure}[b]{0.44\textwidth}
        \includegraphics[width=\textwidth]{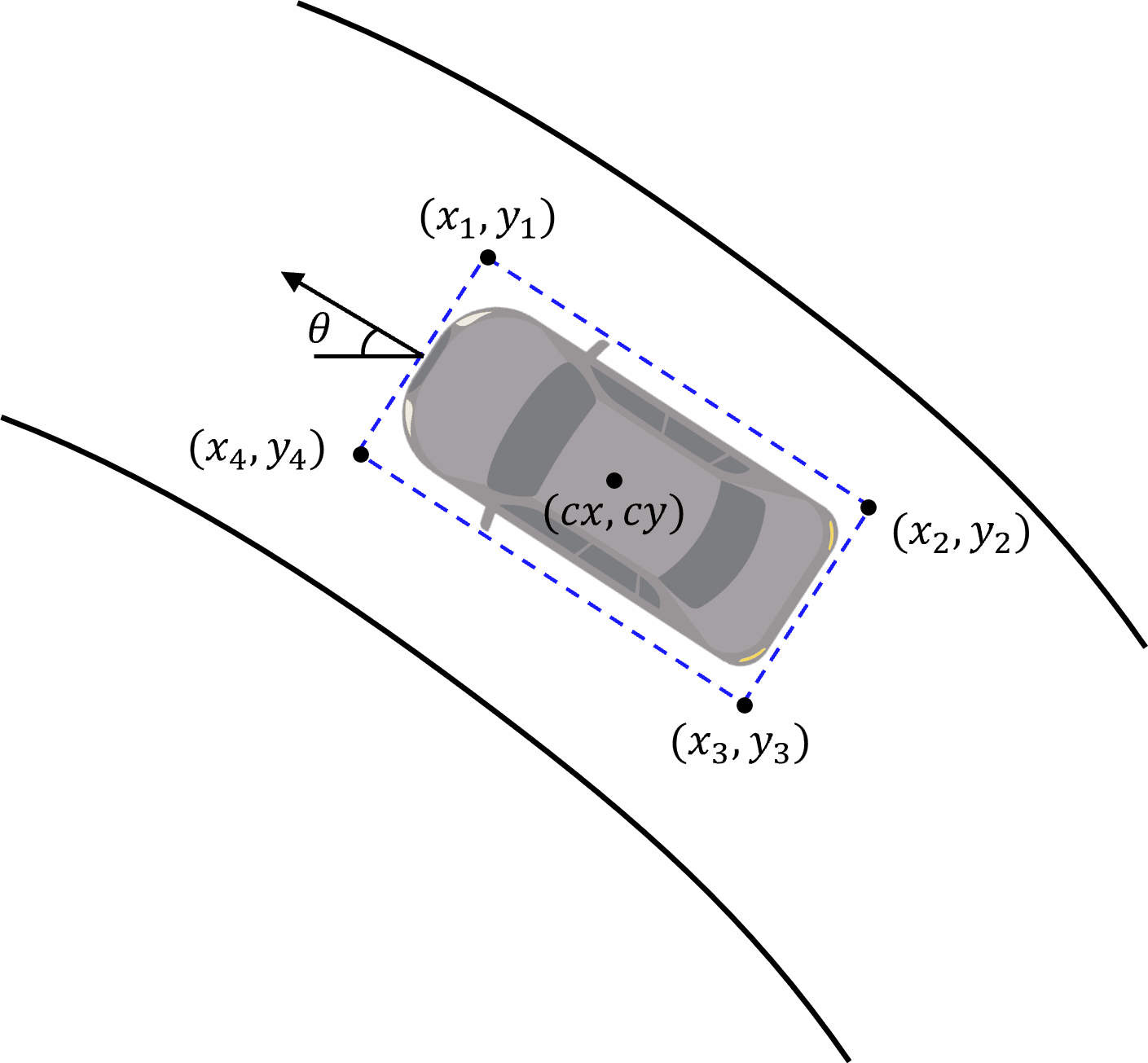}
        \caption{Oriented bounding box}
        \label{subfig:OBB}
    \end{subfigure}    
    \caption{Comparison between (a) AABB and (b) OBB for vehicle detection.}
    \label{fig:obb}
\end{figure}

In particular, OBBs are utilized instead of conventional AABB. Unlike AABB, which represent objects by rectangles aligned with image axes, OBB defines the rotation angle \(\theta\), as well as the object's center coordinates \((cx, cy)\), width \(w\), height \(h\). This approach allows for precise characterization of vehicle positions and orientations across diverse road geometries, thereby facilitating subsequent tracking tasks. Figure~\ref{fig:obb} illustrates the distinction between these two bounding-box methods. While AABBs, widely adopted in previous studies \citep{fonod_2025_13828408}, often encompass excessive background areas when objects appear rotated, polygon-based OBBs accurately capture object orientation and minimize extraneous space, thereby significantly improving position accuracy, especially in complex road environments such as curves or intersections.

\subsection{Model evaluation and data quality validation}
In this section, we assess the performance of the object detection model and the quality of extracted data.

\subsubsection{Model training phase}
\paragraph{Data preparation}
In this study, specifically YOLOv11m model was employed for object detection, targeting three classes of vehicles: cars, buses, and trucks. To achieve precise object detection and facilitate directional tracking, all instances were manually annotated in polygon format using OBB. Specifically, each object was delineated by precisely defining four corner points, accurately capturing its orientation. The final dataset consists of 2,301 video frames with 292,570 annotated object instances in polygon format. The dataset was partitioned into training, validation, and test sets, containing about 233,944, 29,478, and 29,148 instances, respectively. This data split ensures diverse exposure during training while preserving a distinct validation set for performance evaluation.

\paragraph{Hyperparameter settings}
Optimized hyperparameters were selected to achieve effective detection performance. The batch size is set to 4, the learning rate to 0.01, and the IoU threshold to 0.7. The training was performed over 500 epochs, employing early stopping to prevent overfitting. 

\subsubsection{Evaluation metrics}
For evaluating object detection performance, standard metrics including the confusion matrix, Precision-Recall (P-R) curve, mean Average Precision at a 50\% IoU threshold (mAP@50), and mean Average Precision across IoU thresholds from 50\% to 95\% (mAP@50-95) are employed. These metrics comprehensively quantify detection accuracy and localization precision, effectively capturing the overall performance of the model. 

Trajectory extraction performance is evaluated through statistical sampling across observation sites. Fragmented trajectories are identified from the sampled data, and the fragmentation rate ($\hat{p}$) is calculated along with its 95\% confidence interval (CI\(_{0.95}\)) using a binomial proportion estimation:
\begin{equation}\label{eq:ci}
\hat{p}_i = \frac{\text{Number of fragmented trajectories}}{\text{Total sampled trajectories}}, \quad 
\text{CI}_{0.95} = \hat{p} \pm 1.96 \sqrt{\frac{\hat{p}(1 - \hat{p})}{n}}
\end{equation}

The overall trajectory extraction accuracy is then computed as \((1 - \hat{p})\). This evaluation metric ensures a statistically reliable estimate of trajectory extraction performance in our DRIFT open dataset.

\subsubsection{Results}
\begin{figure}[!ht]
    \centering
    \includegraphics[width=\textwidth]{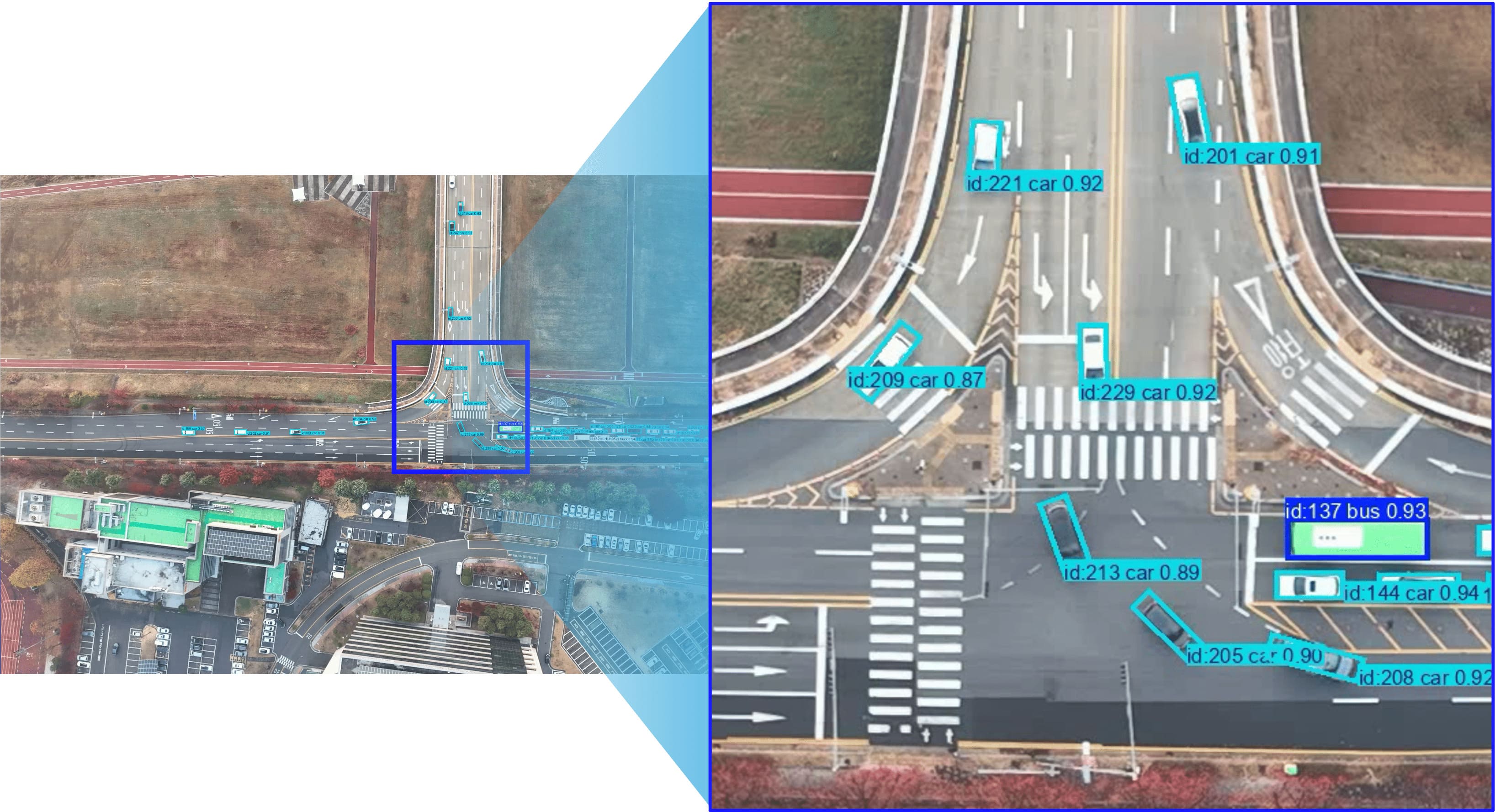}
    \caption{OBB-represented results in the object detection model.}
    \label{fig:od-result}
\end{figure}

\begin{figure}[!ht]
    \centering
    \begin{subfigure}[b]{0.55\textwidth}
        \includegraphics[width=\textwidth]{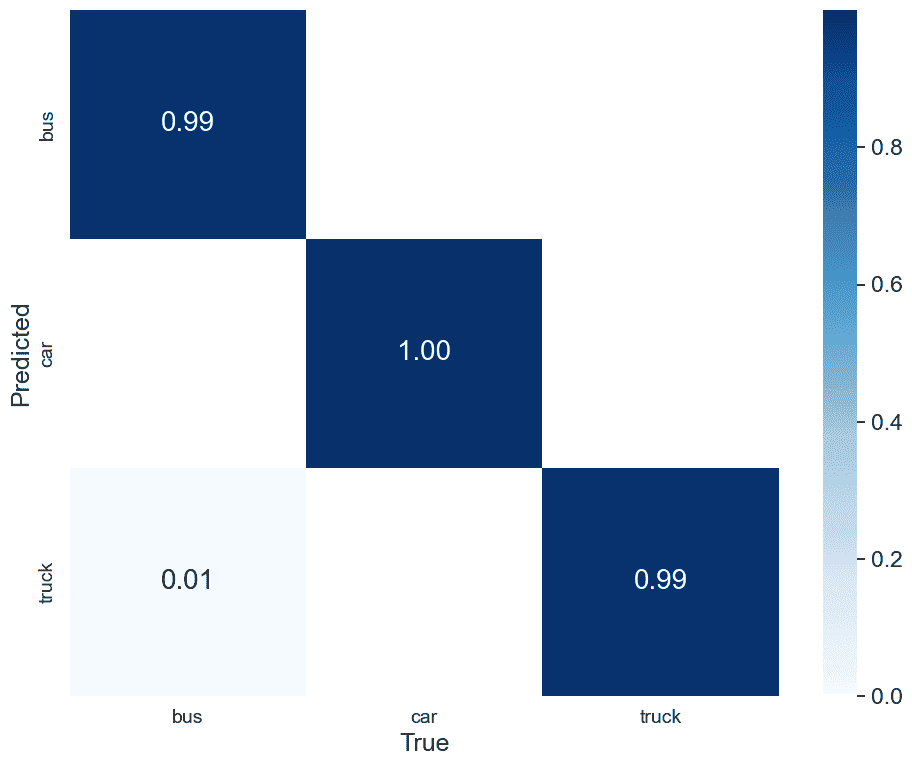}
        \caption{Confusion matrix}
        \label{subfig:confusion matrix}
    \end{subfigure}
    \hfill
    \begin{subfigure}[b]{0.43\textwidth}
        \includegraphics[width=\textwidth]{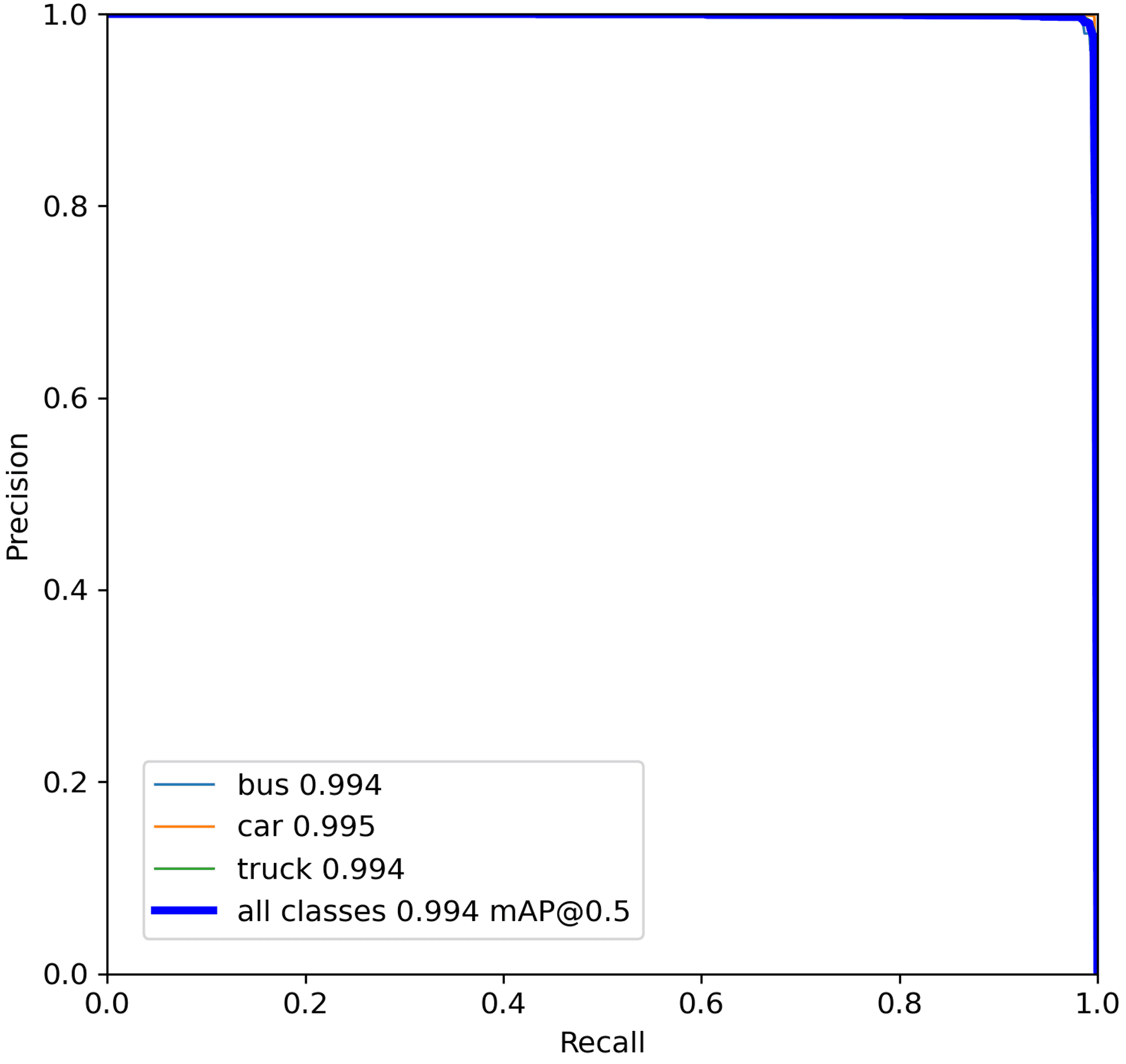}
        \caption{Precision-Recall (P-R) curve}
        \label{subfig:pr-curve}
    \end{subfigure}
    \caption{The results of traffic object detection.}
    \label{fig:od-pr-curve}
\end{figure}

\paragraph{Result of object detection}
Table~\ref{tab:evaluation_results} summarizes the detection results for each object class, demonstrating high mAP@50 and mAP@50-95 scores, thereby validating model robustness. The confusion matrix in Figure~\ref{subfig:confusion matrix} illustrates the model's classification performance, highlighting accurate predictions for each category. An analysis of the Precision-Recall (P-R) curves in Figure~\ref{subfig:pr-curve} illustrate the balanced detection capability across varying confidence thresholds. Qualitatively, Figure~\ref{fig:od-result} visualizes detection outcomes, clearly displaying vehicles detected using OBB annotations, each provided with unique IDs, rotation angles, and confidence scores

\begin{table}[h]
    \centering
    \caption{Evaluation results for object detection performance}
    \label{tab:evaluation_results}
    \renewcommand{\arraystretch}{1.2} 
    \setlength{\tabcolsep}{5pt} 
    \begin{tabular}{lcccccc}
        \hline
        \textbf{Class} & \textbf{\# of frames} & \textbf{\# of instances} & \textbf{Precision} & \textbf{Recall} & \textbf{mAP@50} & \textbf{mAP@50-95} \\
        \hline
        All   & 230 & 29,478 & 0.992 & 0.992 & 0.994 & 0.962 \\
        Bus   & 178 & 534   & 0.987 & 0.987 & 0.994 & 0.963 \\
        Car   & 230 & 27,374 & 0.998 & 0.997 & 0.995 & 0.964 \\
        Truck & 226 & 1,570  & 0.992 & 0.992 & 0.994 & 0.960 \\
        \hline
    \end{tabular}
\end{table}

\paragraph{Result of trajectory extraction}
Overall, the total number of trajectories in all sites in the DRIFT open dataset is about 81,699, and the numbers of trajectories in each site are depicted in Table~\ref{tab:data-stats}. 

\begin{figure}[!ht]
    \centering
    \begin{subfigure}[b]{0.32\textwidth}
        \includegraphics[width=\textwidth]{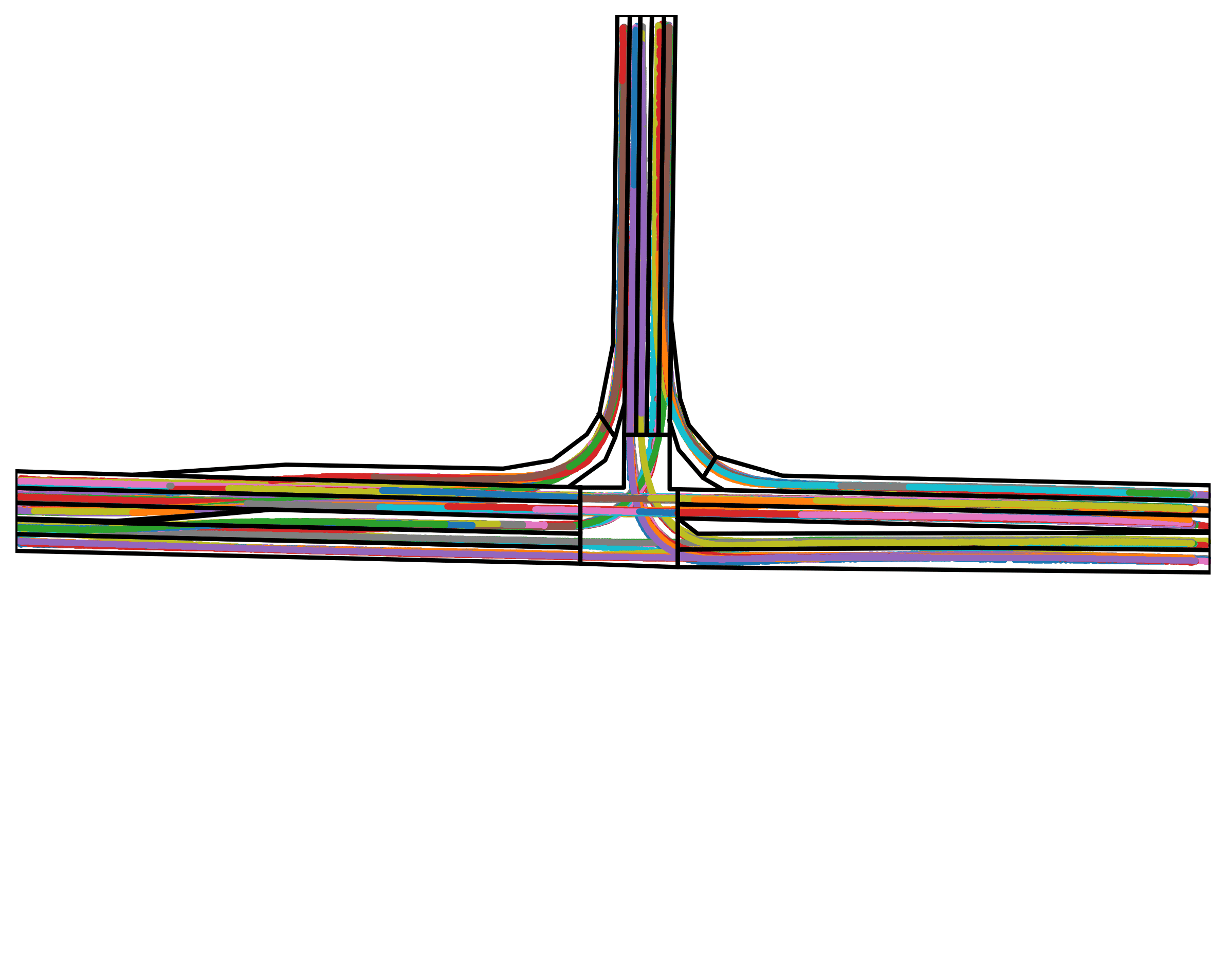}
        \caption{Site A}
        \label{subfig:traj-A}
    \end{subfigure}
    \hfill
    \begin{subfigure}[b]{0.32\textwidth}
        \includegraphics[width=\textwidth]{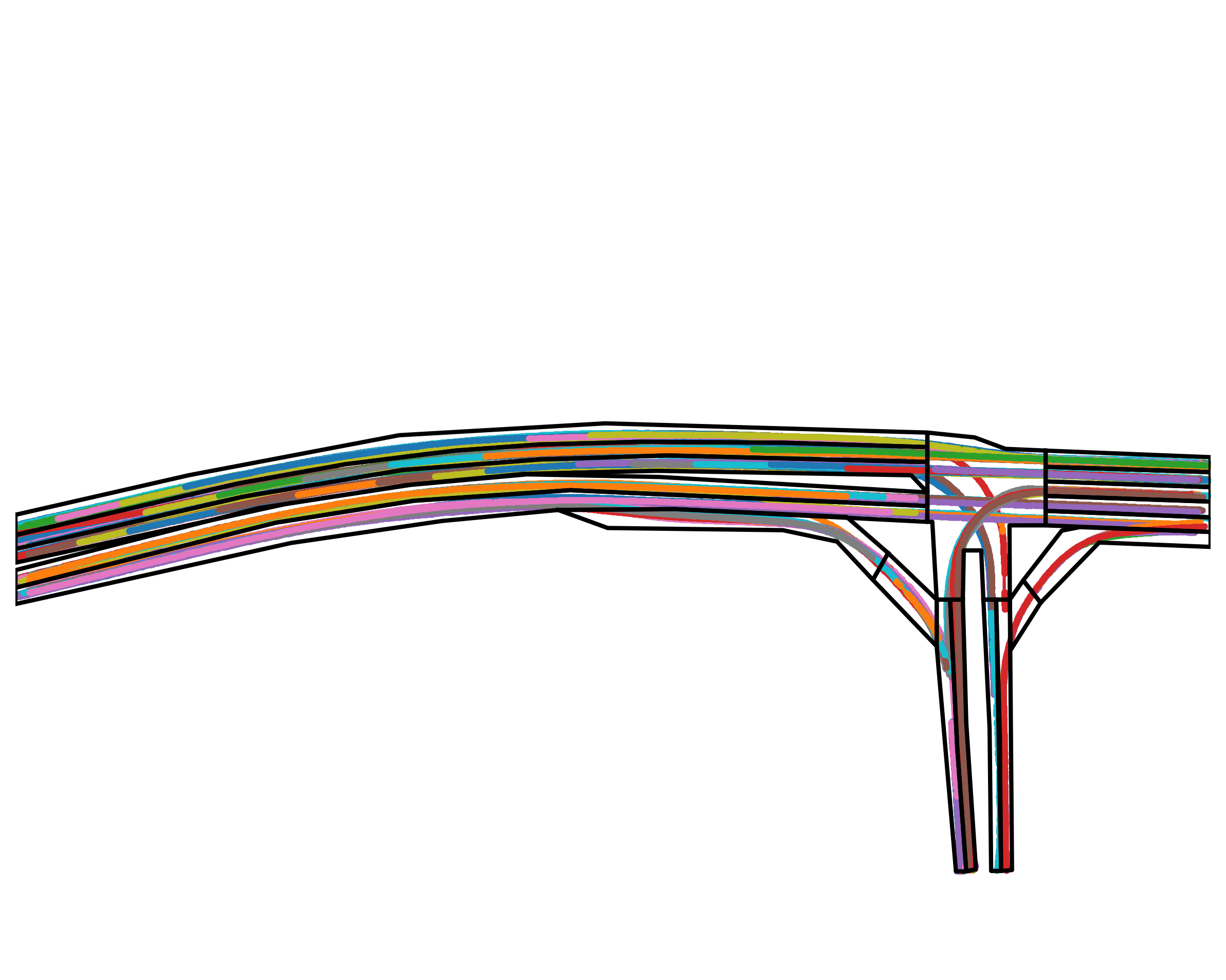}
        \caption{Site B}
        \label{subfig:traj-B}
    \end{subfigure}
    \hfill
    \begin{subfigure}[b]{0.32\textwidth}
        \includegraphics[width=\textwidth]{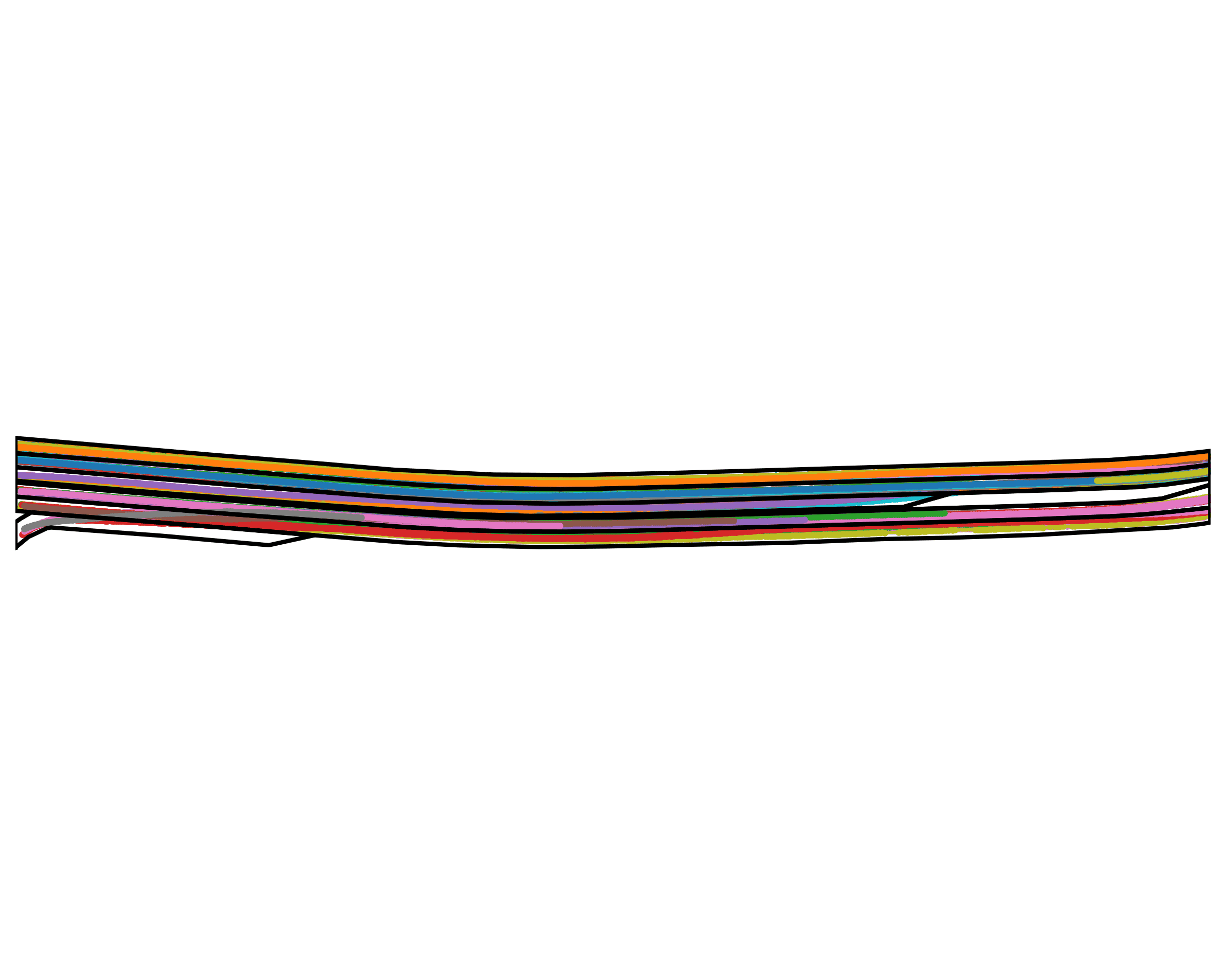}
        \caption{Site C}
        \label{subfig:traj-C}
    \end{subfigure} 
    \hfill
    \begin{subfigure}[b]{0.32\textwidth}
        \includegraphics[width=\textwidth]{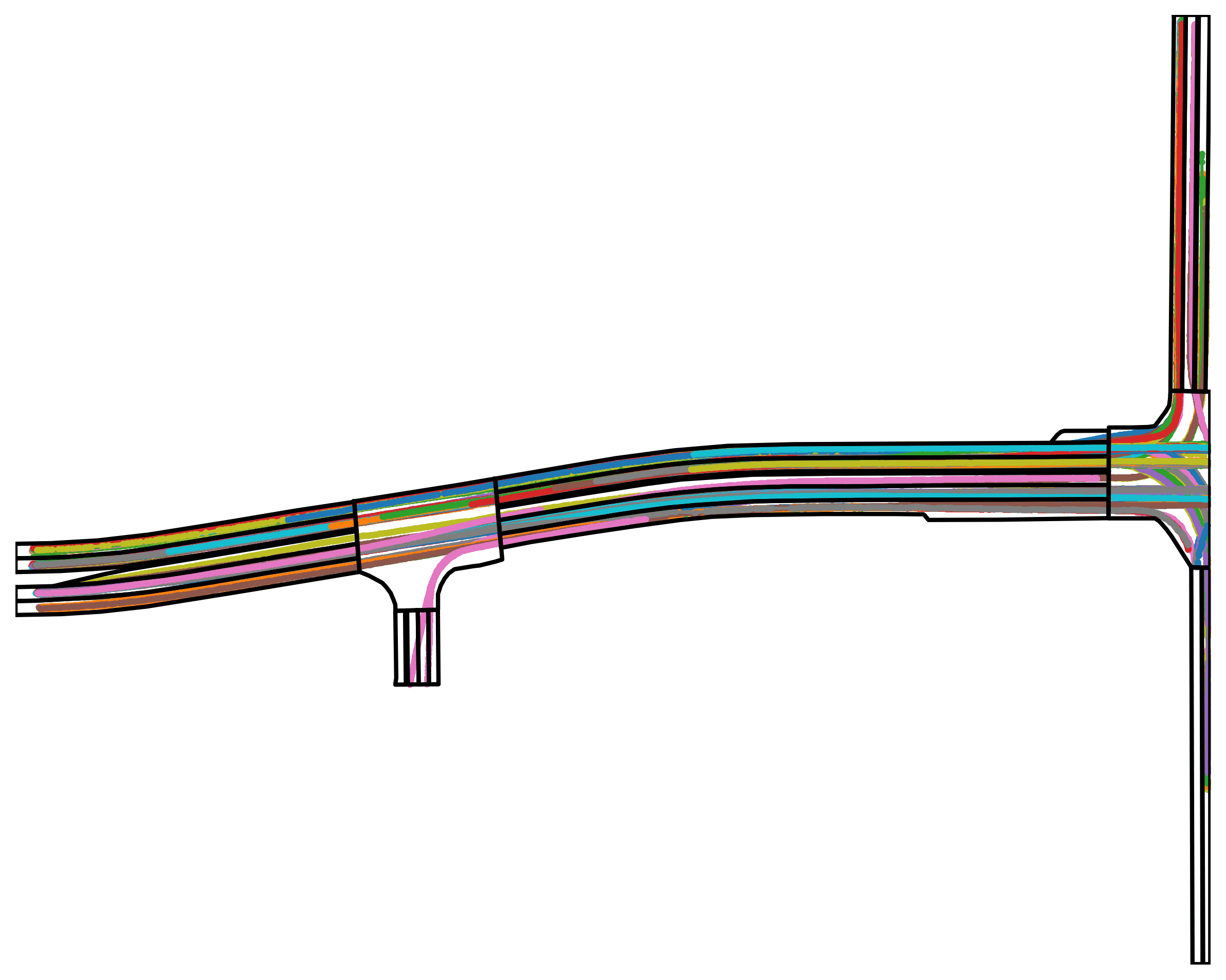}
        \caption{Site D}
        \label{subfig:traj-D}
    \end{subfigure}    
    \hfill
    \begin{subfigure}[b]{0.32\textwidth}
        \includegraphics[width=\textwidth]{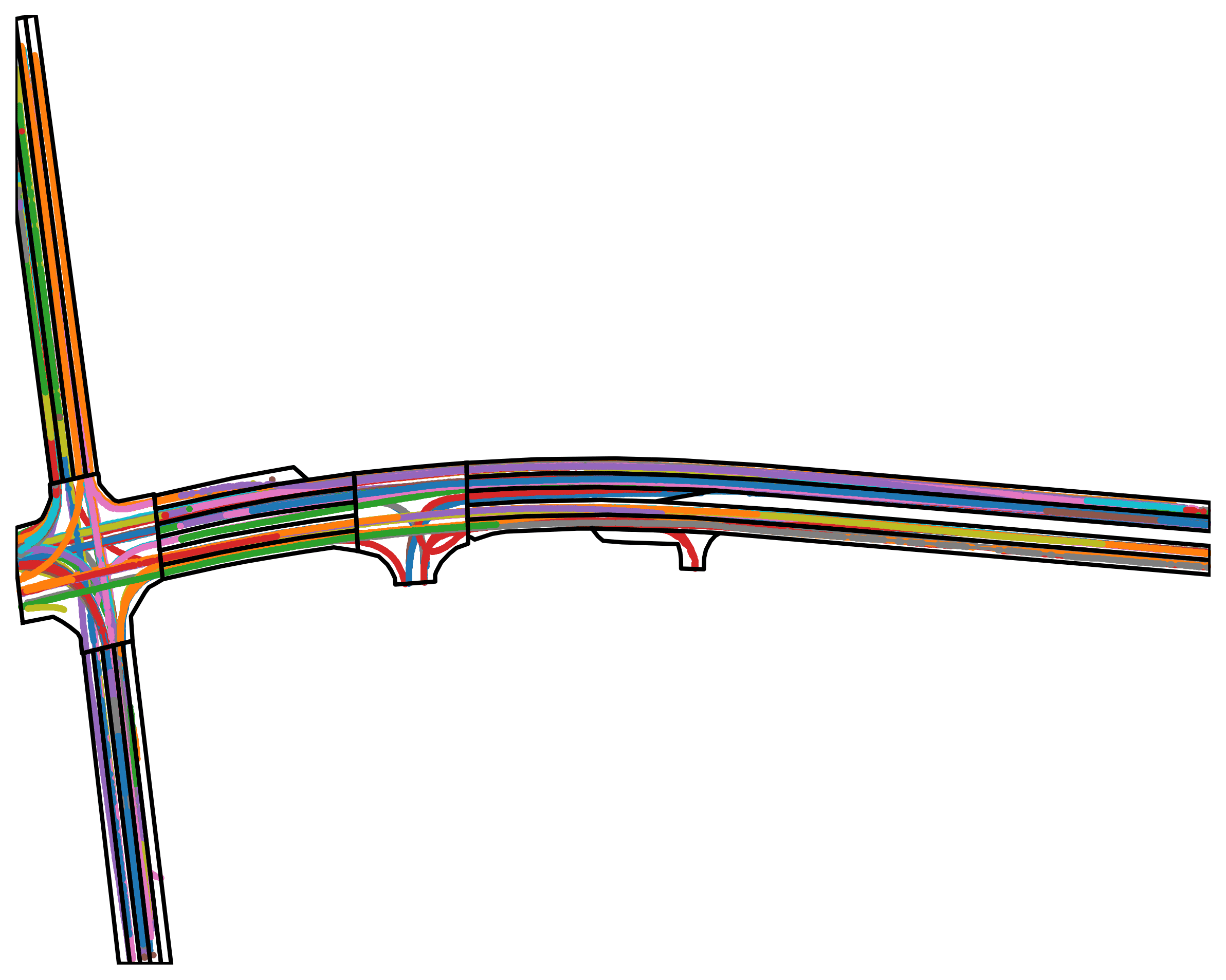}
        \caption{Site E}
        \label{subfig:traj-E}
    \end{subfigure}
    \hfill
    \begin{subfigure}[b]{0.32\textwidth}
        \includegraphics[width=\textwidth]{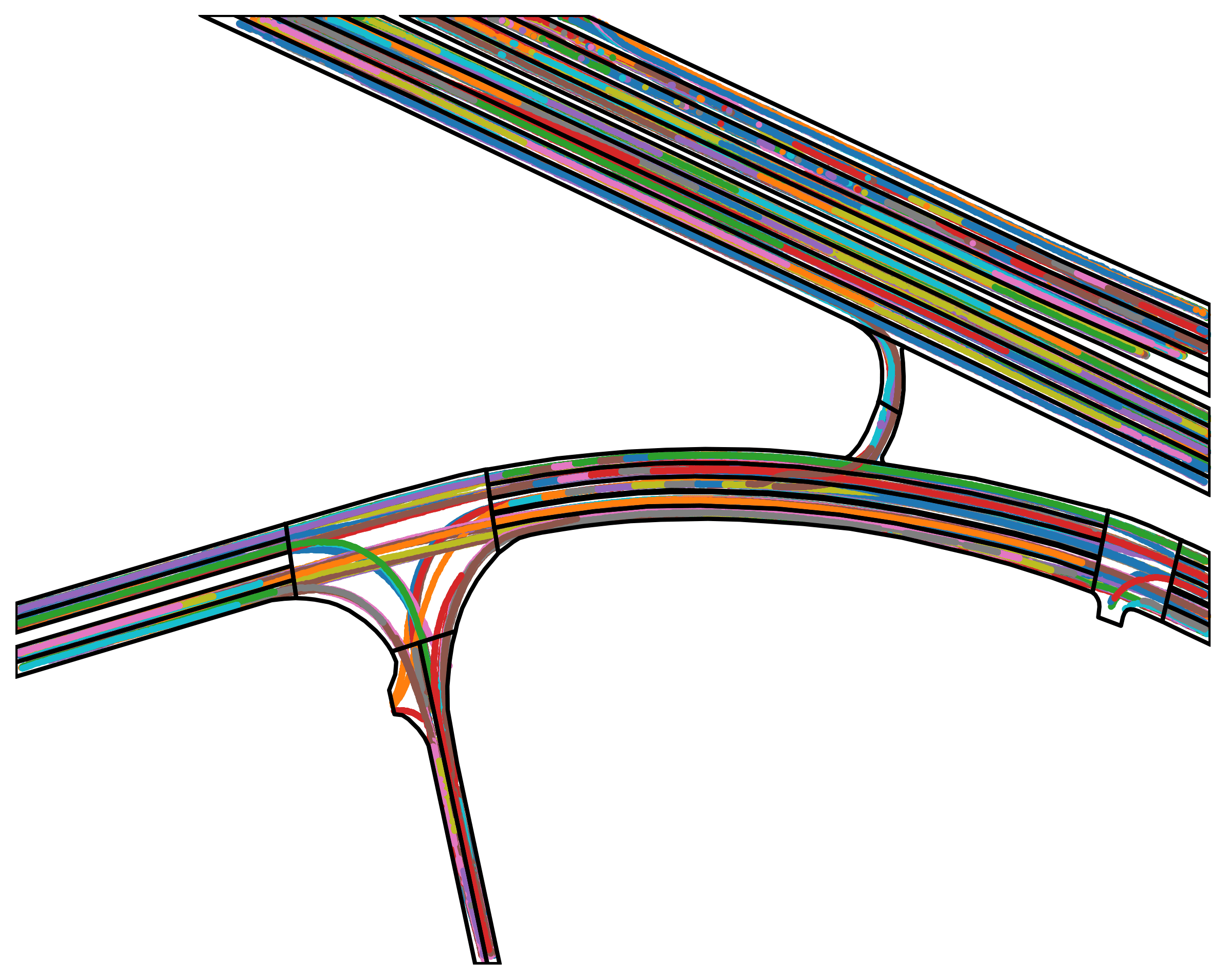}
        \caption{Site F}
        \label{subfig:traj-F}
    \end{subfigure} 
    \hfill
    \begin{subfigure}[b]{0.32\textwidth}
        \includegraphics[width=\textwidth]{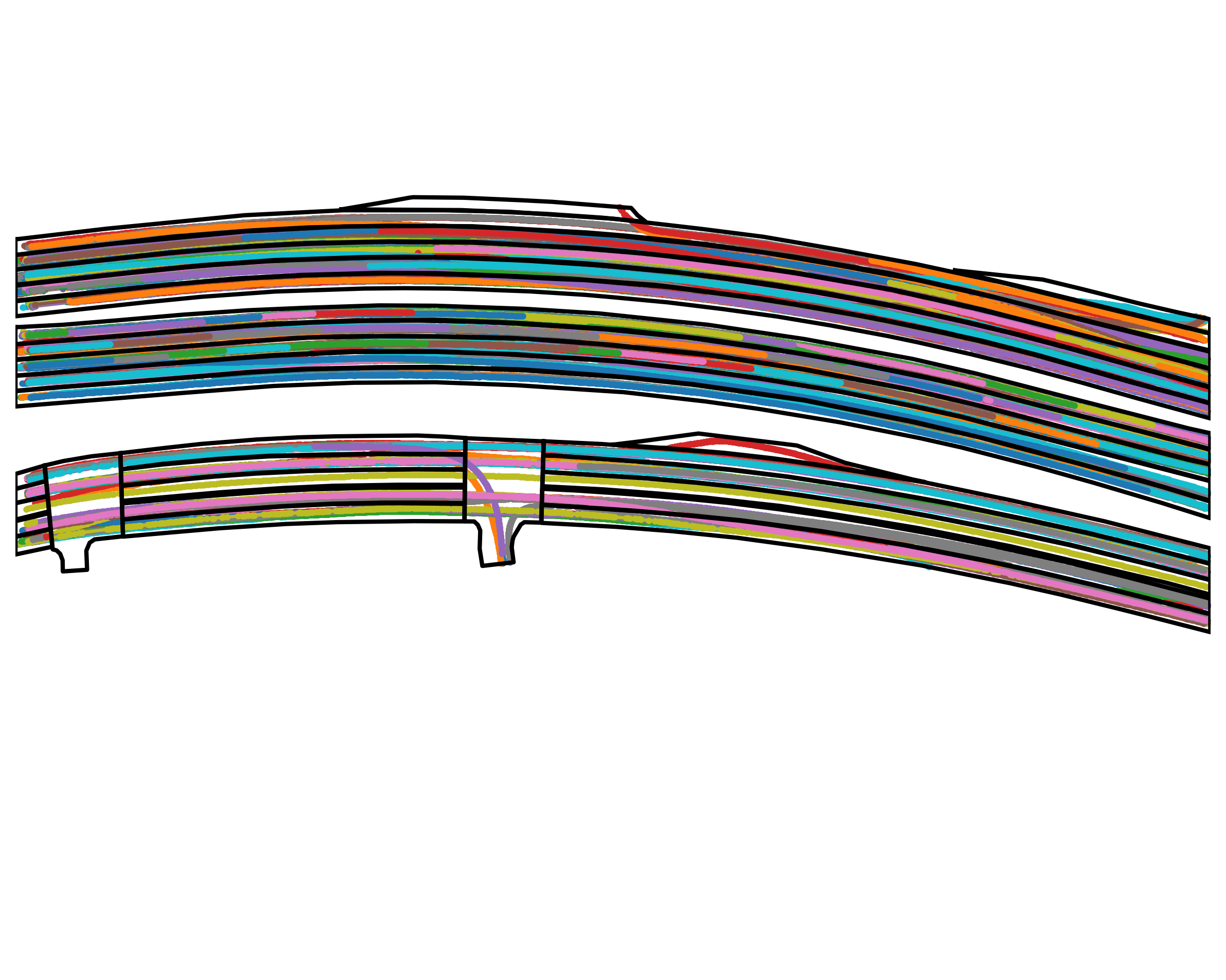}
        \caption{Site G}
        \label{subfig:traj-G}
    \end{subfigure} 
    \hfill
    \begin{subfigure}[b]{0.32\textwidth}
        \includegraphics[width=\textwidth]{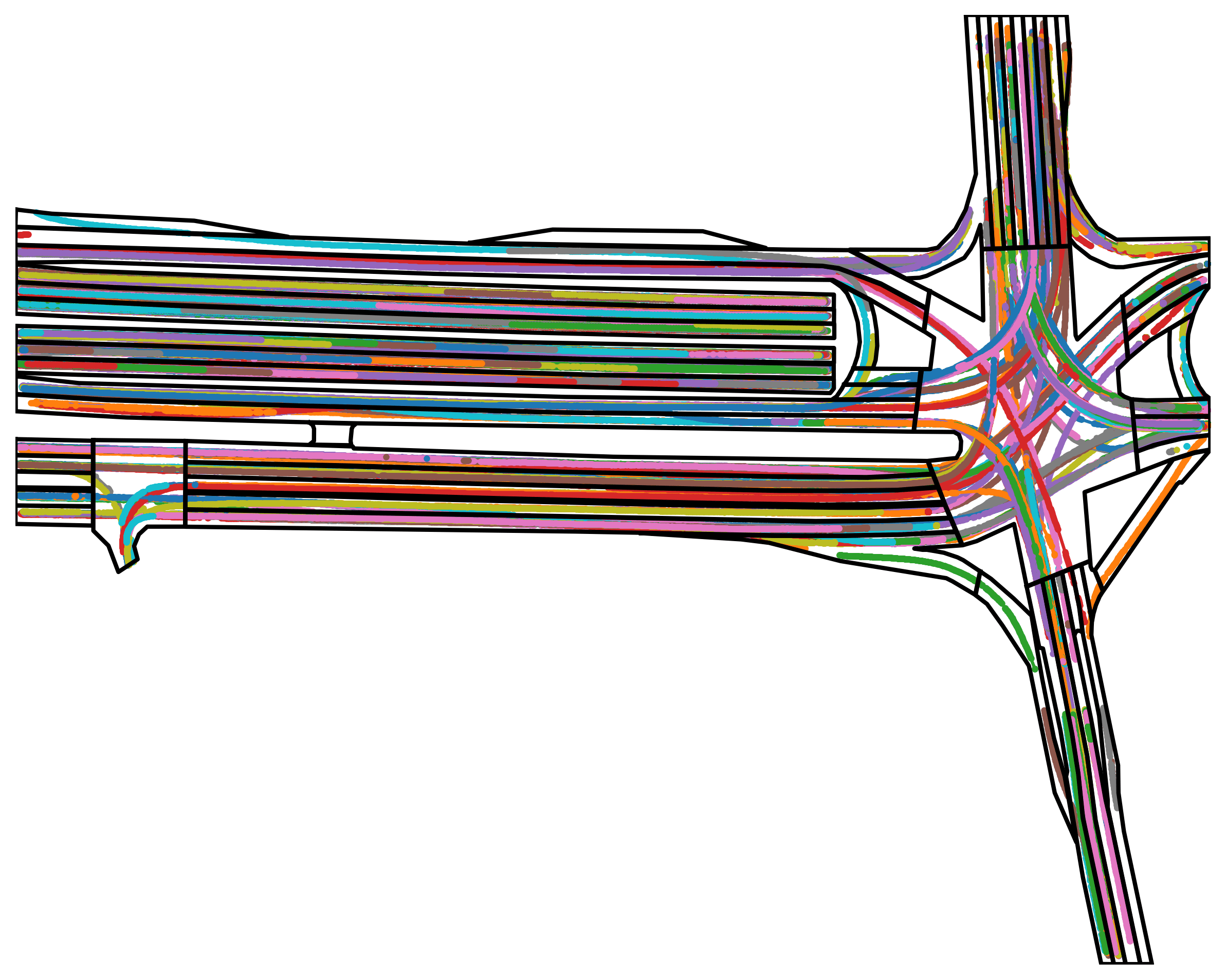}
        \caption{Site H}
        \label{subfig:traj-H}
    \end{subfigure}
    \hfill
    \begin{subfigure}[b]{0.32\textwidth}
        \includegraphics[width=\textwidth]{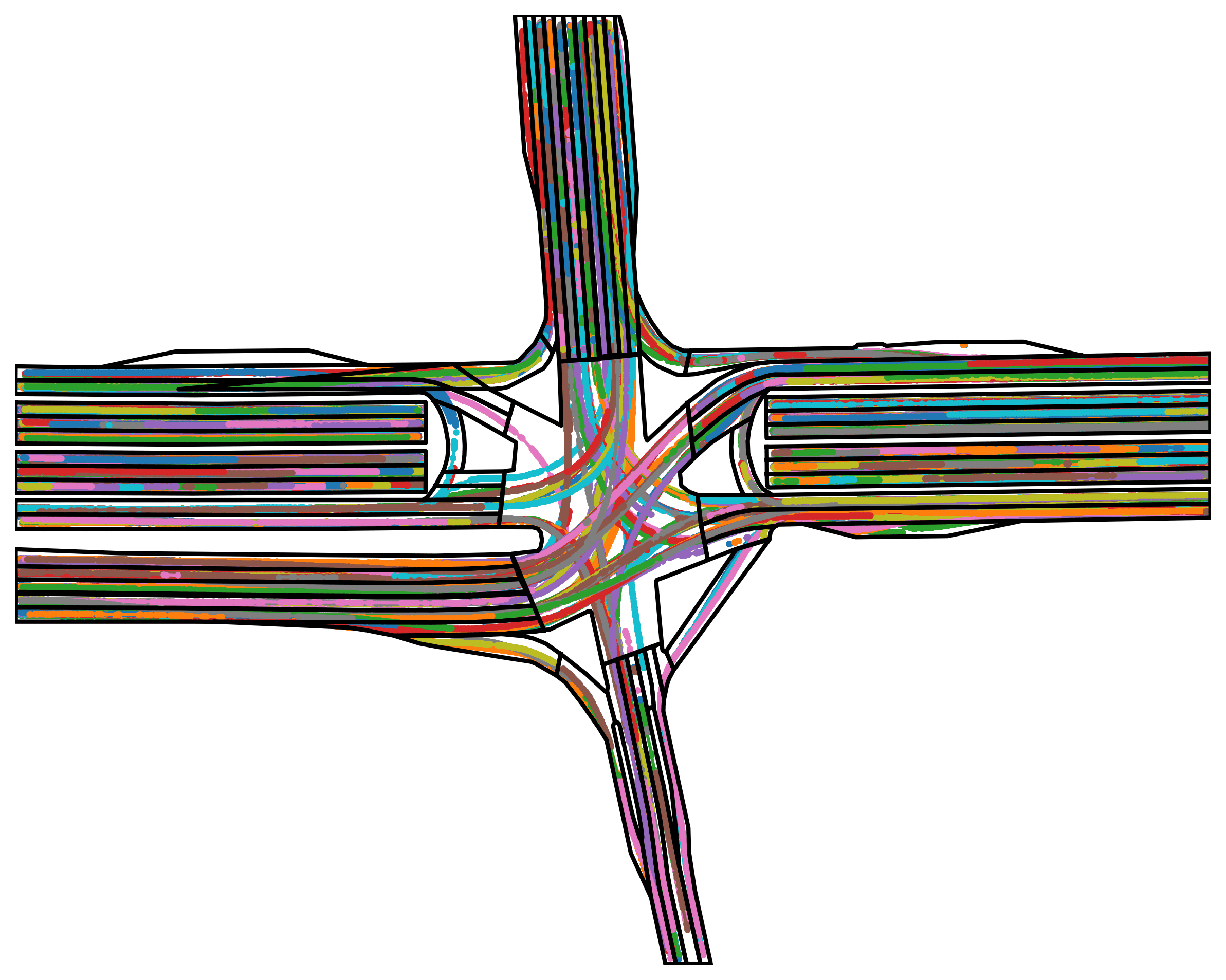}
        \caption{Site I}
        \label{subfig:traj-I}
    \end{subfigure}
    \caption{Extracted trajectories in each test site (represented only RoI lanes)}
    \label{fig:trajectory}
\end{figure}

Trajectory extraction performance was evaluated through statistical sampling, where approximately \(5\%\) of trajectories were randomly selected, resulting in a total of \(3,080\) sampled trajectories. Fragmented trajectories were counted, and the fragmentation rate (\(\hat{p}\)) was estimated along with its CI\(_{0.95}\) as shown in Equation~\ref{eq:ci}. Applying this calculation, the overall trajectory extraction accuracy was estimated to be approximately \(96.98\%\), demonstrating robust and statistically reliable trajectory continuity in drone-derived data. Qualitatively, Figure~\ref{fig:trajectory} illustrates vehicle trajectories derived from drone imagery, confirming trajectory consistency within the designated RoIs.

\subsection{Specification of DRIFT open dataset}
This section provides an overview of the DRIFT open dataset, detailing the traffic-related information it contains and associated analytical tools. The DRIFT dataset primarily offers comprehensive vehicle trajectory data, from which additional vehicle states, such as speed and acceleration, can be derived. These computed vehicle kinematics facilitate analyses across various traffic scales, including microscopic analyses (e.g., lane-changing behaviors, vehicle-to-vehicle time-to-collision), mesoscopic analyses (e.g., flow-density diagrams, time-space diagrams), and macroscopic analyses (e.g., aggregated speed profiles). Further potential applications and significance of the dataset are discussed in detail in subsequent sections.

\subsection{Data format and accessibility}
The primary dataset is structured on a per-frame basis, with detected vehicle data stored in tabular format, as shown in Table~\ref{tab:data-format}. In addition, tools are provided to transform frame-based data into vehicle-specific trajectories, facilitating trajectory-based analyses. Such structured datasets form the foundation for subsequent processing and enable the extraction of more nuanced traffic characteristics.

\begin{table}[!ht]
    \centering
    \caption{Structure of the extracted traffic trajectory data}
    \label{tab:data-format}
    \renewcommand{\arraystretch}{1.2}
    \begin{tabular}{lp{0.65\textwidth}} \toprule
        \textbf{Column} & \textbf{Description} \\ \midrule
        \texttt{track\_id} & Unique identifier assigned to each vehicle throughout its trajectory \\
        \texttt{frame} & Frame index in the video sequence (30 frames per second) \\
        \texttt{center\_x}, \texttt{center\_y} & Horizontal and vertical positions of the vehicle center, respectively \\
        \texttt{width} & Width of the detected vehicle \\
        \texttt{height} & Height of the detected vehicle \\
        \texttt{angle} & Orientation angle of the vehicle in radians \\
        \texttt{x1}, \texttt{y1} & Coordinates of the front-left corner of the vehicle \\
        \texttt{x2}, \texttt{y2} & Coordinates of the front-right corner of the vehicle \\
        \texttt{x3}, \texttt{y3} & Coordinates of the rear-right corner of the vehicle \\
        \texttt{x4}, \texttt{y4} & Coordinates of the rear-left corner of the vehicle \\
        \texttt{confidence} & Confidence score of the detection result (range: 0 to 1) \\
        \texttt{class\_id} & Object class label (1: bus, 2: car, 3: truck) \\
        \texttt{site} & Identifier of the observation site \\
        \texttt{lane} & Lane index where the vehicle is currently located \\
        \texttt{preceding\_id} & Identifier of the vehicle directly ahead \\
        \texttt{following\_id} & Identifier of the vehicle directly behind \\
        \bottomrule
    \end{tabular}
    \raggedright
    \footnotesize\textbf{Note}: Position- and size-related values (e.g., coordinates, width, height) are expressed in pixels.
\end{table}

Vehicle speed is computed based on positional displacement between consecutive frames, whereas lane-changing events are identified through temporal variations in assigned lane codes. TTC values are estimated using the relative speed and spatial positioning between a vehicle and its preceding vehicle. Methodologies for extracting these behavioral attributes (e.g., coordinates, speeds, distances to preceding vehicles) from video data have been extensively developed and validated in our previous studies. For detailed explanations, readers are referred to \citep{noh2022novel, noh2022analyzing, noh2021safetycube, noh2020vision, noh2021analysis, no2024novel}.

In total, the DRIFT dataset provides synchronized trajectory data consisting of approximately 81,699 trajectories across all observation sites. Specific trajectory statistics per site are summarized in Table~\ref{tab:data-stats}.

\begin{table}[h]
    \centering   
    \caption{Status of the DRIFT dataset version 1.0}
    \label{tab:data-stats}
    \renewcommand{\arraystretch}{1.2} 
    \setlength{\tabcolsep}{5pt} 
        \begin{tabular}{ccc}
            \toprule
            \textbf{Site} & \textbf{\# of traj. (avg. frames per traj.)} & \textbf{Traj. by class} \\
            \midrule
            A & 4,288 (954.92)  & Bus: 104, Car: 3,643, Truck: 541  \\
            \midrule
            B & 4,110 (898.33)  & Bus: 94, Car: 3,635, Truck: 381  \\
            \midrule
            C & 2,628 (822.34)  & Bus: 68, Car: 2,454, Truck: 106  \\
            \midrule
            D & 2,630 (1,038.90)  & Bus: 28, Car: 2,452, Truck: 150  \\
            \midrule
            E & 4,327 (804.38)  & Bus: 106, Car: 3,885, Truck: 336  \\
            \midrule
            F & 12,447 (476.43) & Bus: 397, Car: 9,412, Truck: 2,638  \\
            \midrule
            G & 10,020 (536.70) & Bus: 340, Car: 8,461, Truck: 1,219  \\
            \midrule
            H & 16,889 (458.76) & Bus: 622, Car: 14,436, Truck: 1,831 \\
            \midrule
            I & 24,360 (517.17) & Bus: 796, Car: 20,877, Truck: 2,687 \\
            \midrule
            \textbf{Total} & \textbf{81,699} & \textbf{Bus: 2,555, Car: 70,255, Truck: 10,889}\\
            \bottomrule
        \end{tabular}
\end{table}

\subsection{Data processing and analysis tools}
In addition to the dataset itself, DRIFT provides full access via GitHub to source codes, trained models, and analysis tools utilized throughout dataset preparation. The GitHub repository is organized into five directories—\textit{data, extraction, model, utils}, and \textit{vis}—with their respective roles as follows:

\begin{itemize}
\item \textit{data}: Contains datasets in CSV format, along with reference images and sample videos for each observation site, assisting users in comprehending data structure and spatial characteristics.

\item \textit{extraction}: Includes scripts for video stabilization, preprocessing, and feature extraction, ensuring the accuracy and reliability of the data generation process.

\item \textit{model}: Provides training and evaluation scripts for the YOLOv11m vehicle-detection model and the ByteTracker trajectory-extraction model specifically optimized for drone-based perspectives. Users can directly apply or customize these models according to their research objectives.

\item \textit{utils}: Supplies various utility scripts for efficient data handling, including automation scripts for converting video data into frame-by-frame images and facilitating format conversions among CSV, JSON, XML, and other data formats.

\item \textit{vis}: Offers visualization scripts designed for comprehensive analysis at micro-, meso-, and macro-scales, aiding intuitive interpretation and thorough analysis of traffic data.
\end{itemize}

The provided dataset and accompanying analytical tools are designed to facilitate effective data utilization and comprehensive analysis of traffic conditions, enabling users to flexibly manipulate data and efficiently validate models tailored to diverse research applications.


\section{DRIFT open dataset exploration: Multi-Level characteristics and application potentials}\label{chap:analysis}

This section presents several traffic analyses using the proposed DRIFT open dataset and discusses its potential for future research and applications. Our analysis framework is structured around three complementary perspectives: microscopic, mesoscopic, and macroscopic levels of traffic behavior. At the micro-level, the analysis delves into individual vehicle behaviors, specifically investigating lane change maneuvers and TTC metrics to understand dynamic interactions and safety aspects. The meso-level focuses on the performance of road segments by employing time-space diagrams and fundamental diagrams to capture traffic flow patterns, congestion levels, and the efficiency of road infrastructure. Finally, at the macro-level, the study evaluates the level of service at each intersection, synthesizing data across multiple links to provide a comprehensive overview of systemic traffic conditions and inform strategic urban mobility planning. Through this foundational analysis, we aim to offer readers a comprehensive understanding of the dataset, highlighting its utility as a valuable resource for future transportation research and practical applications.

\subsection{Micro-level traffic characteristics}
At the agent level, we analyze lane-change and car-following behaviors using individual vehicle trajectories to gain micro-level insights into traffic dynamics.

\begin{figure}[!ht]
    \centering
    \begin{subfigure}[b]{0.30\textwidth}
        \includegraphics[width=\textwidth]{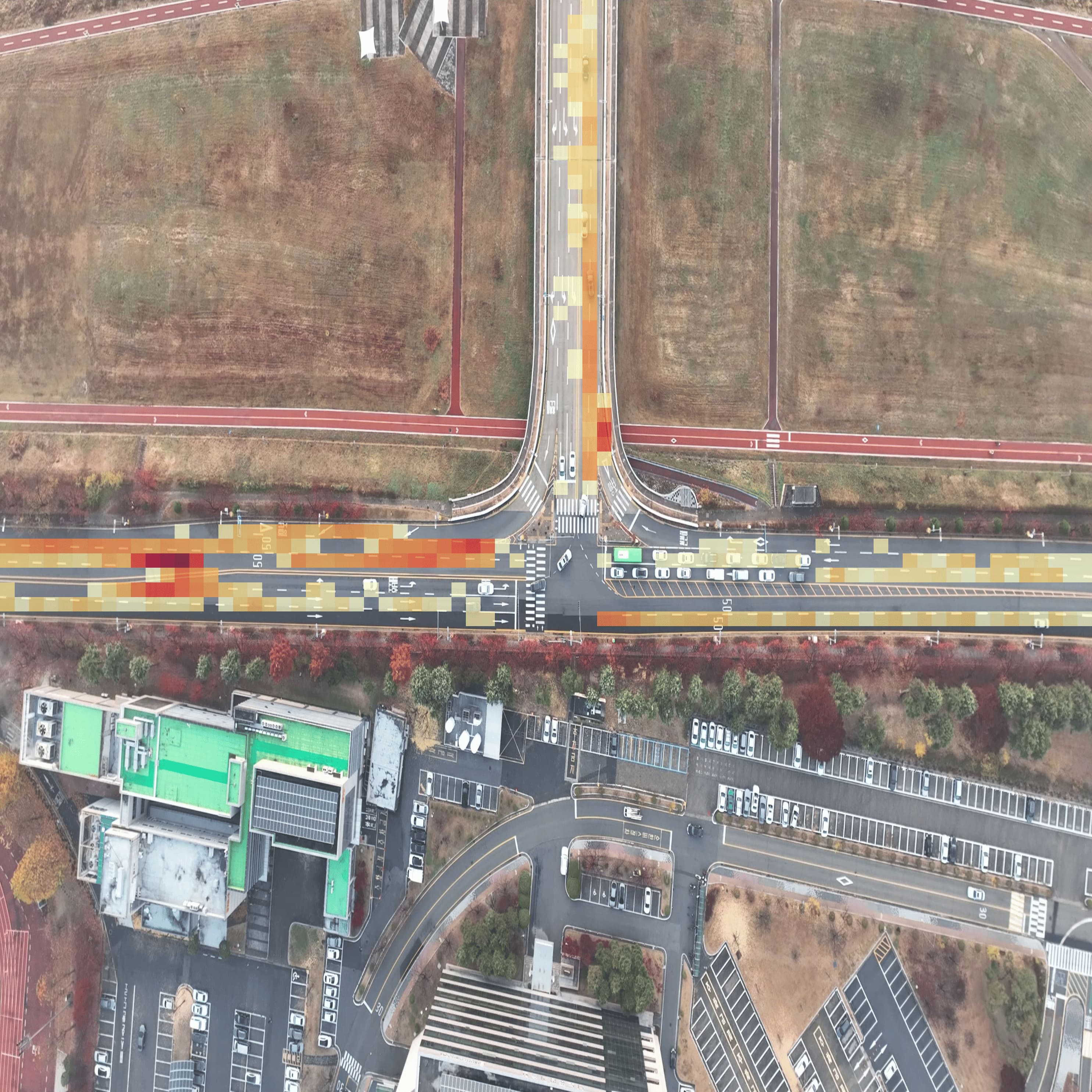}
        \caption{Site A}
        \label{subfig:LC-A}
    \end{subfigure}
    \hfill
    \begin{subfigure}[b]{0.30\textwidth}
        \includegraphics[width=\textwidth]{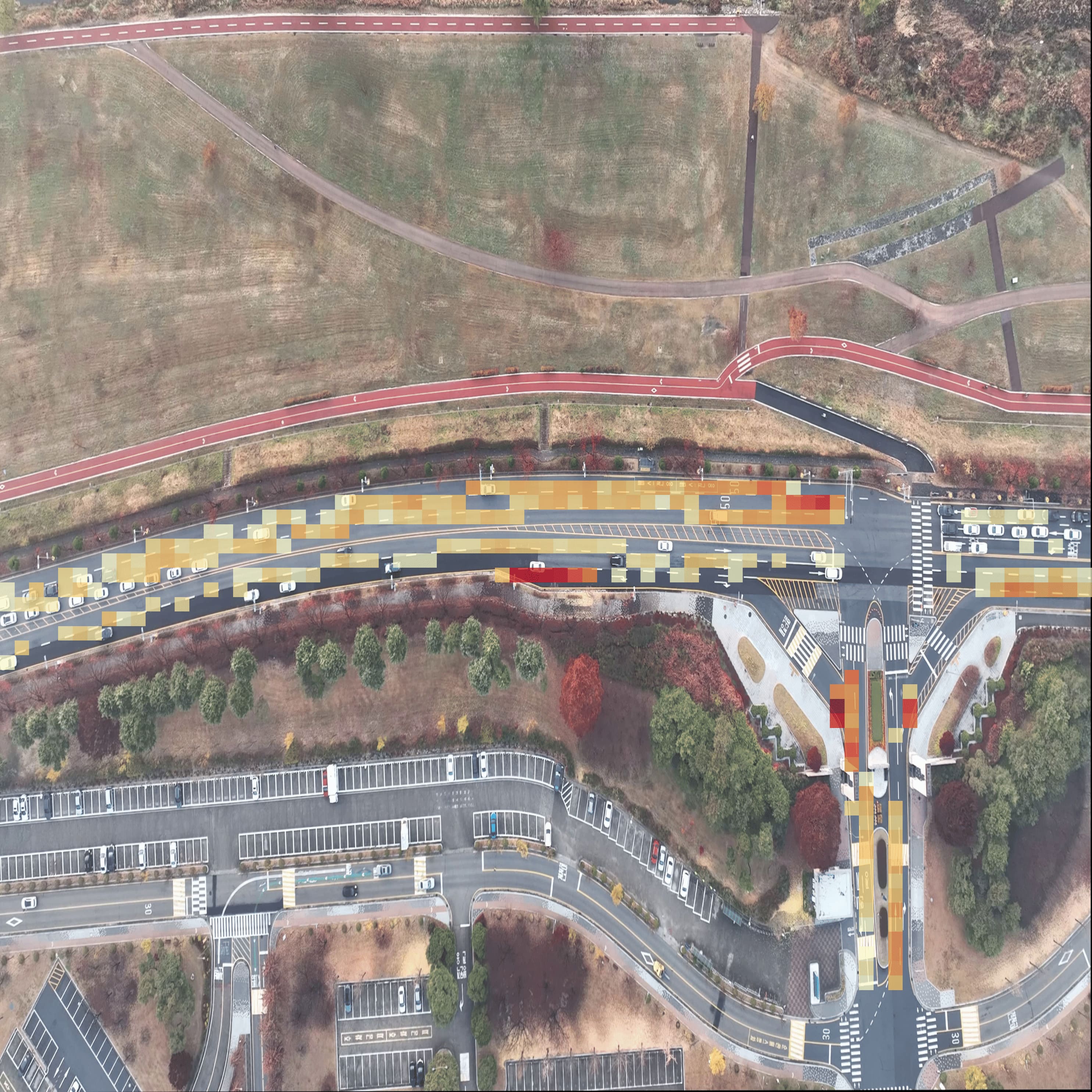}
        \caption{Site B}
        \label{subfig:LC-B}
    \end{subfigure}
    \hfill
    \begin{subfigure}[b]{0.3\textwidth}
        \includegraphics[width=\textwidth]{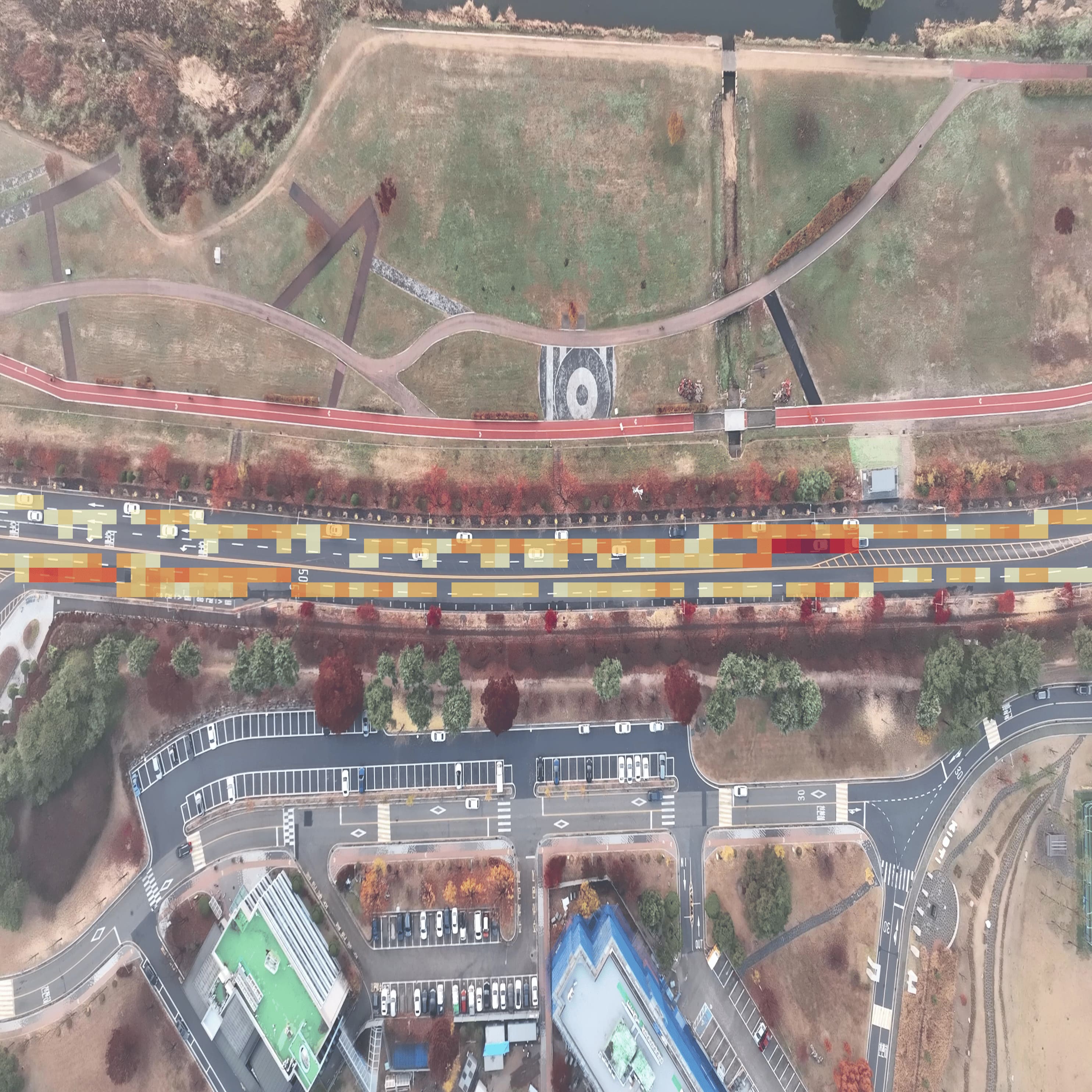}
        \caption{Site C}
        \label{subfig:LC-C}
    \end{subfigure} 
    \hfill
    \begin{subfigure}[b]{0.3\textwidth}
        \includegraphics[width=\textwidth]{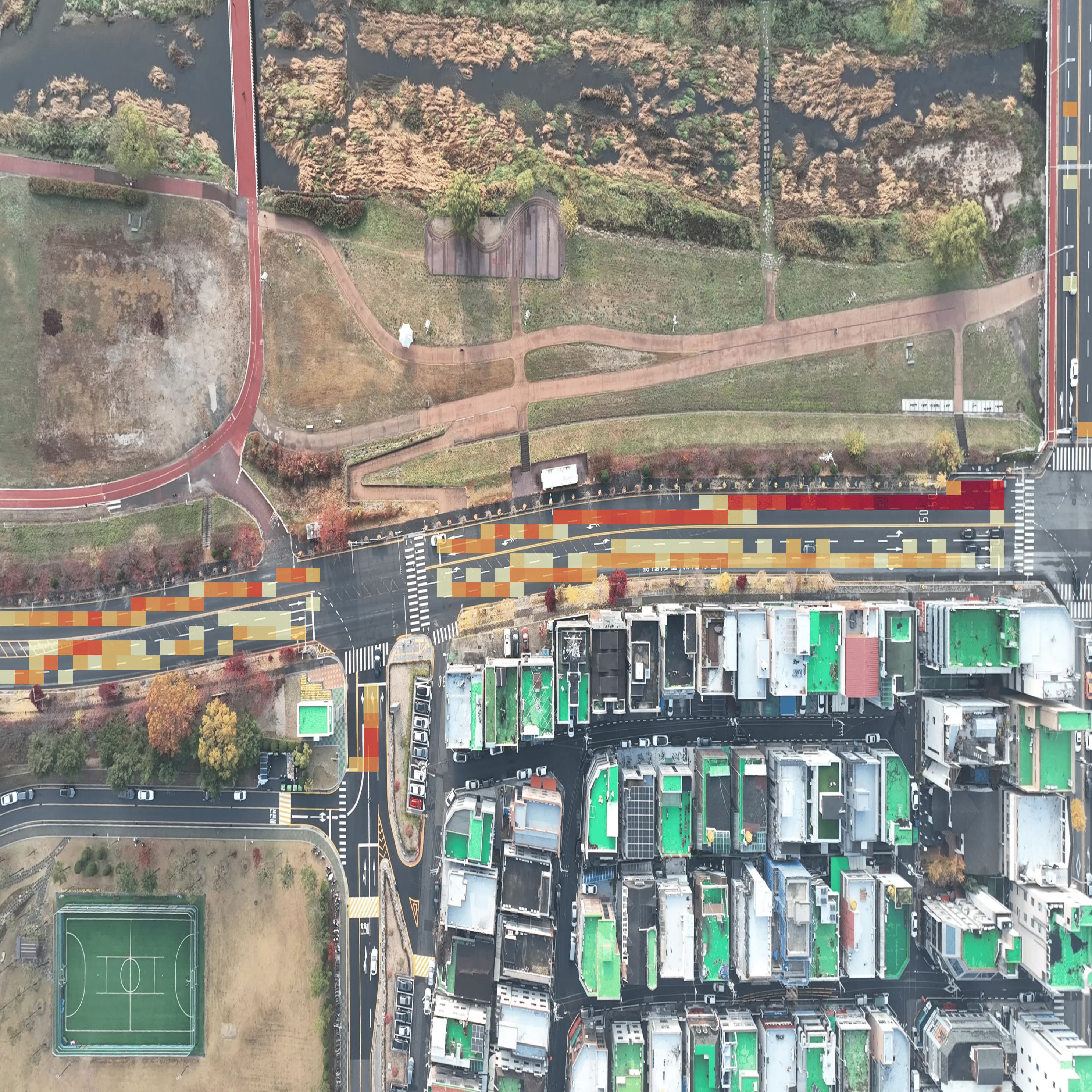}
        \caption{Site D}
        \label{subfig:LC-D}
    \end{subfigure}    
    \hfill
    \begin{subfigure}[b]{0.3\textwidth}
        \includegraphics[width=\textwidth]{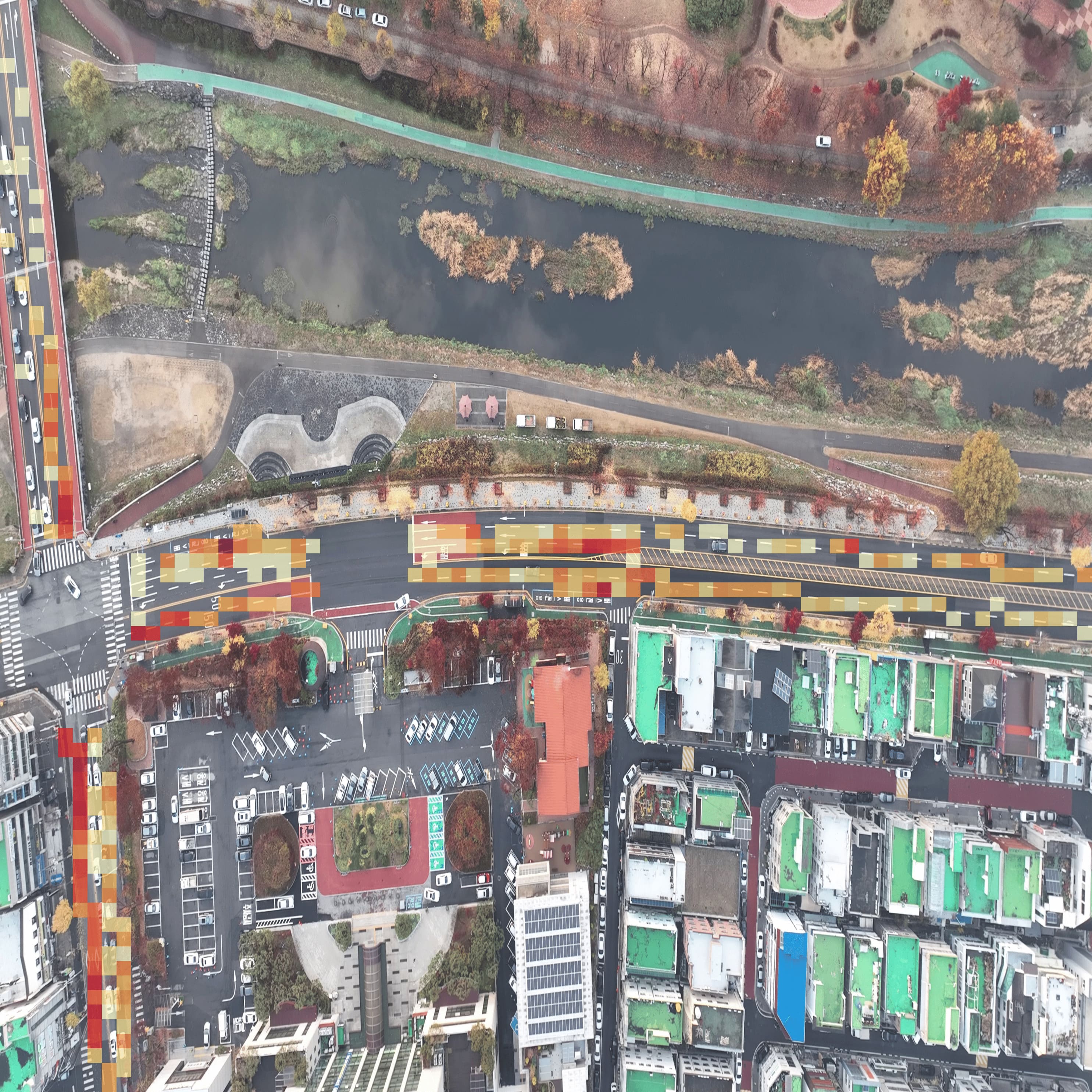}
        \caption{Site E}
        \label{subfig:LC-E}
    \end{subfigure}
    \hfill
    \begin{subfigure}[b]{0.3\textwidth}
        \includegraphics[width=\textwidth]{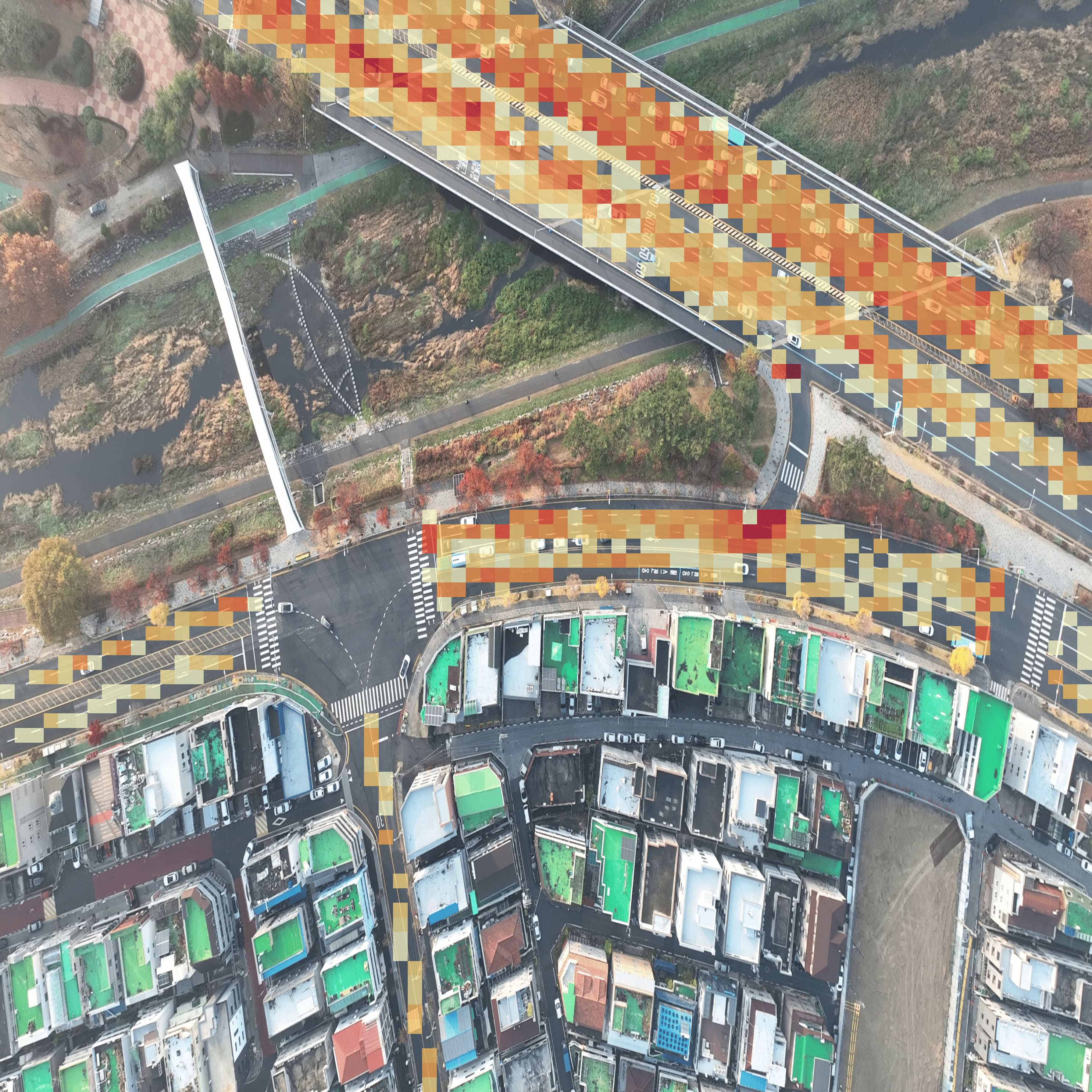}
        \caption{Site F}
        \label{subfig:LC-F}
    \end{subfigure} 
    \hfill
    \begin{subfigure}[b]{0.3\textwidth}
        \includegraphics[width=\textwidth]{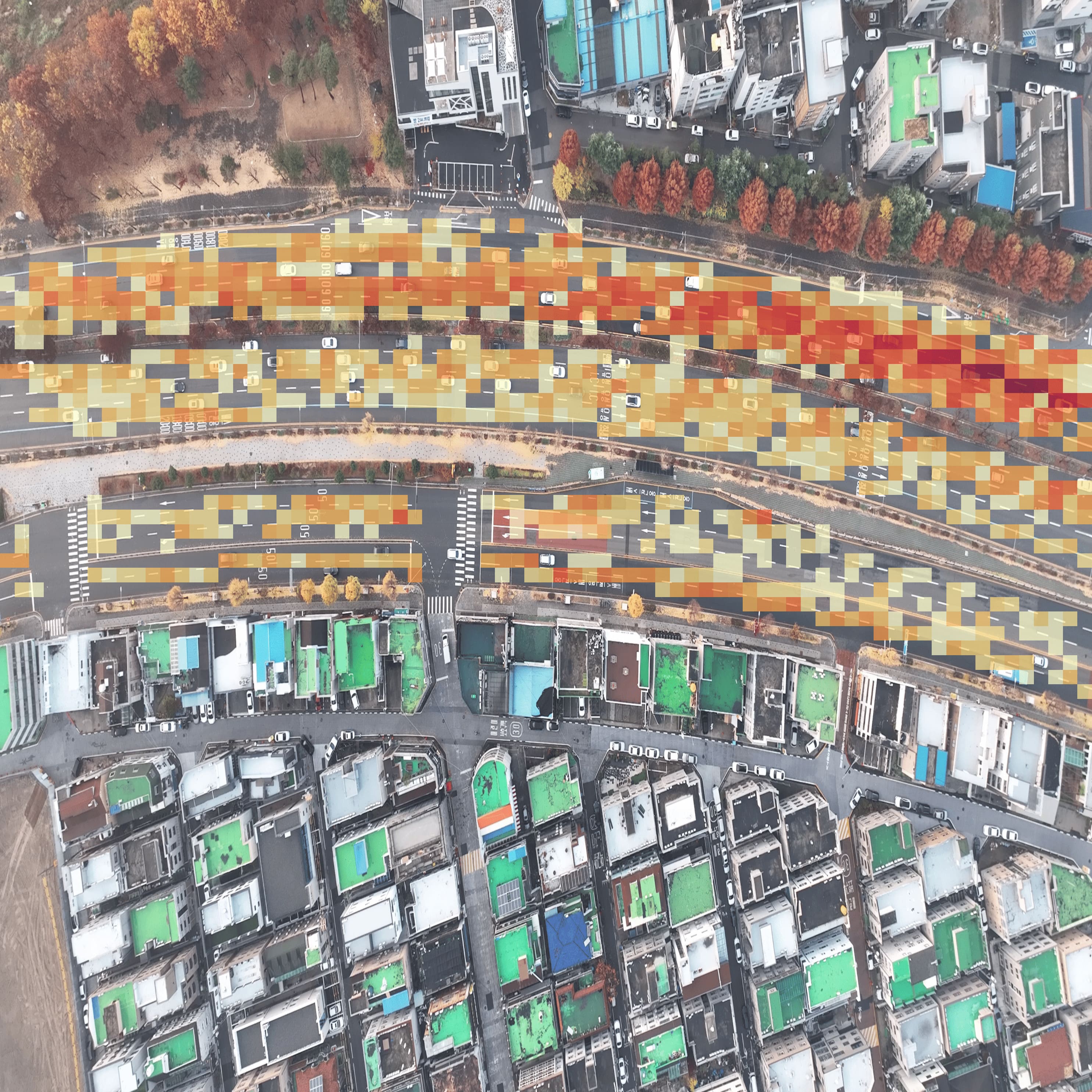}
        \caption{Site G}
        \label{subfig:LC-G}
    \end{subfigure} 
    \hfill
    \begin{subfigure}[b]{0.3\textwidth}
        \includegraphics[width=\textwidth]{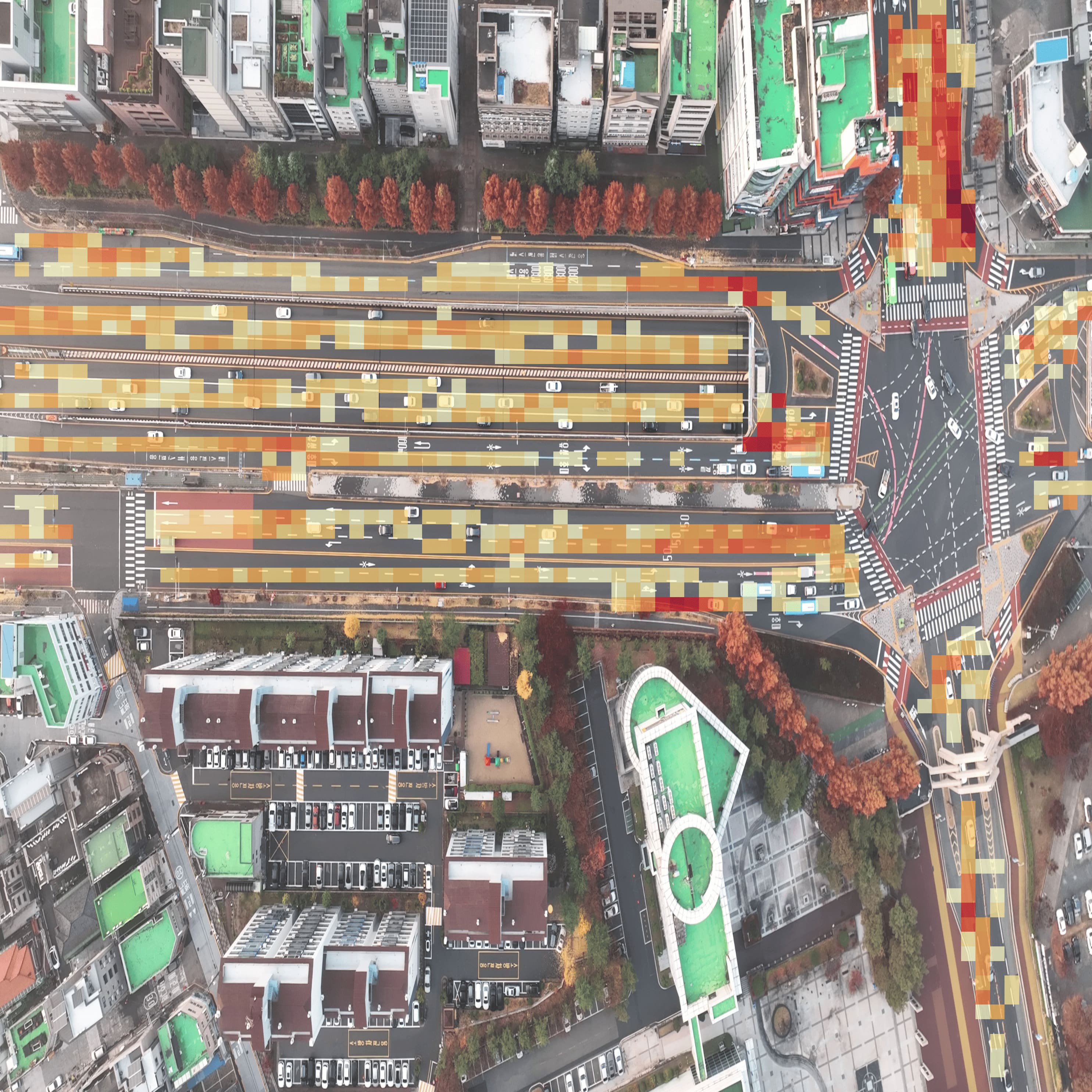}
        \caption{Site H}
        \label{subfig:LC-H}
    \end{subfigure}
    \hfill
    \begin{subfigure}[b]{0.3\textwidth}
        \includegraphics[width=\textwidth]{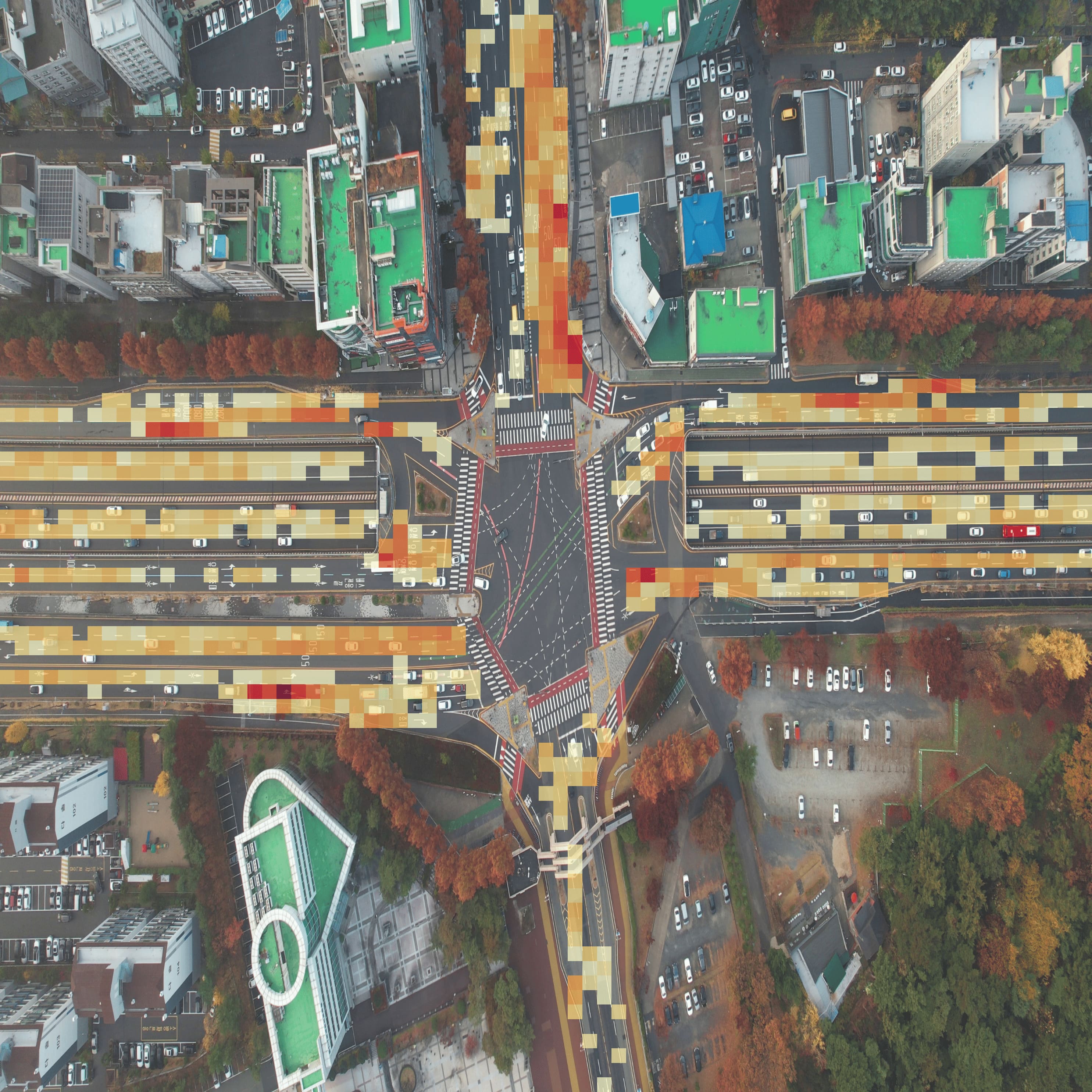}
        \caption{Site I}
        \label{subfig:LC-I}
    \end{subfigure}

    \begin{minipage}{\textwidth}
        \centering
        \includegraphics[width=\textwidth]{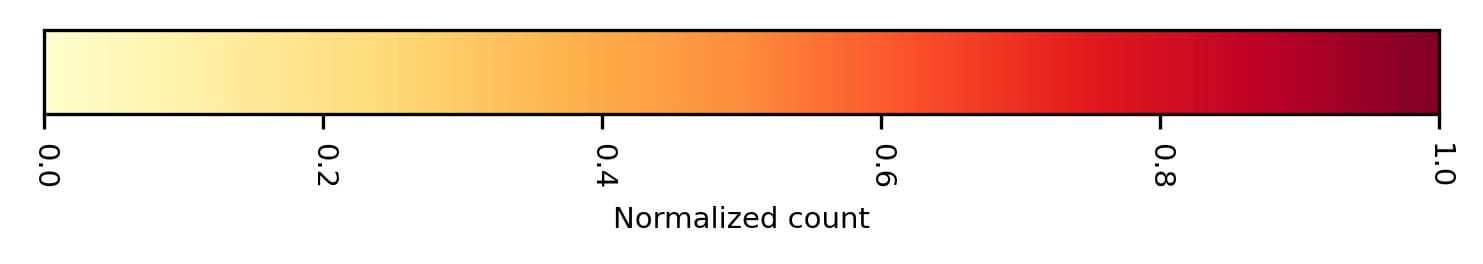}
        \label{subfig:LC-cbar}
    \end{minipage}
    \caption{Distributions of lane change positions in each site.}

    \label{fig:LC}
\end{figure}

Lane-change behavior is a critical aspect of driver interaction, allowing dynamic adaptation to fluctuating traffic conditions. Frequent or abrupt lane-changes are often linked to increased collision risk and traffic flow instability, making them essential indicators for both safety assessments and traffic flow analysis. In this study, lane-changing events are identified through frame-by-frame changes in vehicle orientation, enabling precise detection of lateral maneuvers and their spatial distribution.
Figure~\ref{fig:LC} illustrates the spatial distribution of lane-change frequency across all sites, revealing that lane-change activity is highly concentrated around merging zones, where geometric constraints induce more frequent lateral movements.

\begin{figure}[!ht]
    \centering
    \begin{subfigure}[b]{0.30\textwidth}
        \reflectbox{\rotatebox{180}{\includegraphics[width=\textwidth]{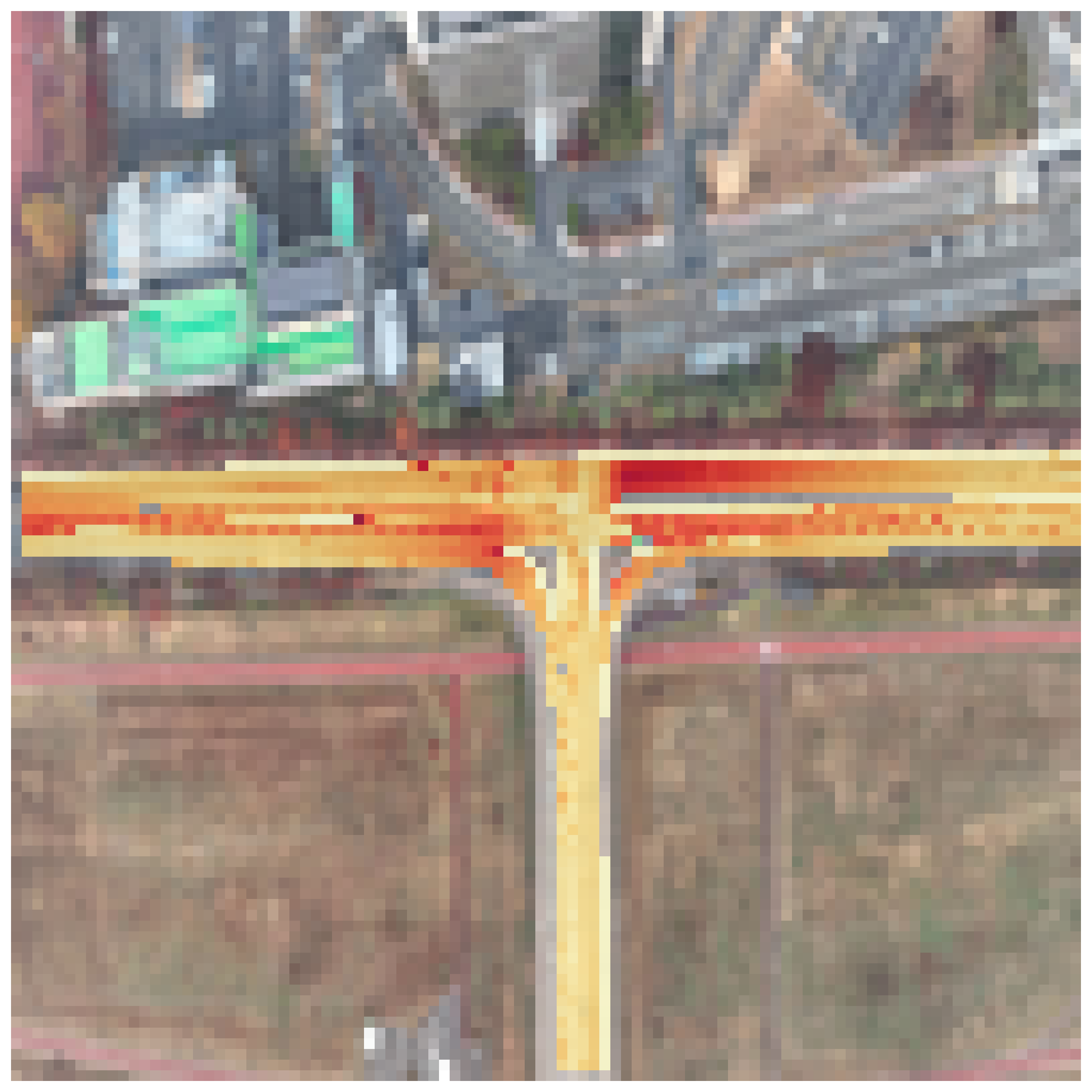}}}
        \caption{Site A}
        \label{subfig:ttc-A}
    \end{subfigure}
    \hfill
    \begin{subfigure}[b]{0.30\textwidth}
        \reflectbox{\rotatebox{180}{\includegraphics[width=\textwidth]{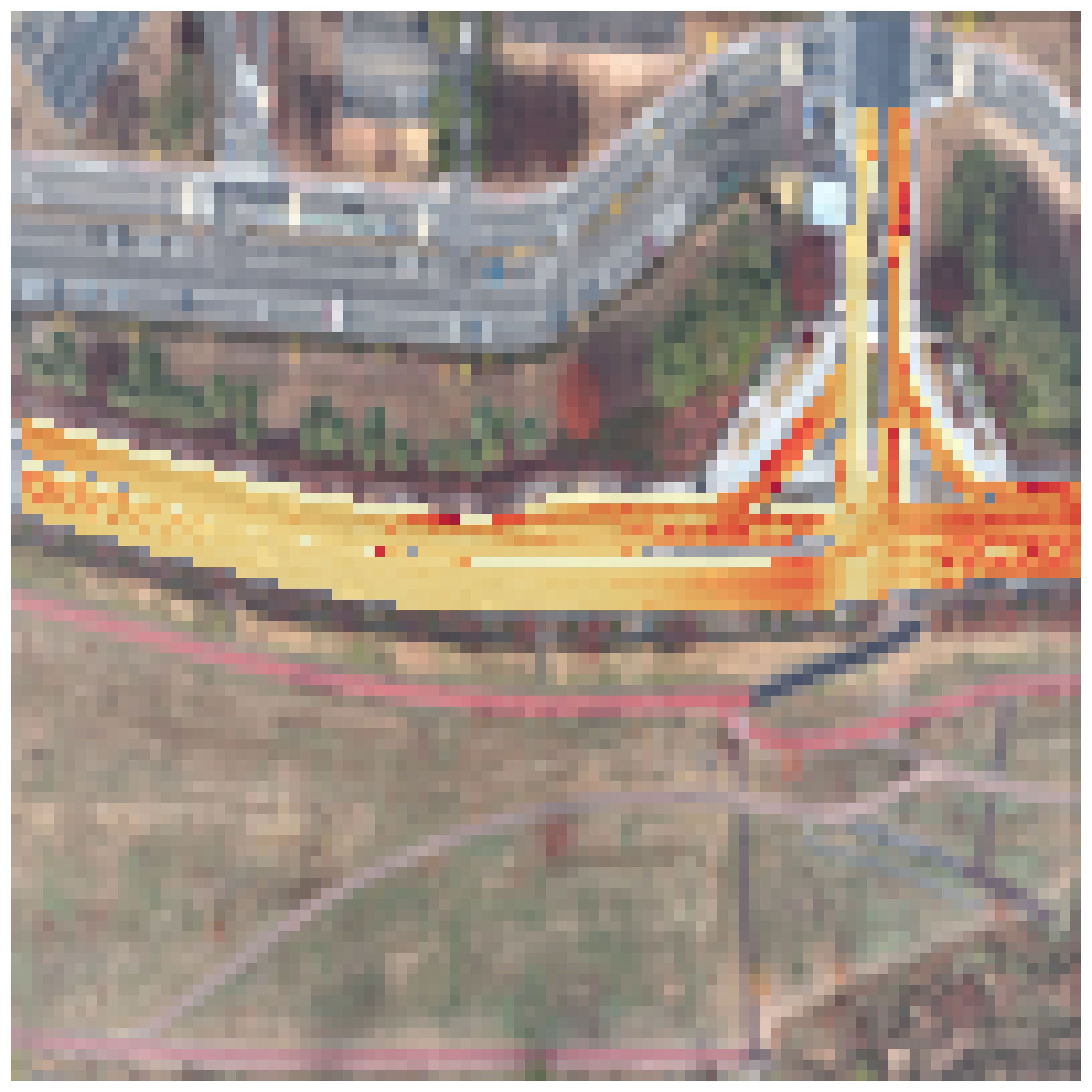}}}
        \caption{Site B}
        \label{subfig:ttc-B}
    \end{subfigure}
    \hfill
    \begin{subfigure}[b]{0.3\textwidth}
        \reflectbox{\rotatebox{180}{\includegraphics[width=\textwidth]{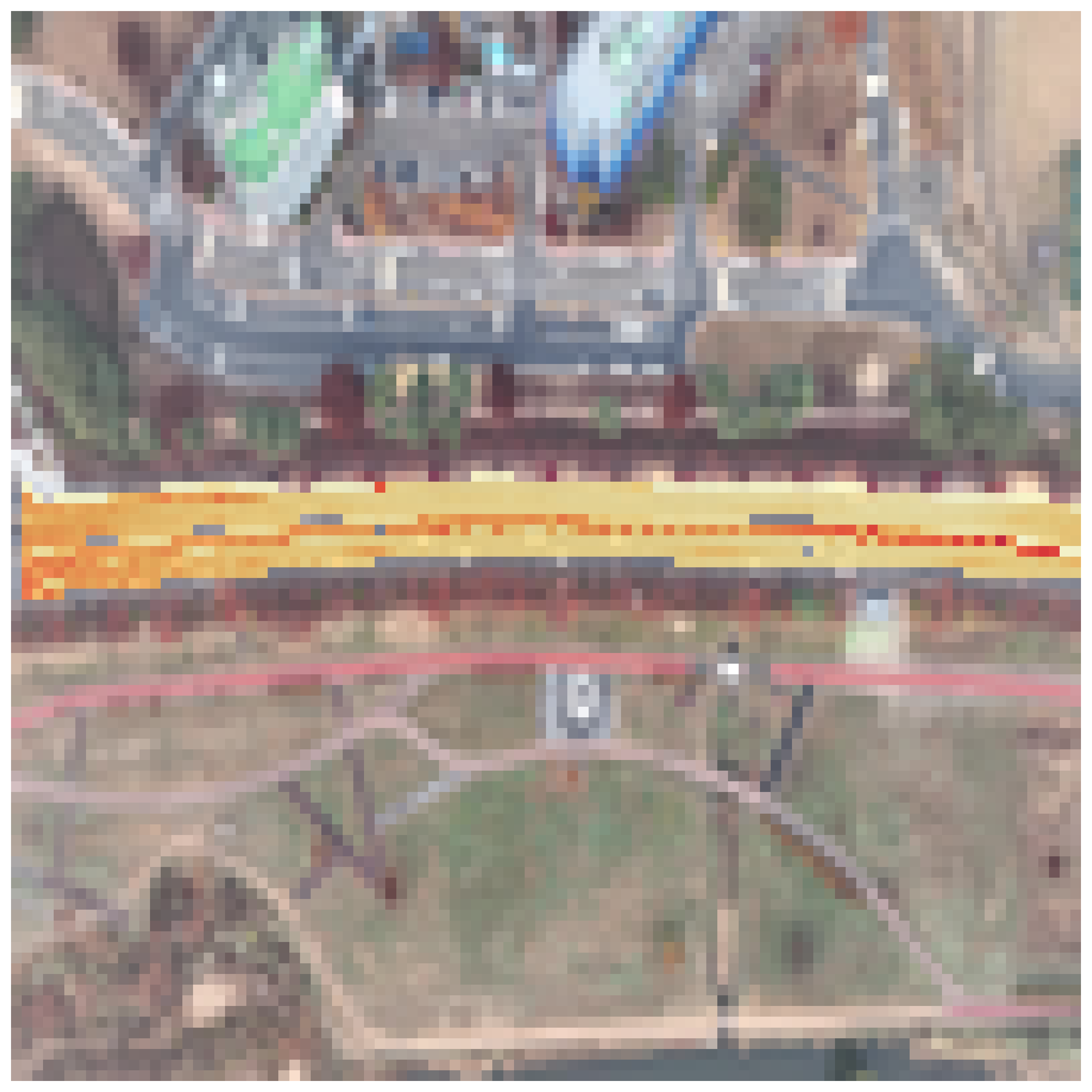}}}
        \caption{Site C*}
        \label{subfig:ttc-C}
    \end{subfigure} 
    \hfill
    \begin{subfigure}[b]{0.3\textwidth}
        \reflectbox{\rotatebox{180}{\includegraphics[width=\textwidth]{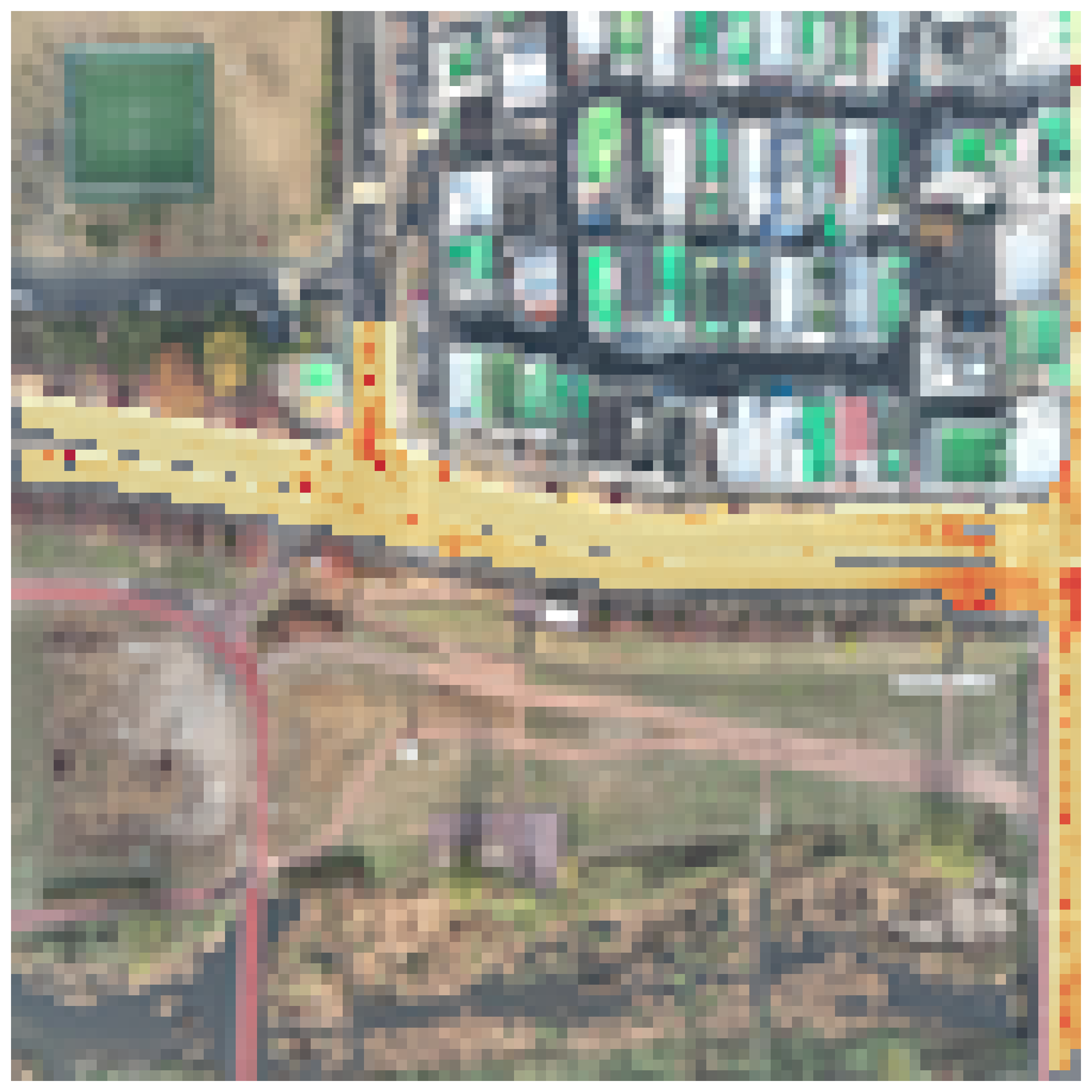}}}
        \caption{Site D}
        \label{subfig:ttc-D}
    \end{subfigure}    
    \hfill
    \begin{subfigure}[b]{0.3\textwidth}
        \reflectbox{\rotatebox{180}{\includegraphics[width=\textwidth]{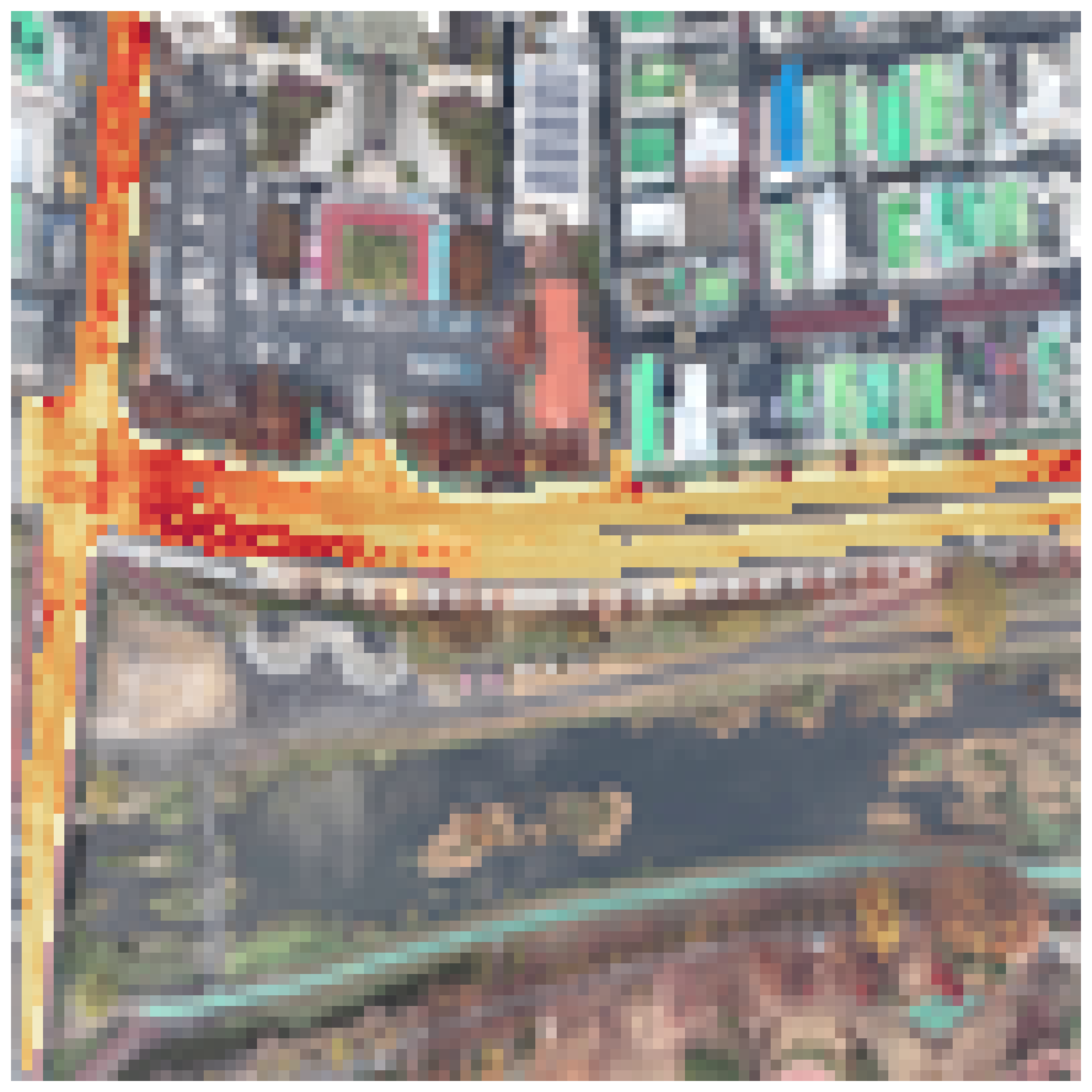}}}
        \caption{Site E}
        \label{subfig:ttc-E}
    \end{subfigure}
    \hfill
    \begin{subfigure}[b]{0.3\textwidth}
        \reflectbox{\rotatebox{180}{\includegraphics[width=\textwidth]{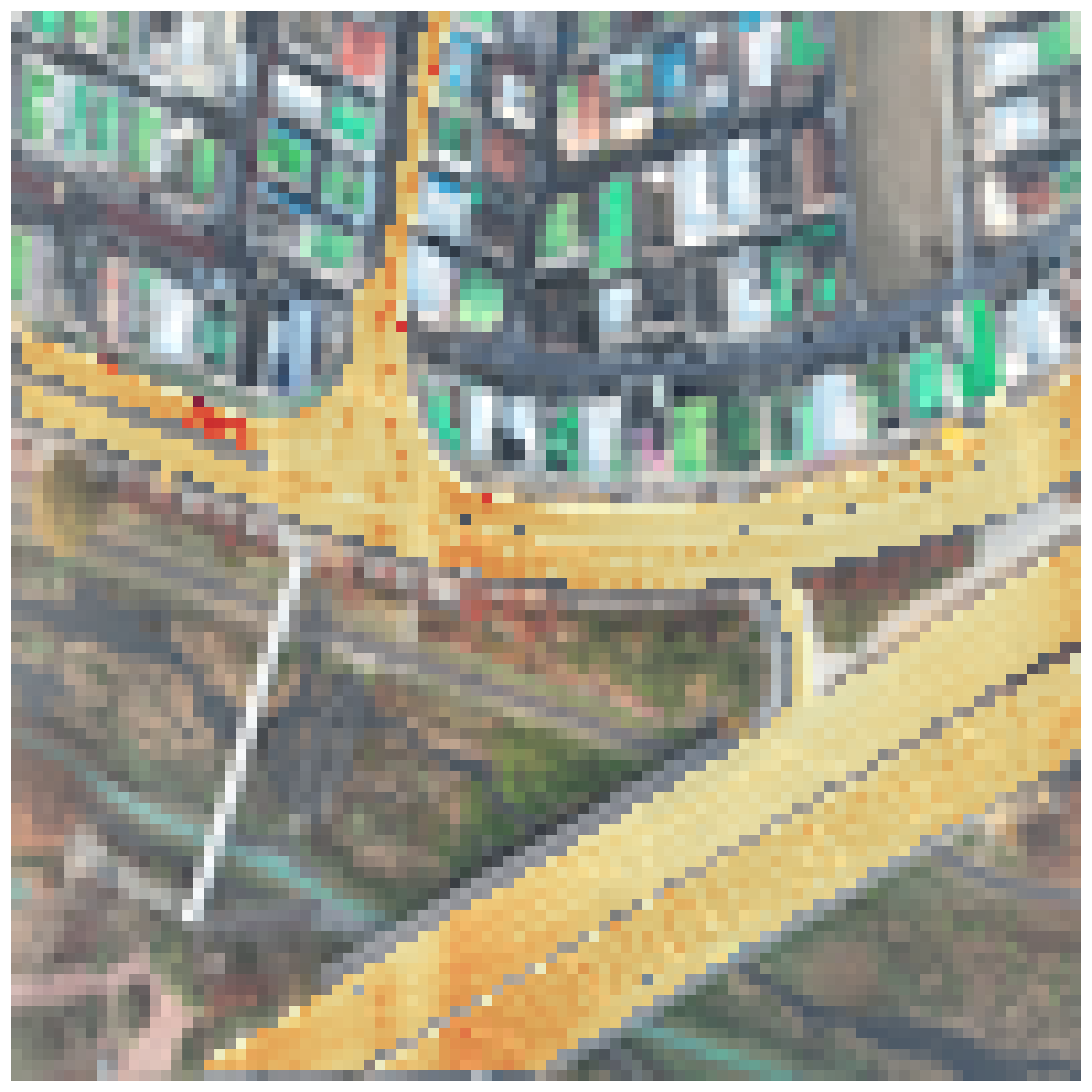}}}
        \caption{Site F}
        \label{subfig:ttc-F}
    \end{subfigure} 
    \hfill
    \begin{subfigure}[b]{0.3\textwidth}
        \reflectbox{\rotatebox{180}{\includegraphics[width=\textwidth]{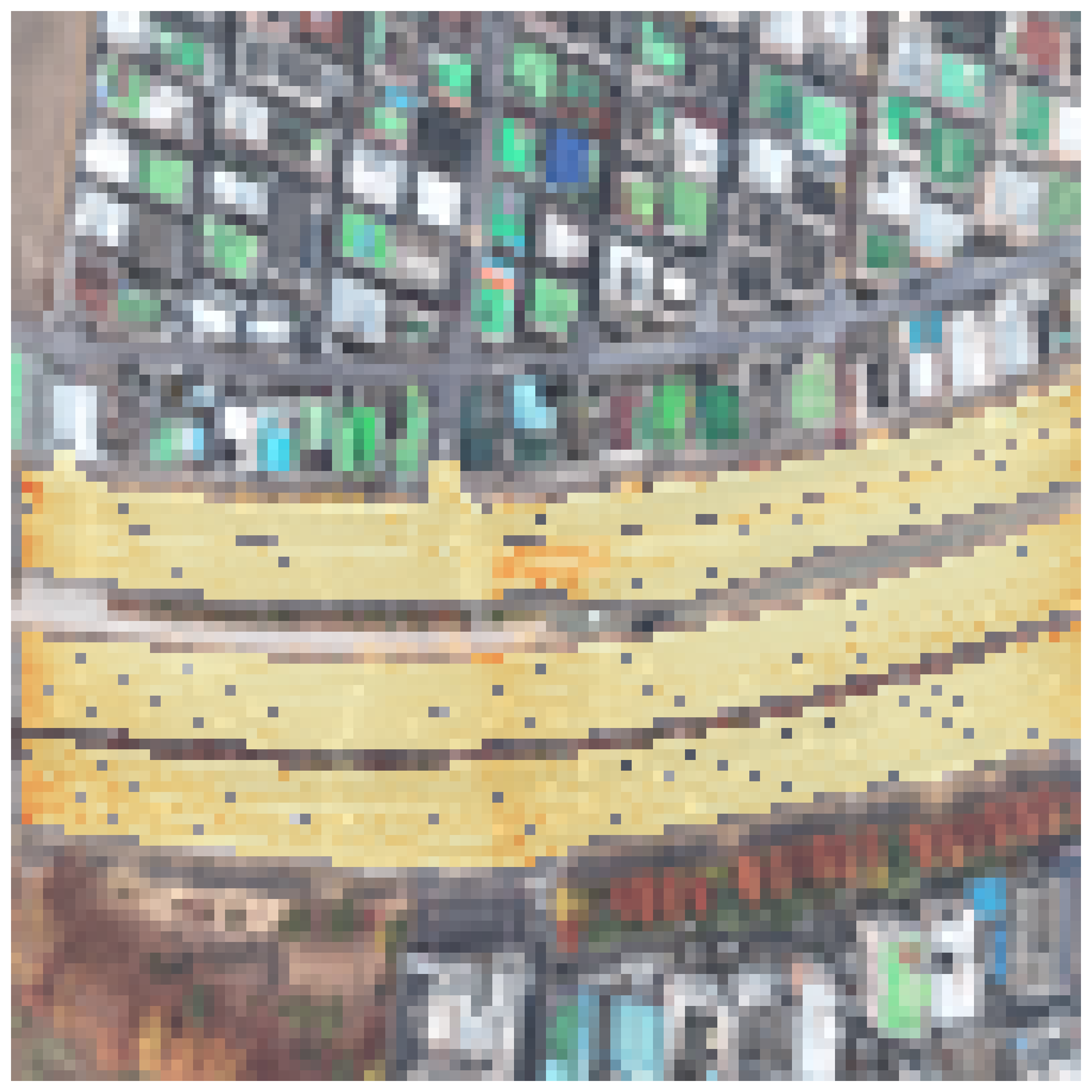}}}
        \caption{Site G*}
        \label{subfig:ttc-G}
    \end{subfigure} 
    \hfill
    \begin{subfigure}[b]{0.3\textwidth}
        \reflectbox{\rotatebox{180}{\includegraphics[width=\textwidth]{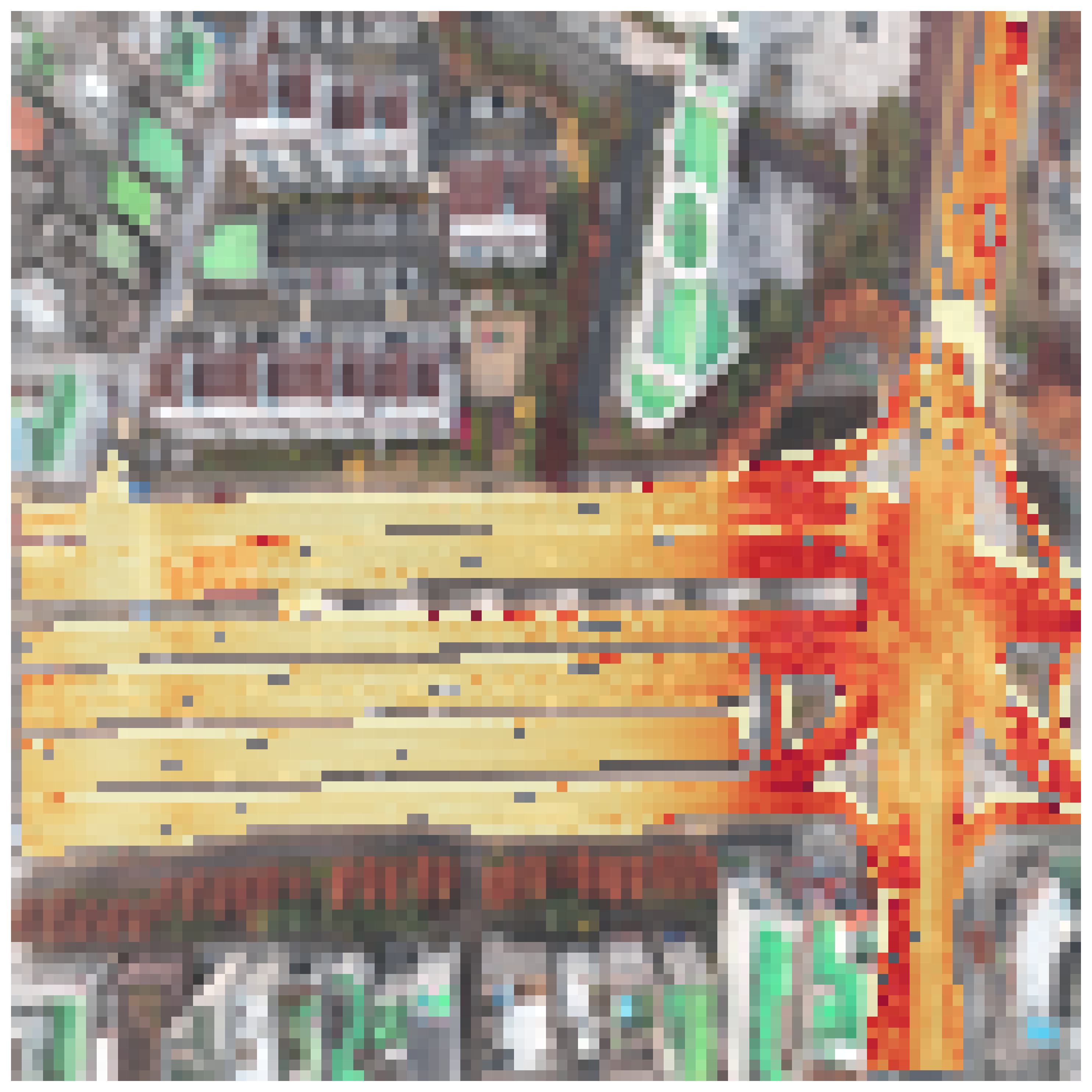}}}
        \caption{Site H}
        \label{subfig:ttc-H}
    \end{subfigure}
    \hfill
    \begin{subfigure}[b]{0.3\textwidth}
        \reflectbox{\rotatebox{180}{\includegraphics[width=\textwidth]{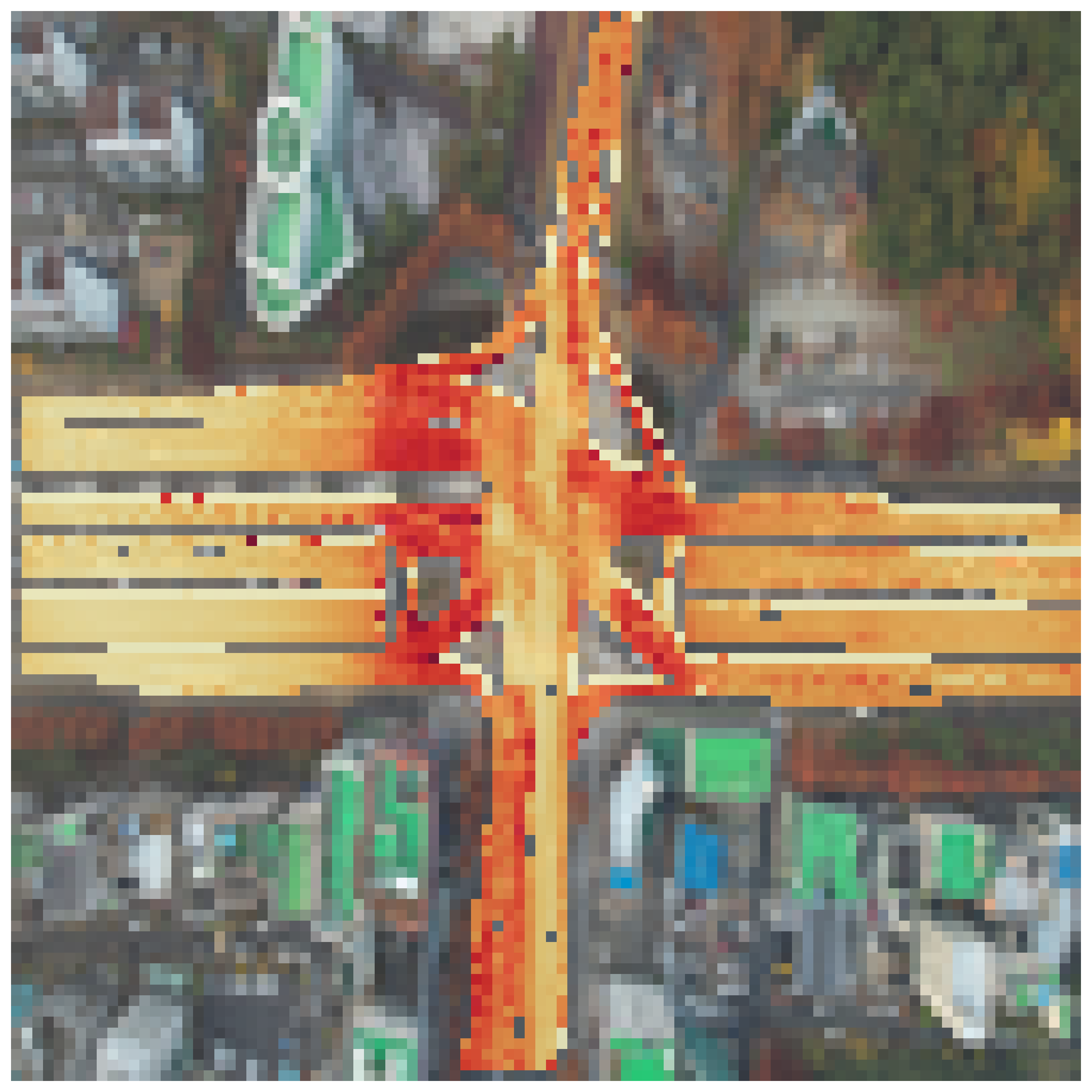}}}
        \caption{Site I}
        \label{subfig:ttc-I}
    \end{subfigure}

    \begin{minipage}{\textwidth}
        \centering
        \includegraphics[width=\textwidth]{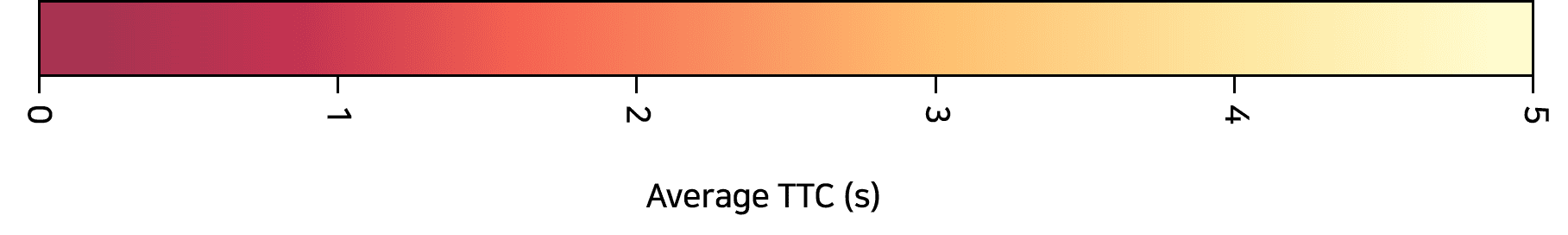}
        \label{subfig:ttc-cbar}
    \end{minipage}
    
    \caption{Distributions of TTC in each site. Sites marked with $^{*}$ indicate locations with no intersection.}
    \label{fig:ttc}
\end{figure}

Time-to-Collision (TTC) is a widely used metric for assessing traffic safety in car-following scenarios. It quantifies the time remaining before a potential collision if vehicles continue at their current speeds. As a measure of temporal proximity, TTC reflects the immediacy of risk in longitudinal interactions, with lower values indicating a higher likelihood of rear-end collisions. In this study, TTC values are computed for each vehicle based on its relative speed and distance to the preceding or following vehicle. Figure~\ref{fig:ttc} presents the spatial distribution of TTC values across the nine sites. The results show that approximately 10 \~ 15\% TTC values are below 2 seconds, with an average TTC of approximately 5 seconds across target sites. These low TTC values are predominately observed in the merging sections, where frequent lane changes occur. 

In summary, the proposed DRIFT dataset offers extensive spatial coverage and high temporal resolution, addressing traditional limitations in microscopic traffic analysis. It provides detailed driving behavior metrics—such as lane-change frequency and acceleration profiles—that enhance simulation realism and enable continuous monitoring for early safety risk detection and proactive interventions.

\subsection{Meso-level characteristics}
In this section, we present meso-level traffic analysis using the DRIFT open dataset, exemplifying its applicability through flow-density and time-space diagrams. First, the flow-density diagram illustrates how traffic flow varies with changes in vehicle density, effectively representing diverse traffic states ranging from free-flow conditions to congestion. 

\begin{figure}[!ht]
    \centering
    \begin{subfigure}[b]{0.60\textwidth}
        \centering
        \includegraphics[width=\textwidth]{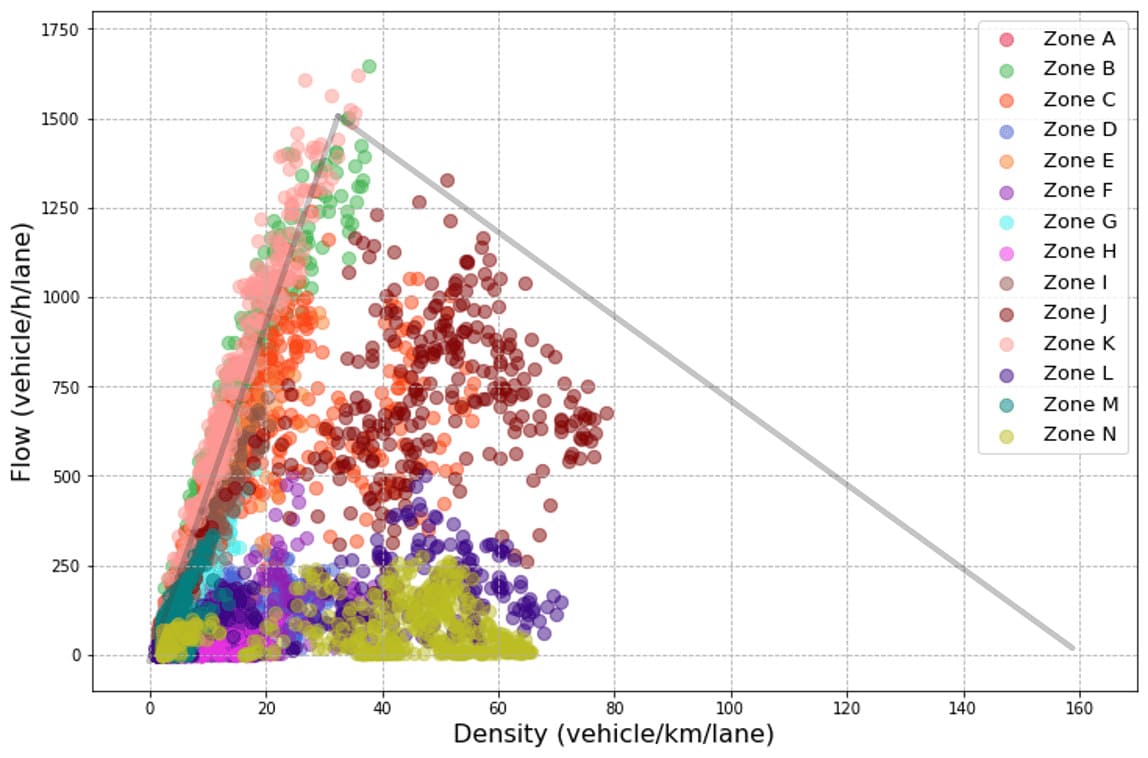}
        \caption{Flow-density relationship}
        \label{fig:q-k-analysis-I}
    \end{subfigure}
    \hfill
    \begin{subfigure}[b]{0.35\textwidth}
        \centering
        \includegraphics[width=\textwidth]{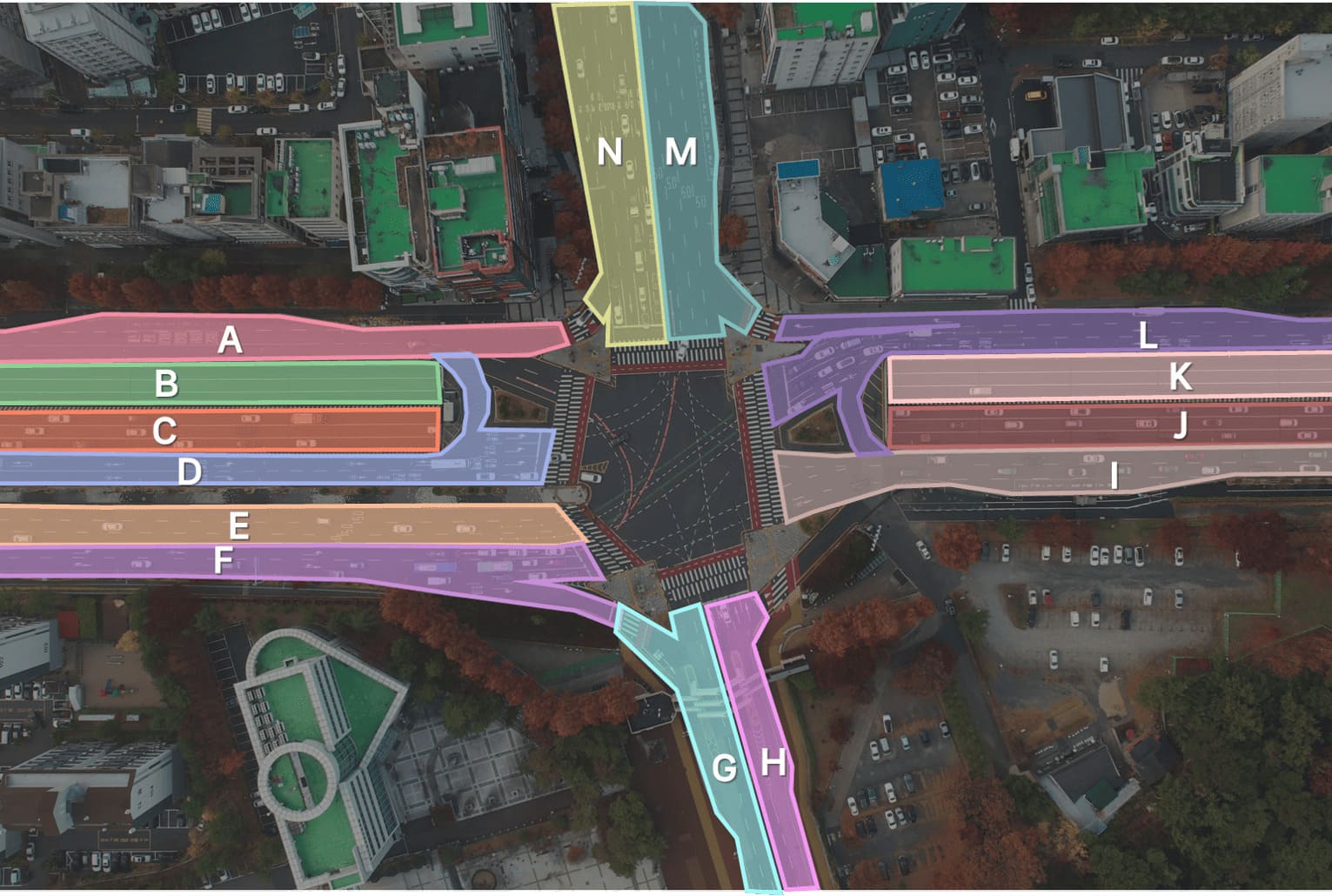}
        \caption{Lane RoI information}
        \label{fig:q-k-analysis-II}
    \end{subfigure}
    \caption{Flow-density diagram highlighting critical densities at different lane RoI in Site I}
    \label{fig:q-k-analysis}
\end{figure}

In this analysis, the flow-density relationship is examined for Site I, which features a complex intersection structure (see Figure~\ref{fig:q-k-analysis}). Flow and density are computed using Edie’s definition \cite{gazis1968traffic}, with vehicle trajectory data aggregated at 5-second intervals. As illustrated in Figure~\ref{fig:q-k-analysis}, the flow–density relationships for individual zones within Site I exhibit notable variation. Specifically,  higher flow values are observed in zones such as B, C, K, and J, which correspond to underpasses, while zones characterized by frequent right-turns and U-turns displayed comparatively lower flow values. These results visualize that road geometry significantly influences congestion formation and enable detailed analysis of traffic characteristics and driving patterns across different segments of the intersection.

\begin{figure}[!htbp]
    \centering
    
    \begin{minipage}{\textwidth} 
        \centering
        \includegraphics[width=0.95\textwidth]{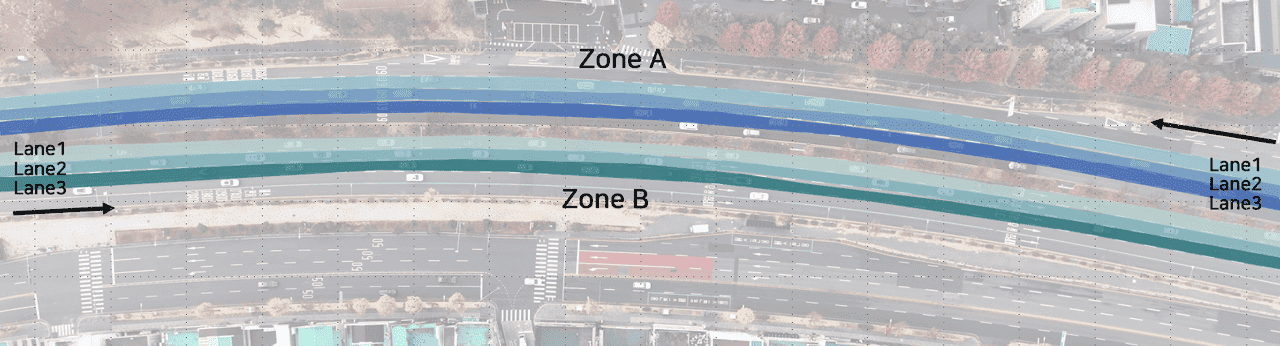}
    \end{minipage}
    
    \vspace{0.2em}  
    
    \begin{minipage}{0.49\textwidth} 
        \centering
        \begin{subfigure}[b]{\textwidth}
            \includegraphics[width=\textwidth]{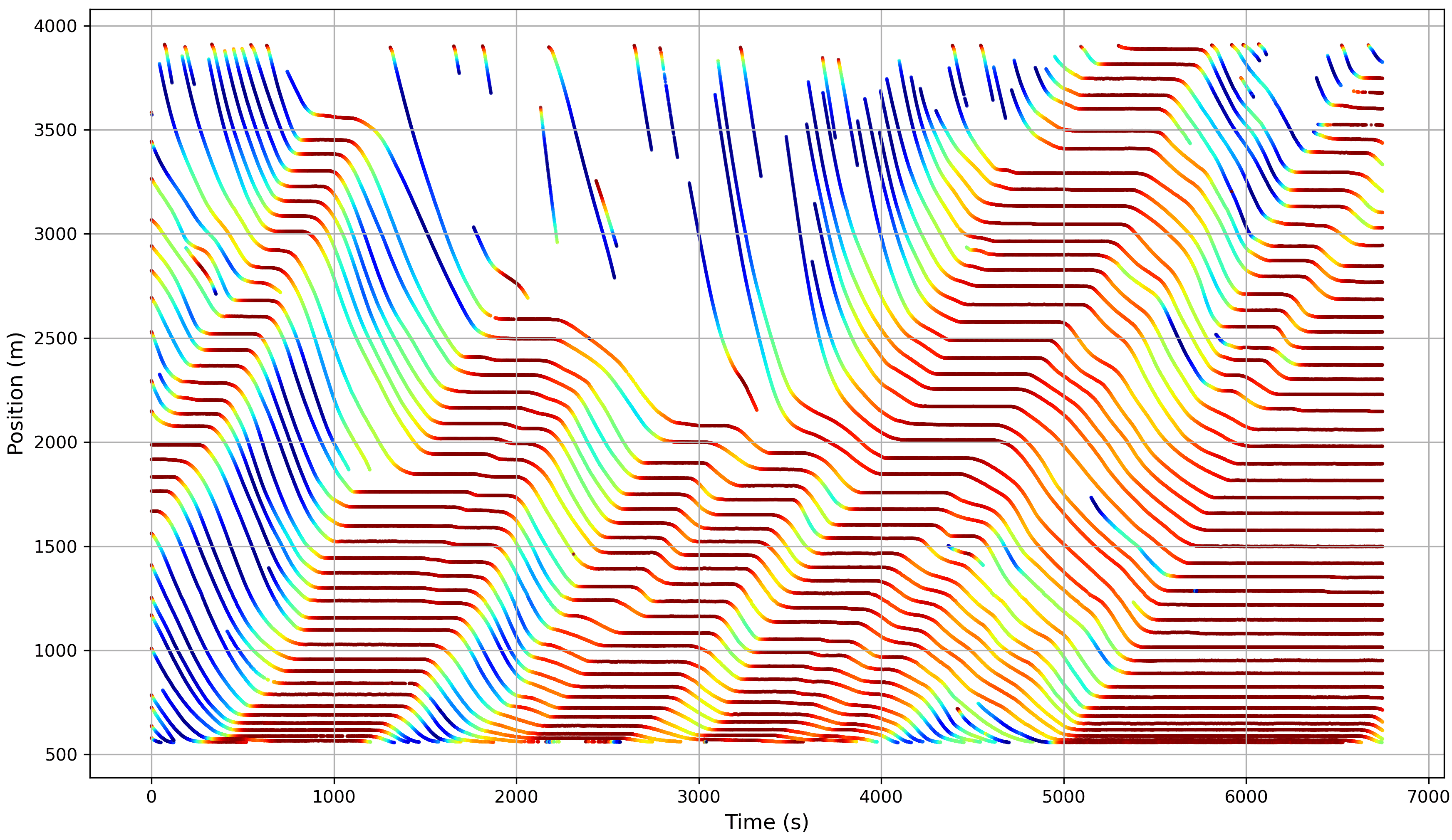}            
            \caption{Zone A - Lane 1}
            \label{subfig:A}
        \end{subfigure}
        \vfill
        \begin{subfigure}[b]{\textwidth}
            \includegraphics[width=\textwidth]{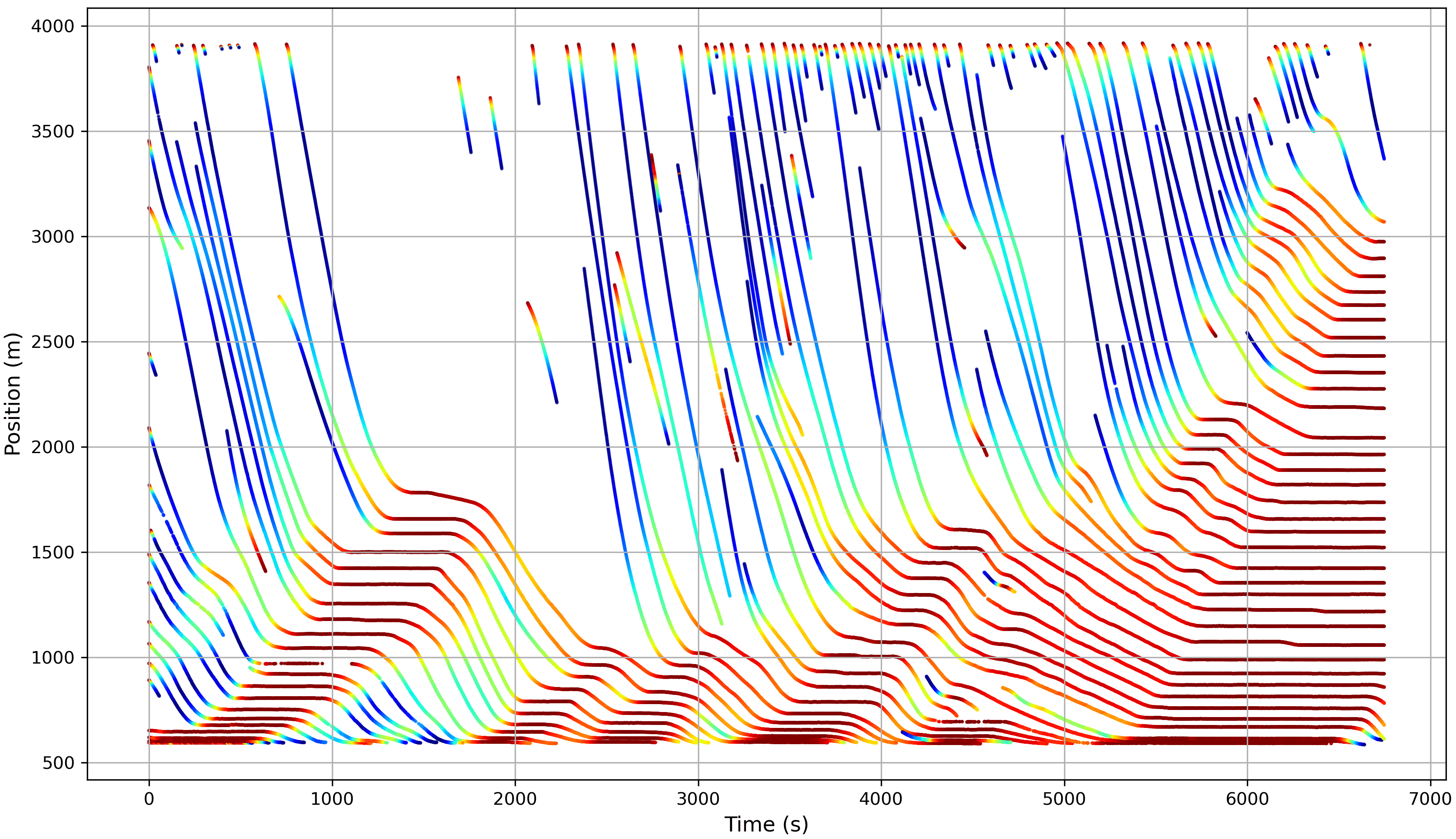}
            \caption{Zone A - Lane 2}
            \label{subfig:B}
        \end{subfigure}
        \vfill
        \begin{subfigure}[b]{\textwidth}
            \includegraphics[width=\textwidth]{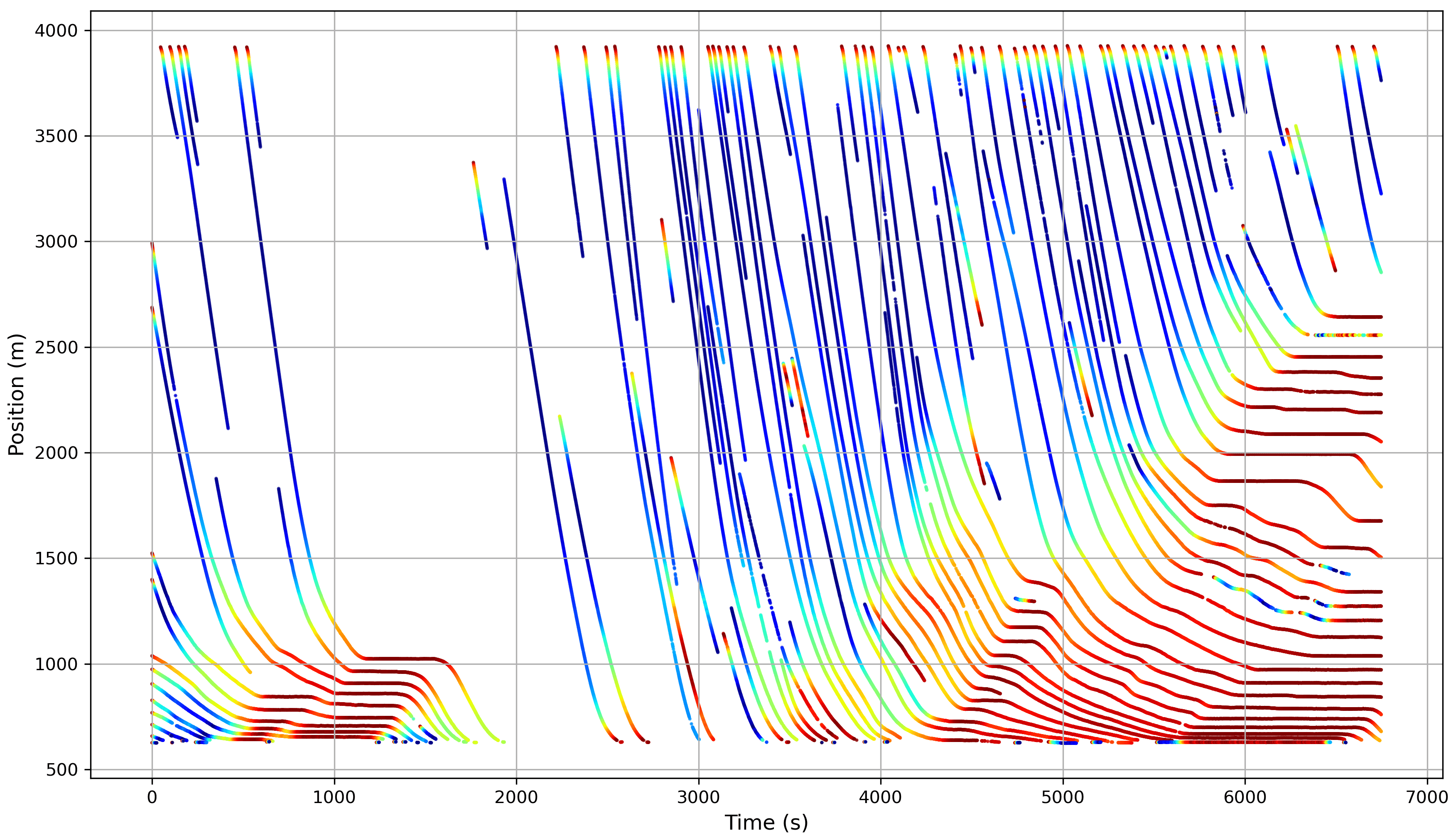}
            \caption{Zone A - Lane 3}
            \label{subfig:C}
        \end{subfigure}
    \end{minipage}
    \hfill
    \begin{minipage}{0.49\textwidth} 
        \centering
        \begin{subfigure}[b]{\textwidth}
            \includegraphics[width=\textwidth]{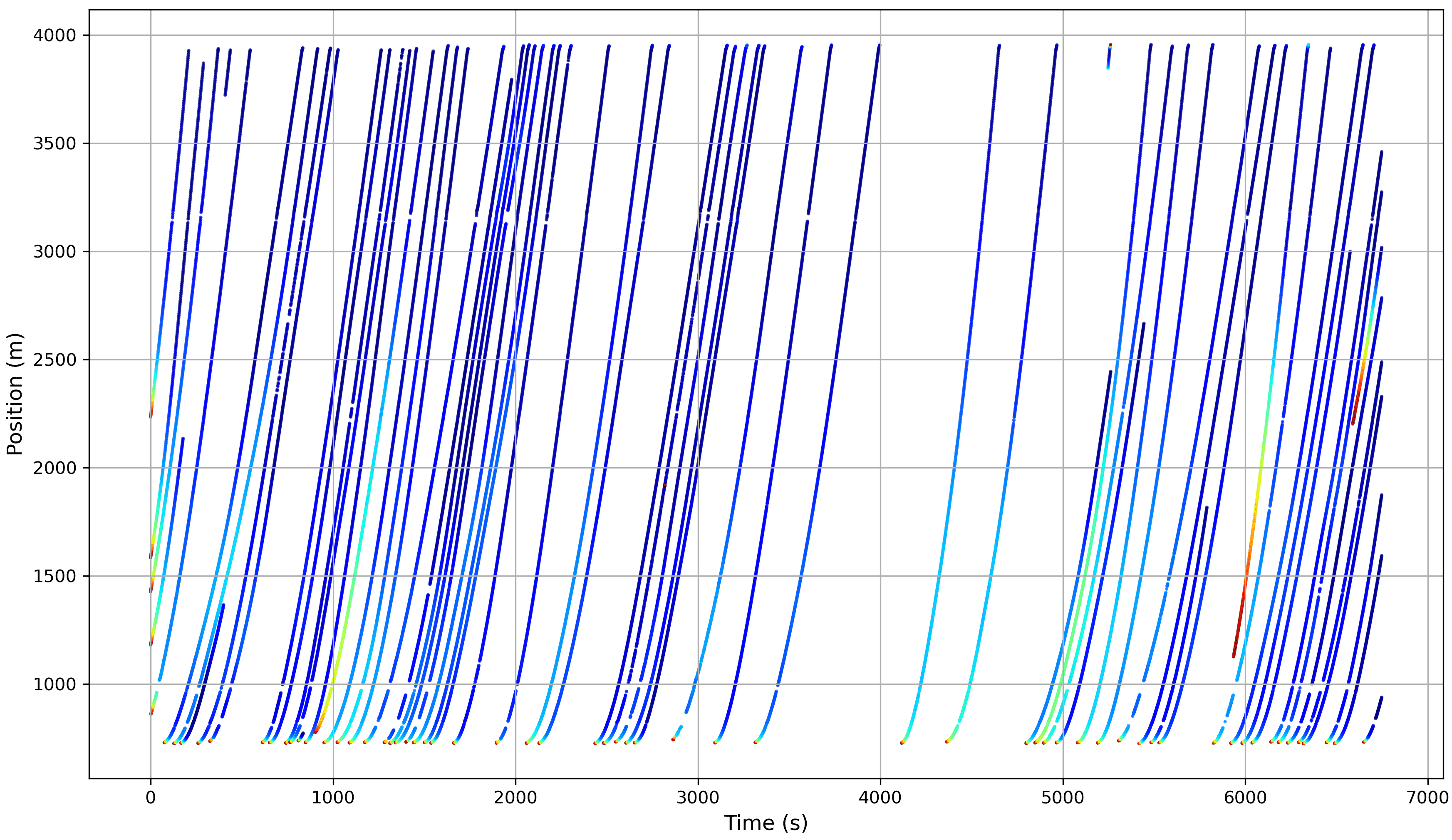}
            \caption{Zone B - Lane 1}
            \label{subfig:D}
        \end{subfigure}
        \vfill
        \begin{subfigure}[b]{\textwidth}
            \includegraphics[width=\textwidth]{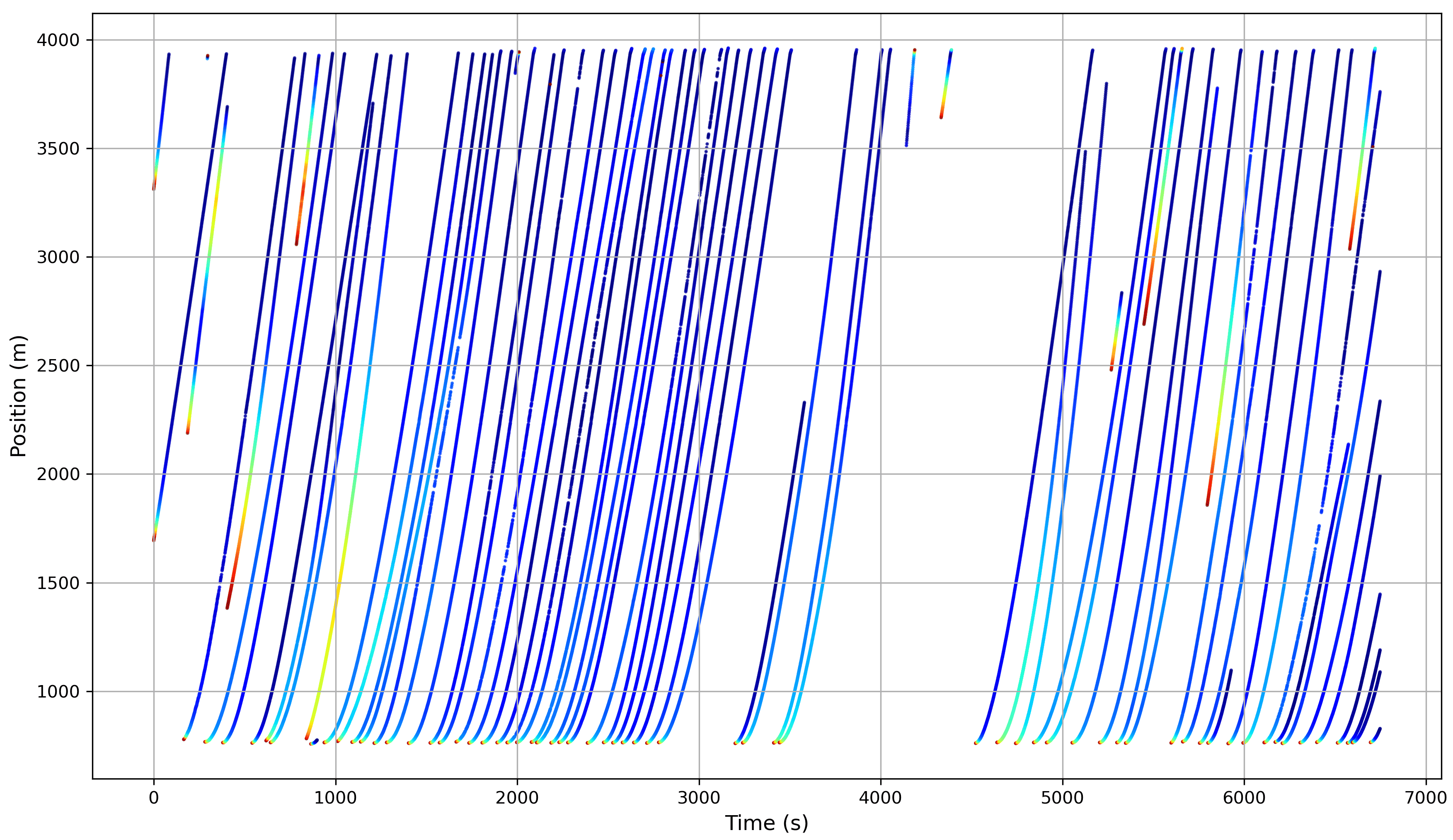}
            \caption{Zone B - Lane 2}
            \label{subfig:E}
        \end{subfigure}
        \vfill
        \begin{subfigure}[b]{\textwidth}
            \includegraphics[width=\textwidth]{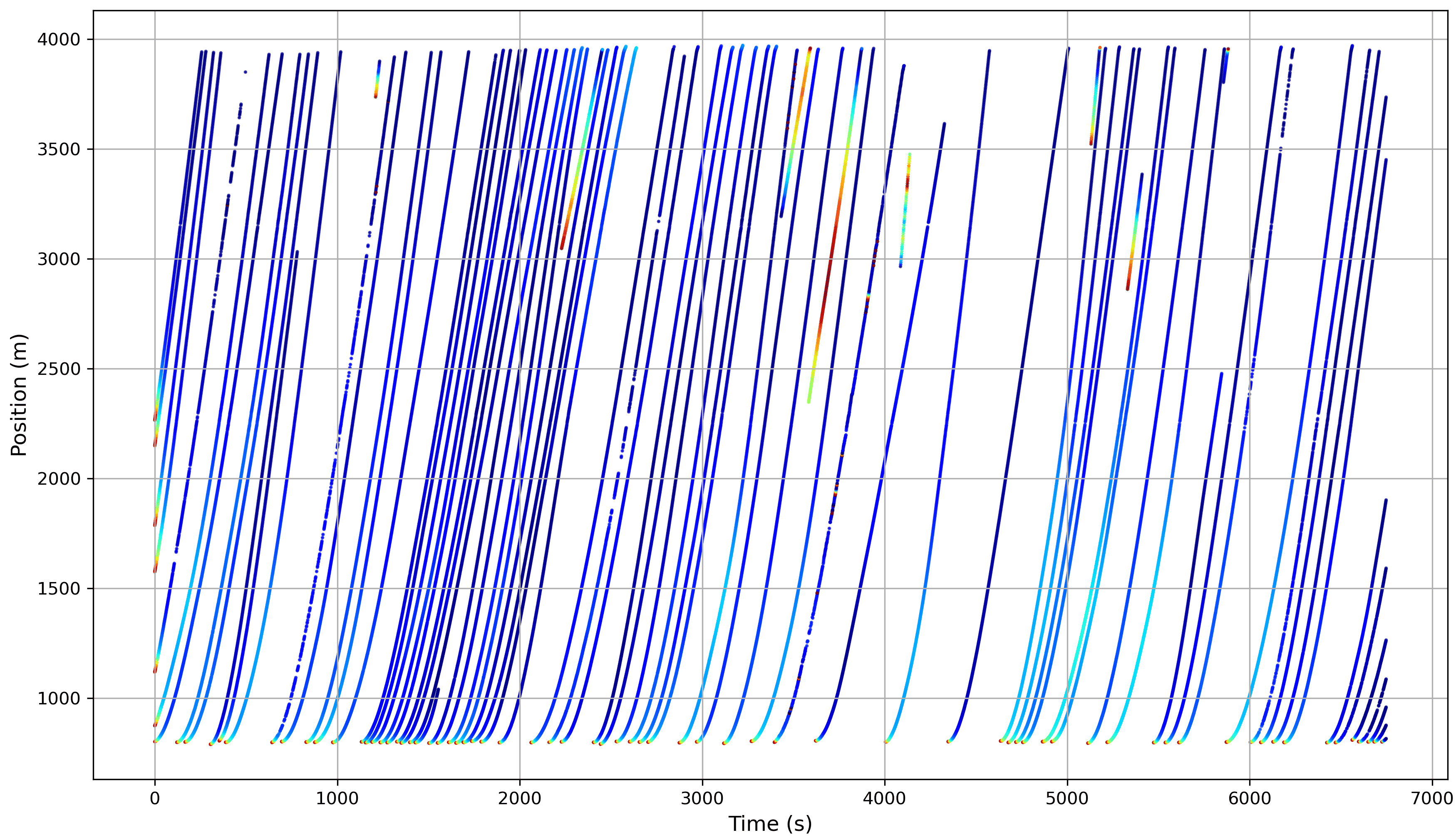}
            \caption{Zone B - Lane 3}
            \label{subfig:F}
        \end{subfigure}
    \end{minipage}
    
    \vspace{0.2em}  
    
    \begin{minipage}{\textwidth} 
        \centering
        \begin{subfigure}[b]{\textwidth}
            \includegraphics[width=\textwidth]{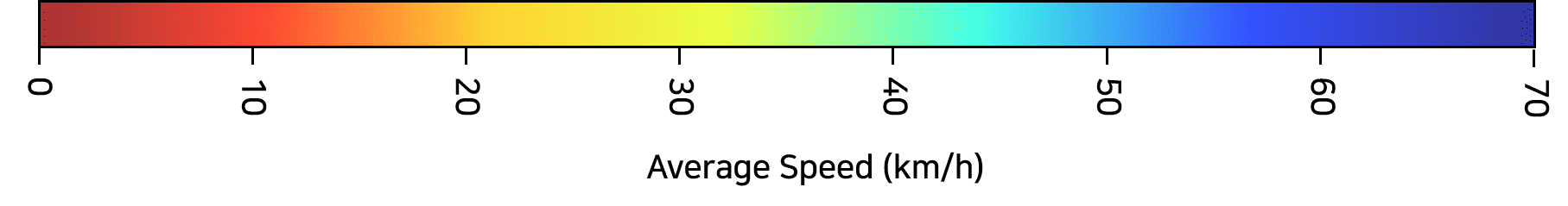}
            \label{subfig:G}
        \end{subfigure}
    \end{minipage}

    \caption{Time-space diagram of Site G illustrating traffic dynamics}
    \label{fig:time-space}
\end{figure}

In addition, a time-space diagram was depicted for Site G to visualize the spatiotemporal evolution of vehicle speeds and positions during the morning peak period. This analysis enabled the precise identification of congestion propagation patterns, bottleneck locations, and critical lane-change zones. Figure~\ref{fig:time-space} illustrates clear directional asymmetries in traffic flow, with markedly different patterns observed between eastbound and westbound movements. For example, while eastbound traffic maintained relatively stable speeds, the westbound direction exhibited repeated slowdowns and queue formation near signalized intersections. These differences reflect underlying commuting demand asymmetries, which become particularly pronounced during peak hours. 
Such findings offer meaningful implications for time-of-day-based traffic control strategies. Understanding when and where congestion forms—particularly with respect to directional flow imbalances—can inform the development of adaptive signal timing, dynamic lane assignments, and localized lane-change restrictions. Moreover, this such analysis contributes significantly to validating and analyzing theoretically how congestion evolves to adjacent segments in urban networks, leading to micro- and meso-multimodal policies such as optimizing lane configurations and establishing no-change zones. Overall, the DRIFT open dataset provides a robust basis for detailed mesoscopic traffic analyses, effectively serving as an analytical bridge connecting microscopic vehicle behaviors with macroscopic traffic flow patterns.

\subsection{Macro-level characteristics}

\begin{figure}[!ht]
    \centering
    \begin{minipage}{\textwidth}
        \centering
        \includegraphics[width=\textwidth]{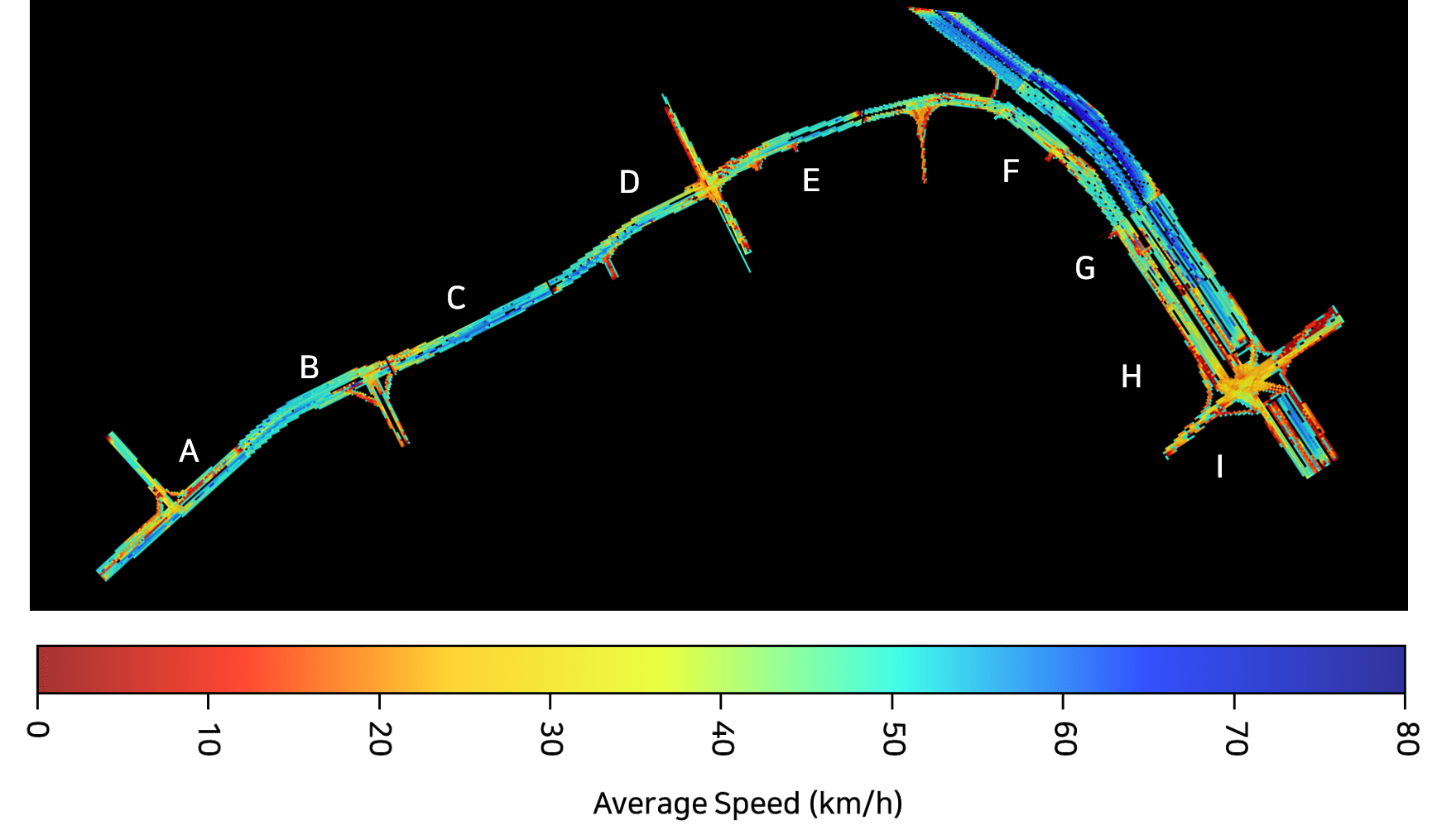}
    \end{minipage}
    \caption{Speed heatmap illustrating velocity distribution across drone sites (A–I)}
    \label{fig:speed-heatmap}
\end{figure}

At the macroscopic level, we performed a grid-based analysis of average lane-level speeds across the study area to support network-wide traffic flow evaluation. Prior research has examined urban traffic conditions using metrics such as traffic volume, average travel speed, and congestion indices, often focusing on the relationships among these aggregated indicators. While conventional approaches capture broad traffic trends effectively, they often fail to address the detailed micro-level driving behaviors like lane changes, sudden deceleration, and braking that underpin these macro-level phenomena. The DRIFT dataset bridges this analytical gap by providing detailed vehicle trajectory data that allow for quantifying the cumulative impact of micro-level behaviors on macro-level congestion. 

Figure~\ref{fig:speed-heatmap} is a visualization of lane-by-lane speed variations at various locations, highlighting areas of recurring bottlenecks. This enables time-based analysis of average speeds and helps to understand how congestion spreads spatially, or how interactions between individual vehicles affect traffic flow across the entire network. Such analysis helps to evaluate the causes of bottlenecks from a microscopic behavioral perspective between vehicles. It also supports the development of adaptive traffic management systems that adjust signal cycles in response to micro-level driver behavior, such as sudden deceleration. In addition, this type of visualization-based analysis is promising for use in scenario-based simulations, including digital twin applications. Overall, DRIFT expands the horizons of transport research by enabling integrated studies at all scopes, from microscopic driver behavior to the macroscopic transport system and infrastructure performance.

\subsection{Discussion}
This study aimed to structure and publicly release the DRIFT open dataset, comprising urban traffic trajectory data collected using drones. A total of 81,699 vehicle trajectories were obtained from a 2.6-km urban corridor encompassing nine interconnected intersections, and the entire data processing pipeline—including object detection and tracking—was fully automated using computer vision techniques. Orthophoto matching using homography was also performed to accurately map trajectories to real-world coordinates. Consequently, the DRIFT dataset provides high-resolution trajectory data with spatial and temporal continuity, addressing limitations inherent in conventional sensor-based methods.

The DRIFT dataset is accompanied by analytical tools and illustrative examples for conducting micro-, meso-, and macro-level traffic analyses. Specifically, at the micro level, vehicle behavior and collision risks can be quantitatively evaluated through lane-change distributions and TTC values. At the meso level, congestion emergence and propagation mechanisms can be examined using flow-density and time-space diagrams. At the macro level, grid-based average speed distributions enable visual identification of bottlenecks and urban congestion patterns.

In fact, several drone-based traffic trajectory datasets have previously been introduced, each providing valuable contributions to the research community. Highway-oriented datasets such as highD \citep{krajewski2018highd} and exiD \citep{moers2022exid} offer detailed lane-change and trajectory data; datasets like rounD \citep{krajewski2020round}, inD \citep{bock2020ind}, and INTERACTION \citep{zhan2019interaction} provide insights into complex interactions among diverse road users at intersections; and large-scale urban datasets such as pNEUMA \citep{barmpounakis2020new}, CitySim \citep{zheng2024citysim}, and Songdo \citep{fonod_2025_13828408} cover extensive urban networks involving multiple intersections. Nevertheless, existing datasets often face limitations, such as difficulties in capturing intricate urban vehicle behaviors, limited spatial continuity due to a focus on isolated intersections or restricted roadway segments, a lack of rigorous validation of extracted trajectories, as well as trajectory inconsistencies or insufficient preprocessing, thus frequently necessitating further refinement prior to practical analyses.

In contrast, the DRIFT dataset incorporates extensive preprocessing—including trajectory validation, stabilization, and error correction—to deliver a refined dataset that is immediately applicable for analysis and modeling without further preprocessing. Moreover, by employing polygon-based OBBs, DRIFT ensures precise tracking of vehicle orientations, especially in intersection areas and along curved roadway segments. Additionally, by capturing a continuously connected network of intersections, DRIFT facilitates integrated multi-scale analyses of urban traffic dynamics. To further enhance usability, pre-trained detection models and visualization tools are provided alongside the dataset.

Nevertheless, the DRIFT dataset remains focused on a specific urban corridor and does not cover an entire urban network. Its applicability may also be partially limited by environmental factors such as weather conditions and visual obstructions. Despite these limitations, DRIFT’s availability as an open-access resource with high reproducibility and usability is expected to substantially support empirical urban traffic research and contribute to the advancement of effective traffic management strategies.

\newpage
\section{Conclusion}\label{chap:conclusion} 

This study presents the DRIFT dataset, a drone-based trajectory dataset designed to ensure spatial continuity of traffic flows across interconnected urban intersections. Unlike existing datasets, which typically focus on isolated intersections or specific roadway segments, DRIFT overcomes spatial discontinuities by collecting data across multiple interconnected intersections. By applying OBB-based object detection at an altitude of 250 meters, the dataset achieves robust vehicle tracking performance. The detection accuracy (99.2\%, mAP@50 of 0.994) and trajectory extraction accuracy (96.98\%) have been validated, allowing researchers to consistently analyze interaction-level dynamics, congestion propagation patterns, and overall traffic flow behaviors at micro-, meso-, and macro-scales.

Although various drone-based datasets have previously been introduced and have served as foundational resources, many require significant preprocessing and data refinement before practical use. In contrast, the DRIFT dataset addresses these limitations by providing structured, cleaned trajectory data readily applicable for immediate analysis and modeling tasks without additional preprocessing.

The publicly released DRIFT dataset is expected to support a variety of applications, including traffic safety assessments, simulation-based forecasting, and real-time traffic management. Future work will involve expanding geographic coverage and integrating real-time analytics to enhance urban traffic management decision-making capabilities. The DRIFT dataset, along with associated analytical tools and pretrained models, has been made publicly available via GitHub, thereby facilitating empirical transportation research and advancing the development of intelligent transportation systems.

\section*{Declarations}
\bmhead{Availability of data and material}
The dataset used in this study is available at our DRIFT open dataset repository (https://github.com/AIxMobility/The-DRIFT).

\bibliography{sn-bibliography} 

\end{document}